\documentclass[11pt,twoside]{article} 
\usepackage{geometry}
\usepackage[style=alphabetic,natbib=true,backend=bibtex,maxnames=10,maxcitenames=1]{biblatex}
\usepackage{hyperref}
\usepackage{url}

\usepackage[utf8]{inputenc} 
\usepackage[T1]{fontenc}    
\usepackage{hyperref}       
\usepackage{url}            
\usepackage{booktabs}       
\usepackage{amsfonts}       
\usepackage{nicefrac}       
\usepackage{microtype}      
\usepackage{xcolor}         

\usepackage{comment}
\usepackage{algorithm}
\usepackage{algpseudocode}
\usepackage{amsthm}
\newtheorem{theorem}{Theorem}
\newtheorem{corollary}{Corollary}[theorem]
\newtheorem{remark}{Remark}
\usepackage{multicol}
\usepackage{xcolor}         
\usepackage{tikz}
\usepackage{tcolorbox,wrapfig}
\usepackage[labelfont={bf},font=small]{caption}
\usepackage{subcaption}
\usepackage{rotating}
\tcbuselibrary{skins}
\usepackage{graphicx}
\usepackage{lipsum}
\usepackage{makecell}
\usepackage{tabu}
\usepackage{wrapfig}
\usepackage{amsmath}
\usepackage{longtable}

\usepackage{hyperref,graphicx,amsmath,amsfonts,amssymb,bm,url,breakurl,epsf,color,MnSymbol,mathbbol,fmtcount,semtrans,caption,multirow,comment, boldline}
\usepackage{wrapfig}
\usepackage{enumitem}
\usepackage{amssymb}

\usepackage[utf8]{inputenc} 
\usepackage[T1]{fontenc}    
\usepackage{url}            
\usepackage{booktabs}       
\usepackage{amsfonts}       
\usepackage{nicefrac}       
\usepackage{microtype}      
\usepackage{dsfont}

\usepackage{xcolor}

\newlist{myitemize}{enumerate}{1}
\setlist[myitemize]{font=\color{darkred}\bfseries\itshape}

  \usepackage{mathtools}

\usepackage{titlesec}

\usepackage{tikz}
\usepackage{pgfplots}
\usetikzlibrary{pgfplots.groupplots}

\usepackage{movie15}

\usepackage{caption}
\usepackage[bottom,hang,flushmargin]{footmisc} 

\newcommand{\tsn}[1]{{\left\vert\kern-0.25ex\left\vert\kern-0.25ex\left\vert #1 
    \right\vert\kern-0.25ex\right\vert\kern-0.25ex\right\vert}}

\definecolor{darkred}{RGB}{150,0,0}
\definecolor{darkgreen}{RGB}{0,150,0}
\definecolor{darkblue}{RGB}{0,0,200}
\hypersetup{colorlinks=true, linkcolor=darkred, citecolor=darkgreen, urlcolor=darkblue}

\newenvironment{fminipage}%
  {\begin{Sbox}\begin{minipage}}%
  {\end{minipage}\end{Sbox}\fbox{\TheSbox}}

\newtheorem{lemma}{Lemma}[section]

\newtheorem{definition}{Definition}

\newcommand{\maj}{{\text{maj}}}

\DeclareMathOperator{\tr}{tr}

\newcommand{\cut}[1]{\textcolor{red}{}}

\newcommand{\M}{\mathbf{M}}

\newcommand{\G}{\mathbf{G}}

\newcommand{\Hb}{{\mathbf{H}}}

\newcommand{\Vb}{\mathbf{V}}

\newcommand{\mub}{\boldsymbol{\mu}}

\newcommand{\thetab}{{\boldsymbol{\theta}}}

\newcommand{\x}{\mathbf{x}}

\newcommand{\w}{\mathbf{w}}

\newcommand{\vb}{\mathbf{v}}

\newcommand{\h}{\mathbf{h}}

\newcommand{\Sc}{{\mathcal{S}}}
\newcommand{\Bc}{{\mathcal{B}}}

\newcommand{\Lc}{\mathcal{L}}

\newcommand{\Pc}{\mathcal{P}}

\newcommand{\beq}{\begin{equation}}
\newcommand{\eeq}{\end{equation}}
\newcommand{\bea}{\begin{align}}
\newcommand{\eea}{\end{align}}

\newcommand{\R}{\mathbb{R}}

\newcommand{\nn}{\notag}

\DeclarePairedDelimiterX{\inp}[2]{\langle}{\rangle}{#1, #2}

\newcommand{\Id}{\mathds{I}}

\newcommand{\ones}{\mathbf{1}}

\providecommand{\norm}[1]{\lVert#1\rVert}

\newcommand{\HT}{\textbf{H}}
\newcommand{\hT}{\textbf{h}}
\newcommand{\MT}{\textbf{M}}

\author{Ganesh Ramachandra Kini$^{\ddag}$, Vala Vakilian$^{\dagger}$, Tina Behnia$^{\dagger}$, Jaidev Gill$^{\dagger}$, \\ Christos Thrampoulidis$^\dagger$
\vspace{8pt} 
\\
$^\ddag$University of California, Santa Barbara, USA 
\vspace{3pt}
\\
$^\dagger$University of British Columbia, Canada 
\thanks{This work is supported by an NSERC Discovery Grant, NSF Grant CCF-2009030, and by a CRG8-KAUST award. JG and CT gratefully acknowledge the support of NSERC Undergraduate Student Research Grant.  The authors also acknowledge use of the Sockeye cluster by UBC Advanced Research Computing.
}
}
\title{Symmetric Neural-Collapse Representations with Supervised Contrastive Loss: The Impact of ReLU and Batching}

\begin{document}

\maketitle

\begin{abstract}
  Supervised contrastive loss (SCL) is a competitive and often superior alternative to the cross-entropy loss for classification. While prior studies have demonstrated that both losses yield symmetric training representations under balanced data, this symmetry breaks under class imbalances. This paper presents an intriguing discovery: the introduction of a ReLU activation at the final layer effectively restores the symmetry in SCL-learned representations. We arrive at this finding analytically, by establishing that the global minimizers of an unconstrained features model with SCL loss and entry-wise non-negativity constraints form an orthogonal frame. Extensive experiments conducted across various datasets, architectures, and imbalance scenarios corroborate our finding.  Importantly, our experiments reveal that the inclusion of the ReLU activation restores symmetry without compromising test accuracy. This constitutes the first geometry characterization of SCL under imbalances. Additionally, our analysis and experiments underscore the pivotal role of batch selection strategies in representation geometry. By proving necessary and sufficient conditions for mini-batch choices that ensure invariant symmetric representations, we introduce batch-binding as an efficient strategy that guarantees these conditions hold.

\end{abstract}

\section{Introduction}\label{sec:intro}
 \vspace{-0.2cm}


The prevalence of deep-neural networks (DNNs) has led to a growing research interest in understanding their underlying mechanisms. 
A recent research thread, focusing on classification tasks, explores whether it is possible to describe the structure of weights learned by DNNs when trained beyond zero-training error. The specific characteristics of this structure will depend on the DNN being used, the dataset being trained on, and the chosen optimization hyperparameters. Yet, \emph{is it possible to identify macroscopic structural characteristics that are common among these possibilities?}

In an inspiring study, \cite{NC} demonstrates this is possible for the classifiers and for the embeddings when training with cross-entropy (CE) loss and balanced datasets. Through extensive experiments over multiple architectures and datasets with an equal number of examples per class, they found that the geometries of  classifiers and of centered class-mean embeddings {(outputs of  the last hidden layer)} consistently converge during training to a common simplex equiangular tight frame (ETF), a structure composed of vectors that have equal norms and equal angles between them, with the angles being the maximum possible. Moreover, they observed neural-collapse (NC), a property where embeddings of individual examples from each class converge to their class-mean embedding. 
%

Numerous follow-up studies have delved deeper into explaining this phenomenon and further investigating how the converging geometry changes with class imbalances. 
The unconstrained-features model (UFM), proposed independently by \cite{mixon2020neural,fang2021exploring,graf2021dissecting,lu2020neural}, plays a central role in the majority of these follow-up studies. Specifically, the UFM serves as a theoretical abstraction for DNN training, in which the network architecture is viewed as a powerful black-box that generates embeddings without any restrictions in the last {(hidden)} layer. For CE loss, the UFM minimizes $\min_{\w_c\in\R^d,\h_i \in\R^d} \Lc_{\text{CE}}\big(\{\w_c\}_{c\in[k]},\{\h_i\}_{i\in[n]}\big)$ in which both classifiers $\w_c$ for the $k$ classes and  embeddings $\h_i$ for the $n$ training examples are unconstrainedly optimized over $\R^d$. \cite{zhu2021geometric,graf2021dissecting,fang2021exploring} have verified that the global minimum of this non-convex problem satisfies NC and follows the ETF geometry observed in DNN experiments by \cite{NC}. Recent works by \cite{seli, fang2021exploring,behnia2023implicit} have demonstrated that the global optimum of the UFM changes when classes are imbalanced. Yet, the new solution still predicts the geometry observed in DNN experiments, providing evidence that, despite its oversimplification, the UFM is valuable in predicting structural behaviors.

Expanding beyond the scope of CE minimization, \cite{graf2021dissecting} also used the UFM to determine whether optimizing with the supervised-contrastive loss (SCL)  results in any alterations to the geometric structure of the learned embeddings.\footnote{SCL is designed to train only embeddings and is an extension of the well-known contrastive loss used for unsupervised learning, adapted to supervised datasets \citep{khosla2020supervised}; see Equation \eqref{eq:SCL_full}.}
%

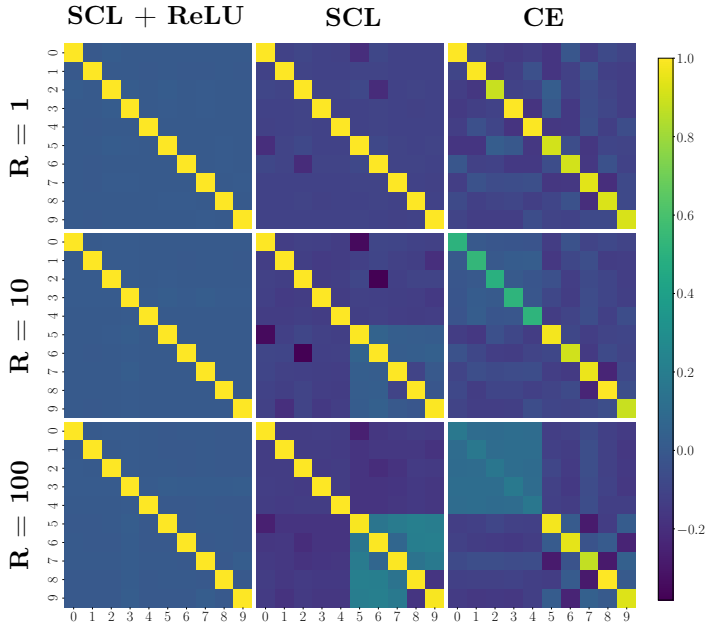
\begin{figure}
     \centering
            \begin{tikzpicture}
                \node at (0.0,0.0) 
                {\includegraphics[width=0.55\textwidth]{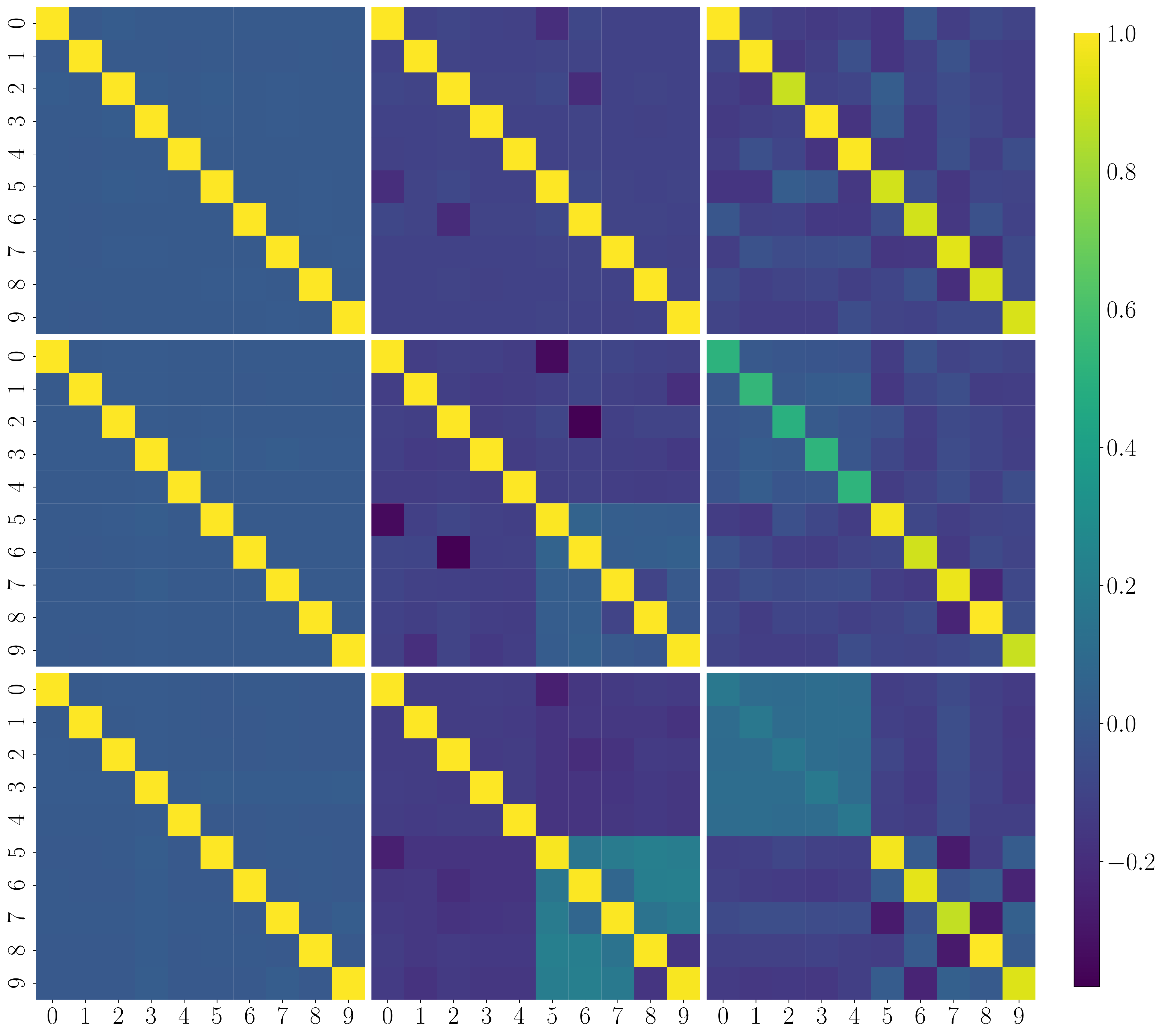}};
                 \node at (-2.9,4.2) [scale=0.9]{\textbf{SCL + ReLU}};
                \node at (-0.3,4.2) [scale=0.9]{\textbf{SCL}};
                \node at (2.2,4.2) [scale=0.9]{\textbf{CE}};
                \node at (-4.7,2.6)  [scale=0.9, rotate=90]{\textbf{R = 1}};
                \node at (-4.7,0.1)  [scale=0.9, rotate=90]{\textbf{R = 10}};
                \node at (-4.7,-2.4)  [scale=0.9, rotate=90]{\textbf{R = 100}};
            \end{tikzpicture}
    \captionsetup{width=0.95\linewidth}
    \caption{Gram matrices, $\G_\M/\max_{i,j}|\G_\M[i,j]|$, of class-means of learned feature embeddings at last epoch of training, with ResNet-18 on MNIST. \textbf{SCL+ReLU:} the mean feature embeddings for different classes are mutually orthogonal, forming an OF, regardless of imbalance (imbalance ratio $R=1,10,100$). This invariance does not hold in the absence of ReLU for \textbf{SCL}. Further, \textbf{CE} feature geometry is also sensitive to imbalance. The label distribution is STEP imbalanced, with the first five classes as majorities and the rest as minorities. See Sec.\ref{sec:exp_fig1_detail} in the appendix for more information.}
    \label{fig:heatmap}
\end{figure}

%
%
\vspace{-4pt}
They proved for balanced datasets that the global solution of $\min_{\h_i\in\R^d}\Lc_{\text{SCL}}\big(\{\h_i\}_{i\in[n]}\big)$ remains a simplex ETF, suggesting that CE and SCL find the same embedding geometries. In Fig.~\ref{fig:heatmap}, we investigate the geometry of class-means in presence of class-imbalance. The prediction in \cite{graf2021dissecting} only applies to balanced data (SCL with $R=1$ in Fig.~\ref{fig:heatmap}) and no prior work has explicitly characterized the geometry of SCL {with} class imbalances. Observing the middle column of the figure, note that the geometry of embeddings changes drastically\footnote{Analogous finding is reported by \cite{zhu2022balanced}, although not visualized as done here.} as the imbalance ratio $R$ increases. This behavior of SCL is consistent with the CE embeddings geometry, which, as discussed previously,  also changes with the imbalance; see last column of Fig.~\ref{fig:heatmap}.

In this paper, we arrive at a surprising finding: simply introducing a ReLU activation at the final layer restores the symmetry in SCL-trained representations. This phenomenon is clearly illustrated in the first column of Fig.~\ref{fig:heatmap}. Below is a detailed description of our contributions.

\subsection{Summary of contributions}

We conduct an in-depth examination of the geometry of training representations (embeddings) of SCL, particularly in relation to varying levels of dataset imbalances. Our study leads us to identify and capture, both analytically and empirically, and for the first time, the impact on the representation geometry of: (i) a straightforward architectural modification, specifically the incorporation of a ReLU activation at the final layer, and (ii) the batch-selection strategy.


\noindent\textbf{Impact of ReLU.}~We find that the addition of a ReLU activation to the last-layer of the architecture results in an orthogonal frame (OF) geometry. That is, a  geometry structure composed of class-mean embedding vectors that have equal norms and are mutually orthogonal to each other. A preliminary experimental validation of this finding is illustrated in Fig.~\ref{fig:heatmap}. The experiments correspond to STEP-imbalanced MNIST data with five majority/minority classes and imbalance ratio $R=1,10, 100$. Each heatmap represents the pairwise inner products between the class-means of learned feature embeddings. 
The figure (first column) reveals that with the addition of ReLU at the last layer, SCL  learns orthogonal features \emph{regardless of class imbalance.} This is in stark contrast to the varying geometries of vanilla SCL (optimized commonly without ReLU) and CE loss.
Extensive additional experiments for Long-tailed (LT) distributions, other datasets and architectures, are presented in Sec.~\ref{sec:Deepnet_SCLGeom_exp} and also Sec.~\ref{app:exp_appex} in the appendix. We also present experimental findings that confirm ReLU symmetrises the geometry without compromising test accuracy.

\begin{figure}[t!]
    \hspace*{-16pt}
    \begin{subfigure}[b]{\textwidth}
        \centering
        \begin{tikzpicture}
            \node at (0.0,0) 
            {\includegraphics[width=0.85\textwidth]{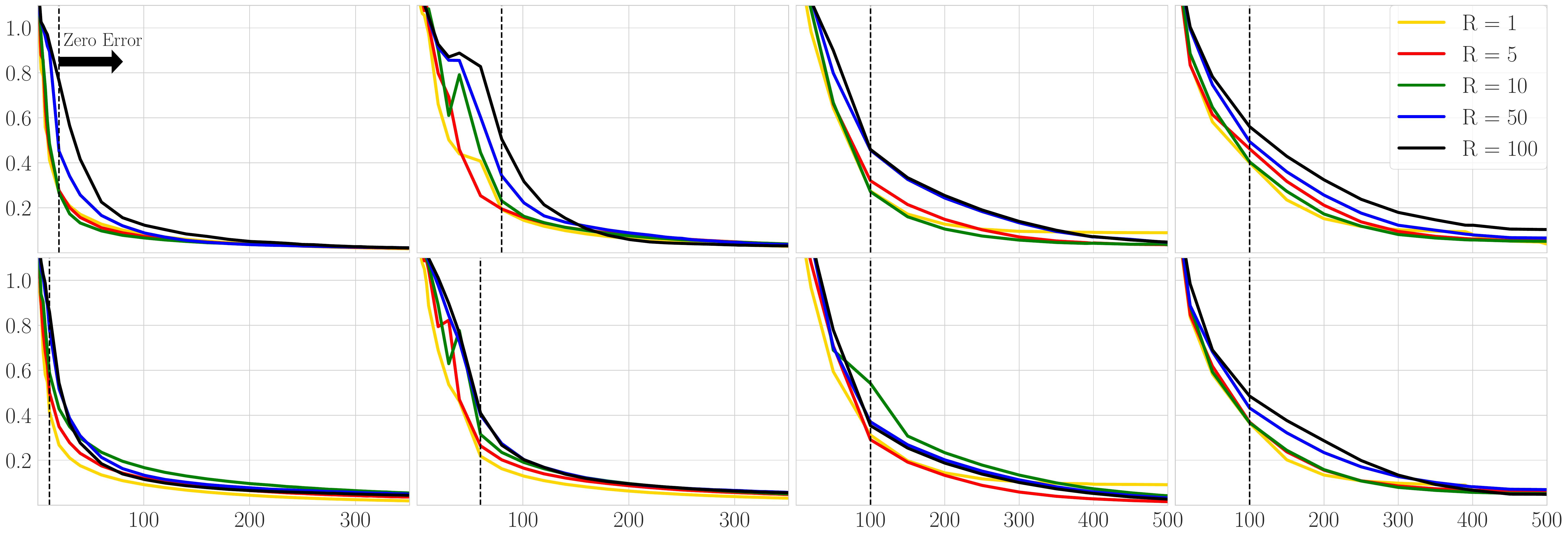}};
            \node at (-4.7,2.6) [scale=0.9]{\textbf{MNIST}};
            \node at (-1.4,2.6) [scale=0.9]{\textbf{CIFAR10}};
            \node at (1.8,2.6) [scale=0.9]{\textbf{CIFAR100}};
            \node at (5,2.6) [scale=0.9]{\textbf{Tiny ImageNet}};
            \node at (-7.0,1.2)  [scale=0.9, rotate=90]{\textbf{Step}};
            \node at (-7.0,-1)  [scale=0.9, rotate=90]{\textbf{Long-tail}};
            \node at (0.0,-2.6) [scale=0.9]{\textbf{Epoch}};
        \end{tikzpicture}
    \end{subfigure}
    \captionsetup{width= 0.95\linewidth}
    \caption
    {
    Distance of learned embeddings to the OF geometry as measured by $\|\frac{\G_\M}{\| \G_\M\|_F} - \frac{\Id_{k}}{\|\Id_{k}\|_F}\|_F$ where $\G_\M=\M^\top\M$ is the Gram matrix of class-mean embeddings. For MNIST and CIFAR10 we use ResNet-18 and for the more complex CIFAR100 and Tiny-ImageNet, we use ResNet-34 and train using batch-binding (see Sec.~\ref{sec:batching}). Regardless of the imbalance level $R$, the embeddings consistently converge to the OF geometry.
    %
    %
    }
    \label{Fig:Deepnet_SCL_Geom_GMu}
\end{figure}
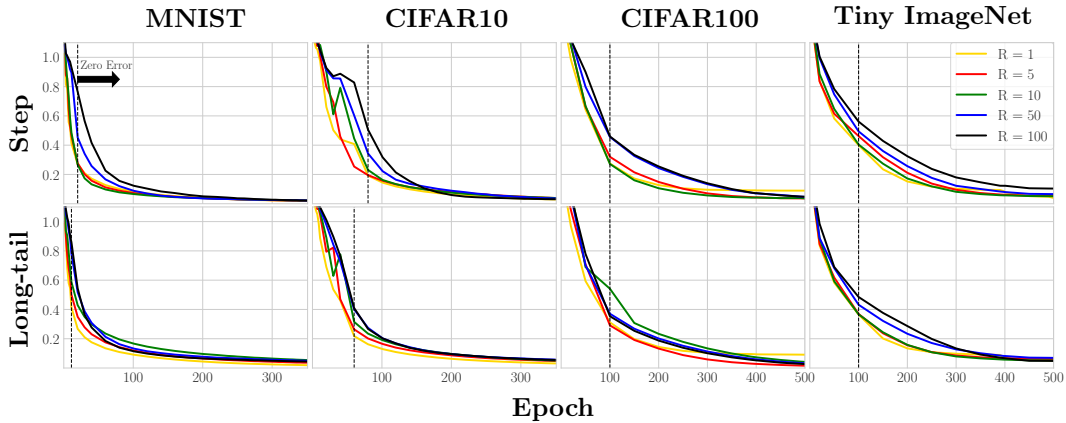

We support this finding by theoretically  investigating the global minimizers of an extended version, denoted as \ref{eq:SCL_UFM_full}, of the original unconstrained-features model (UFM). This refinement accounts for the presence of ReLU activations which we impose on the last-layer by minimizing SCL over the non-negative orthant. Concretely, we show that all global solutions of $\min_{\h_i\geq 0}\Lc_{\text{SCL}}\big(\{\h_i\}_{i\in[n]}\big)$ satisfy NC and the corresponding class-mean features form an OF. This explains the  convergence of feature geometry to an OF as consistently observed in our experiments; for example, see Fig. \ref{Fig:Deepnet_SCL_Geom_GMu}. 

\noindent\textbf{Batching and its implications.}~Furthermore, we identify the crucial impact of batch selection strategies  during SCL training in shaping the learned embedding geometry. Concretely, by analyzing the \ref{eq:SCL_UFM_full}, we establish straightforward criteria for the batching scheme, ensuring that any global minimizer of \ref{eq:SCL_UFM_full} forms an OF. These conditions explain the substantial degradation in convergence towards an OF when employing a fixed batch partition instead of randomly reshuffling the batches at each training epoch. Intriguingly, we demonstrate that any arbitrary batching partition can be transformed into one that fulfills our theoretical conditions by performing so-called \textit{batch-binding}. Through extensive experiments in Sec.~\ref{Sec:Binding_examples} and in the appendix, we show that batch-binding consistently accelerates the convergence of the learned geometry towards an OF and even improves convergence to ETF geometry under balanced training in the absence of ReLU.

\vspace{-0.25cm}
\section{Setup}\label{sec:setup}
\vspace{-0.2cm}
%
%
%


\noindent\textbf{Notation.}~
For a positive integer $n$, $[n] := \{1,2,3,...,n\}$. For matrix $\Vb\in\R^{m\times n}$, $\Vb[i,j]$ denotes its $(i,j)$-th entry, $\vb_j$ denotes the $j$-th {column}, $\Vb^\top$ its transpose. We denote $\|\Vb\|_F$ the Frobenius norm of $\Vb$. We use $\Vb \propto \mathbf{X}$ whenever the two matrices are equal up to a scalar constant. We use $\Vb \geq 0$ to denote the entry-wise non-negativity, i.e., $\Vb[i,j] \geq 0, \forall i \in [m],j \in [n]$. Finally, $\Id_m$ denotes the identity matrix of size $m$.

Consider $k$-class classification and training set $S=\{(\x_i,y_i)\}_{i = 1}^{n}$ where $\x_i\in\R^d,\,y_i\in[k]$ represent  data samples and  corresponding class labels. We use $n_c$ to denote the number of examples in class $c\in[k]$, and $\h_\thetab(\cdot)$ to denote the 
 last-layer feature-embeddings of a deep-net parameterized by $\thetab$. We compute the SCL on a given batch $B\subseteq[n]$ of training examples as follows,
     \begin{align}\label{eq:scl_main}
     \Lc_B(\thetab) := \sum_{i\in B}\frac{1}{n_{B,y_i}-1}\sum_{\substack{j\in B \\ y_j = y_i, j\neq i}}\log\bigg(1+\sum_{\ell \neq i,j}\exp{\Big(\frac{1}{\tau}\big({\h_\thetab(\x_i)^\top\h_\thetab(\x_\ell) - \h_\thetab(\x_i)^\top\h_\thetab(\x_j)}\big)\Big)}\bigg),
 \end{align}
where $\tau$ is a positive scalar temperature hyper-parameter,
{and $n_{B,y_i}$ is the number of examples in batch $B$ belonging to class $c= y_i$. To train a deep-net, we minimize SCL \eqref{eq:scl_main} over the network parameters $\thetab$ on a set of batches $\mathcal{B}$ chosen from the training set. As introduced by \citet{khosla2020supervised}, SCL requires normalized features, so we assume $\norm{\h_{\thetab}(\x_i)} = 1$}.
Furthermore, we assume that $d\geq k$ and $n_c, n_{B,c}\geq2, c\in[k]$.\footnote{When using SCL, it is common practice to add an augmented version of the datapoints in each batch to itself \citep{khosla2020supervised,graf2021dissecting}. This practice, referred to as \textit{batch duplication}, helps ensure that each datapoint in a batch has at least one other example in the same class to compare to during training.}
{We let $\HT_{d \times n} = \begin{bmatrix} \hT_1, \cdots, \hT_n \end{bmatrix}$ denote an embeddings matrix where each column corresponds to one of the examples in the training set, {i.e., $\h_i:=\h_\thetab(\x_i)$.  The \emph{embeddings geometry} or so-called  \emph{implicit geometry}\footnote{
The specific terminology is adopted from \cite{behnia2023implicit}.} refers to the norms and pairwise-angles of these vectors $\h_i, i\in[n]$. Note these quantities correspond exactly to the entries of the
Gram matrix $\Hb^\top\Hb$.} To characterize the geometry, we need the following definitions.} 


\begin{definition}[Neural Collapse (NC)]\label{def:nc}
    NC occurs if $\h_i = \h_j, \ \forall \ i,j: y_i = y_j$. 
\end{definition} 
\begin{definition}[$k$-Orthogonal Frame ($k$-OF)]\label{def:of}
    We say that $k$ vectors $\Vb = [\vb_1,\cdots,\vb_k] \in \R^{d\times k}$ form a $k$-OF {if $\Vb^\top\Vb \propto \Id_k$, i.e., for each pair of $(i,j)\in[k],$ $\|\vb_i\| = \|\vb_j\|$ and $\vb_i^\top\vb_j = 0$.} 
\end{definition}
{When the within-class variation of embeddings is negligible, i.e., $\Hb$ satisfies NC, it suffices to focus on the class-mean embeddings ${\mub}_c = \frac{1}{n_c}\sum_{i: y_i = c} \h_i, c\in[k]$ and the respective matrices $\MT_{d\times k} =  \begin{bmatrix} {\mub}_1, \cdots, {\mub}_k \end{bmatrix}$ and $\G_\M = \M^\top\M$ instead of $\Hb$ and $\Hb^\top\Hb.$ With these, we can formally define the OF geometry for embeddings.}

\begin{definition}[OF geometry] \label{def:ncof}
    We say that a feature-embedding matrix $\Hb$ follows an OF geometry if it satisfies NC and the class-means form a $k$-OF, i.e., $\G_\M\propto\Id_k$.
\end{definition}

\section{{SCL with ReLU learns OF geometries:} Empirical findings}\label{sec:Deepnet_SCLGeom_exp}
\noindent\textbf{Experimental setup.}~ Following the setup of \citet{graf2021dissecting}, our models consist of a backbone (ResNet, DenseNet, etc) with a normalizing layer to output features with unit norm. For all geometric convergence results, we apply a ReLU on output features before normalization. We employ {Stochastic Gradient Descent} (SGD) with a learning rate of $0.1$, momentum set to $0.9$ and with no weight decay.  Furthermore, as in \citet{graf2021dissecting,khosla2020supervised}, we set the temperature parameter $\tau = 0.1$ as \citet{khosla2020supervised} have found that it yields optimal performance. In addition, we empirically find that convergence to OF is not highly dependent on $\tau$ for values near $0.1$, yet the specific choice affects the speed of convergence. Code is available \href{https://github.com/valavakilian/SCL_Geometry_And_Batching}{here.}

We study the behavior of models trained with SCL under, {1)} $R$-\textit{STEP} imbalance having $k/2$ majority classes with $n_\maj$ examples per class and $k/2$ minority classes with $n_{\min} = \nicefrac{n_{\maj}}{R}$ examples per class, and {2)} $R$-\textit{Long-tailed} (LT) imbalance where the number of training datapoints exponentially decreases across classes such that $n_c = n_1 R^{-(c-1/k-1)}$, for $c \in [k]$. Regardless of the imbalance ratio $R$, we ensure that $n_c \geq 2$ by adding a vertically flipped version of each image as a method of batch duplication. Unless stated, we do not perform any additional data augmentation on the datasets, and use a batch size of $1024$ with random reshuffling. 

Finally, when studying the impact of ReLU on generalization, we use a Nearest Center Classifier (NCC) on the output features to evaluate model performance. For such experiments, following the setup of \cite{khosla2020supervised}, we employ a projection head by adding a 2 layer non-linear MLP (with or without ReLU at the last layer) and compare the test accuracy under difference imbalance ratios.

\begin{wraptable}[13]{r}{0.55\textwidth}\label{tab:acc}
\vspace*{-5pt}
    \centering
    \resizebox{0.55\columnwidth}{!}{%
    \begin{tabular}{|l|l|l|}
        \cline{1-3}
        \multicolumn{1}{|c}{Imbalance Ratio (R)} & \multicolumn{1}{|c}{w/o ReLU} & \multicolumn{1}{|c|}{w/ ReLU} \\ \hline
        1 & \multicolumn{1}{c}{72.17 $\pm$ 0.23}  & \multicolumn{1}{|c|}{72.32 $\pm$ 0.60} \\ \hline
        10 (Step) & 56.58 $\pm$ 0.50 & 58.16 $\pm$ 0.76\\ \hline
        10 (LT) & 57.10 $\pm$ 1.04 & 57.88 $\pm$ 1.05 \\ \hline
        100 (Step) & 43.49 $\pm$ 0.30 & 43.80 $\pm$ 0.25 \\ \hline
        100 (LT) & 37.19 $\pm$ 2.50 & 39.71 $\pm$ 0.09 \\ \hline
    \end{tabular}
    }
    \captionsetup{width=\linewidth}
    \caption{Test accuracy comparison for a ResNet-18 trained using SCL on CIFAR100 with and without ReLU after the projection head. We use NCC for classification. Note that the addition of ReLU does not compromise the accuracy. For details, see Sec.~\ref{sec:generalization} in the appendix.} 
    \label{tab:cifar100_main}
\end{wraptable}
\noindent\textbf{Metrics.}~Since in all experiments we observe the within-class variations of the last-layer embeddings ($\h_i$) become negligible towards the end of training (see NC plots in Fig.~\ref{fig:Deepnet_SCL_Geom_NC} in the appendix), we focus here on the geometry of class-mean embeddings $\M$. 
Thus, to measure the distance of learned embedding to the OF geometry, we compute the distance metric $\Delta_{\G_\M} := \|\frac{\G_\M}{\|\G_\M\|_F} - \frac{\Id_{k}}{\|\Id_{k}\|_F}\|_F$.

\noindent\textbf{Observations on Geometry.}~In Fig.~\ref{Fig:Deepnet_SCL_Geom_GMu}, we plot the distance $\Delta_{\G_\M}$ between the learned feature-embeddings and the OF geometry as training progresses. We consistently observe that the learned features during training converge to the OF geometry, irrespective of the imbalance level $R$ of the training set and the imbalance pattern (STEP or LT). This suggests that the feature geometry learned by SCL
is invariant to the training label distribution. 

Interestingly, the invariance that is revealed by our study is distinct for SCL with ReLU as opposed to the case of SCL without ReLU, and that of CE and MSE loss. CE and MSE are known to be sensitive to imbalances \citep{seli, fang2021exploring, liu2023inducing, behnia2023implicit, dang2023neural}. Additionally, for different values of $R$, there is no significant difference between the speed of convergence, unlike the CE loss \citep{seli}, where it is empirically observed that the rate worsens with larger imbalance.

\noindent\textbf{Generalization.} To ensure that ReLU does not yield any adverse effects on test accuracy, we trained ResNet-18 (with a projection head) with and without ReLU on CIFAR10, CIFAR100, Tiny Imagenet and evaluated the balanced test accuracy when training under label imbalance. Following \citep{zhu2022balanced,khosla2020supervised,chen2020simple}, we consider the features before a projection head for evaluating the test accuracy of the models. Our results, e.g. in Tab.~\ref{tab:cifar100_main}, indicate that the addition of ReLU does \emph{not} compromise test accuracy. For a detailed discussion on the results see Sec.~\ref{sec:generalization} in the appendix.

\section{Theoretical justification: SCL 
with non-negativity constraints}\label{sec:theo_results} 
In this section, we analytically justify the convergence of the embeddings learned by SCL to the OF geometry. For our theoretical analysis, we use the Unconstrained Features Model (UFM) \citep{mixon2020neural,fang2021exploring,graf2021dissecting,ULPM,lu2020neural,tirer2022extended}, where we treat the last-layer features $\Hb$ as free variables, removing the dependence to the network parameters $\thetab$. However, we refine the UFM, and consider the SCL minimization over the non-negative orthant to accommodate for the presence of ReLU activations in the last layer.\footnote{\cite{tirer2022extended} (and several follow ups, e.g. \cite{sukenik2023deep}) has also studied incorporating the ReLU activation in the UFM, albeit focusing on MSE loss.} Following \citet{khosla2020supervised}, we further constrain the embeddings $\h_i$ to be normalized. We prove, irrespective of the labels distribution, that the global optimizers form an OF. Formally, consider the following refined UFM for SCL, which we call {$\text{UFM}_+$} for convenience:
%
\begin{align}\label{eq:SCL_UFM_full}
    \widehat{\Hb} \in \arg\min_{\Hb} \Lc_{\text{SCL}}
    (\Hb)~~\text{{subj. to}}~~\Hb \geq 0\,\text{ and }\|\h_i\|^2 = 1,\, \forall i\in[n].
    \tag{$\text{UFM}_+$}
\end{align}
Recall $\Hb\geq 0$ denotes entry-wise non-negativity. We begin our analysis by considering the full-batch loss in Sec.~\ref{sec:full_batch_theory}, and extend our results to the SCL minimized on mini-batches in Sec.~\ref{sec:mini_batch_theory}. We provide specific conditions that mini-batches must satisfy in order to have the same global optimum as the full-batch version. 

\subsection{Full-batch SCL}\label{sec:full_batch_theory}
Here, we focus on the full-batch SCL, where the entire training set is treated as a single batch. Specifically, we consider \ref{eq:SCL_UFM_full} 
with $\Lc_{\text{SCL}}
(\Hb)$ being the full-batch SCL defined as,
\begin{align}\label{eq:SCL_full}
    \Lc_\text{full}(\Hb):=\sum_{i\in[n]}\frac{1}{n_{y_i}-1}\sum_{j \neq i, y_j = y_i}\log\Big(\sum_{\ell \neq i}\exp{\big(\h_i^\top\h_\ell - \h_i^\top\h_j\big)}\Big).
\end{align}
Thm.~\ref{thm:scl_full} specifies the optimal cost and optimizers of \ref{eq:SCL_UFM_full} with the full-batch SCL \eqref{eq:SCL_full}.

\begin{theorem}[Full-batch SCL minimizers]\label{thm:scl_full} Let $d\geq k$.
For any $\Hb$ feasible in \ref{eq:SCL_UFM_full}, it holds,
  \begin{align}\label{eq:scl_full_lb}
      \Lc_\text{\emph{full}}(\Hb) \geq \sum_{c \in [k]}n_c\log\left(n_c-1+(n-n_c)e^{-1}\right).
  \end{align}
  Moreover, equality is achieved if and only if $\Hb$ satisfies NC and the class-means form a k-OF.
\end{theorem}

We defer the proof details to Sec.~\ref{app:proof_details} in the appendix. The bound relies on successive uses of Jensen's inequality and the fact that for feasible $\Hb$, each pair of training samples $i,j\in[n]$ satisfies $0\leq\h_i^\top\h_j\leq 1$. We complete the proof by verifying that equality is attained only if $\h_i^\top\h_j=1$ when $y_i=y_j$ and $\h_i^\top\h_j=0$ when $y_i\neq y_j$. Rather, to achieve the optimal cost, features with similar labels must align (NC) and the class-mean features must form an OF.

\begin{wrapfigure}[13]{r}{0.35\textwidth} 
     \centering
            \begin{tikzpicture}
                \node at (0.0,0.0) 
                {\includegraphics[width=0.28\textwidth]{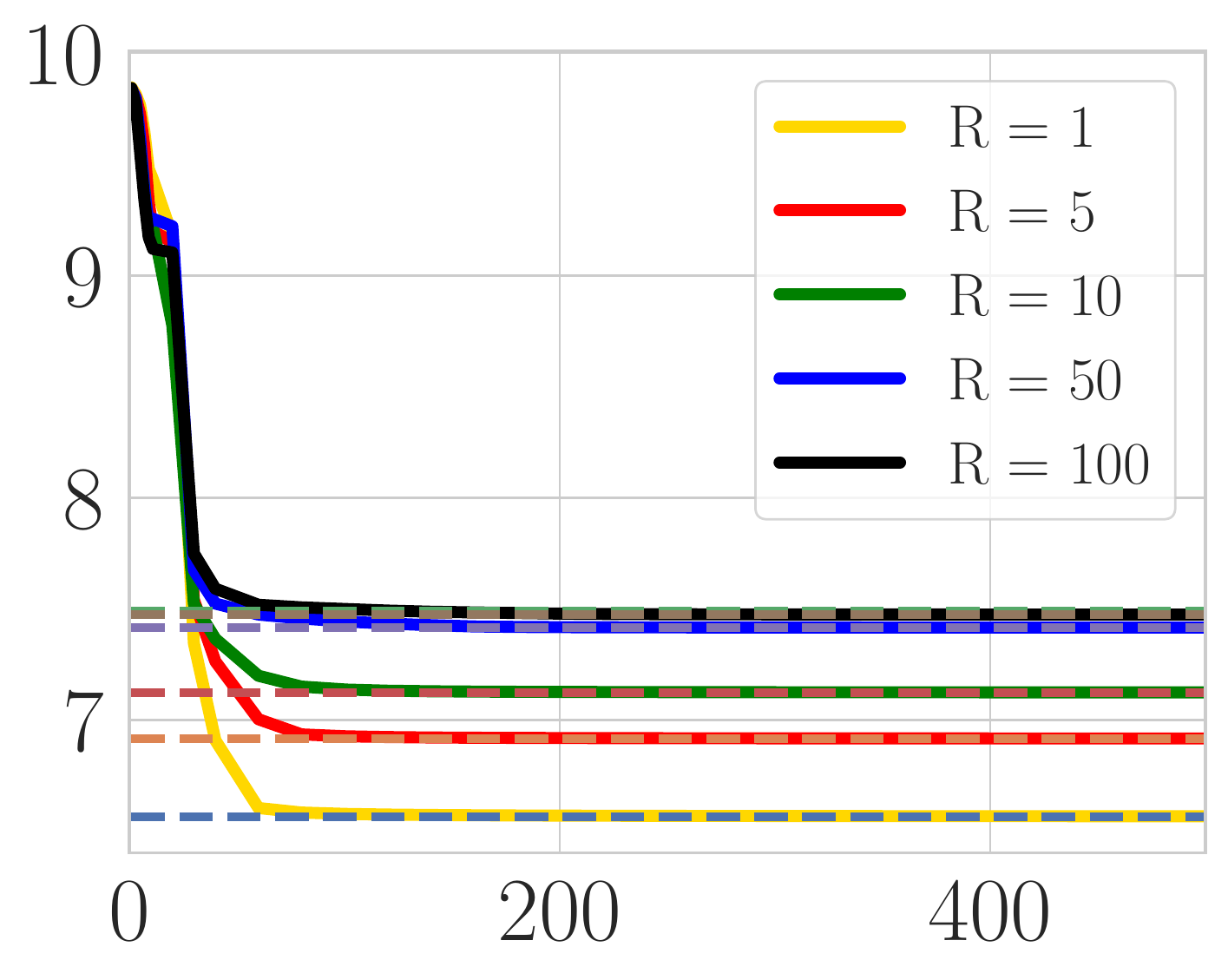}};
                 \node at (0.1,-2.1) [scale=0.7]{\textbf{Epoch}};
    			\node at (-2.4,0) [scale=0.9, rotate = 90]{\textbf{Loss}};
            \end{tikzpicture}
    \captionsetup{width=0.9\linewidth}
    \caption{Full-batch SCL converges to the lower bound (dashed lines) in Thm.~\ref{thm:scl_full}.}
    \vspace{0.03in}
    \label{fig:loss_convergence}
\end{wrapfigure}

Thm.~\ref{thm:scl_full} shows that any optimal embedding geometry learned by the full-batch SCL with ReLU constraints uniquely follows the OF geometry (Defn. \ref{def:ncof}) and the conclusion is independent of the training label distribution.
We have already seen in Sec.~\ref{sec:Deepnet_SCLGeom_exp}, that  this conclusion is empirically verified by deep-net experiments.
 To further verify the lower bound on the cost of the SCL loss given by the theorem, we compare it in Fig.~\ref{fig:loss_convergence} with the empirical loss of a  ResNet-18 model trained with full-batch SCL \eqref{eq:SCL_full} on $R$-STEP MNIST dataset and $n=1000$ total examples. Note the remarkable convergence of the loss to the lower bound (dashed horizontal lines) as training progresses.

\subsection{Mini-batch SCL}\label{sec:mini_batch_theory}
Thm.~\ref{thm:scl_full} shows that minimizing SCL over the training set as a single batch recovers an OF in the feature space, regardless of the training label distribution. However, in practice, SCL optimization is performed over batches chosen from training set \citep{khosla2020supervised, graf2021dissecting}. Specifically, we have a set of batches $\mathcal{B}$ and we compute the loss on each batch $B\in\mathcal{B}$ as in \eqref{eq:scl_main}. While in the full-batch SCL \eqref{eq:SCL_full}, all pairs of training samples interact with each other, in the mini-batch version, two samples $(i,j)$ interact only if there exists a batch $B\in\mathcal{B}$ such that $i,j\in B$. A natural question is whether the mini-batch construction impacts the embeddings learned. In this section, we study the role of batching on the embeddings geometry more closely.

Similar to the previous section, we consider \ref{eq:SCL_UFM_full}, where we directly optimize the SCL over embeddings $\Hb$. However, this time we study the mini-batch SCL defined as follows,
 \begin{align}\label{eq:scl_batch}
     \Lc_{\text{batch}}(\Hb):=\sum_{B \in \mathcal{B}}\sum_{i\in B}\frac{1}{n_{B,y_i}-1}\sum_{\substack{j\in B\\j \neq i, y_j = y_i}}\log\bigg(\sum_{\substack{\ell \in B \\ \ell \neq i}}\exp{\left(\h_i^\top\h_\ell - \h_i^\top\h_j\right)}\bigg),
 \end{align}
 where recall $n_{B,c} = |\{i:i\in B,y_i=c\}|$. Thm.~\ref{thm:scl_mini} is an extension of Thm.~\ref{thm:scl_full} for the mini-batch SCL \eqref{eq:scl_batch}. The proof proceeds similarly to that of Thm.~\ref{thm:scl_full}, where the loss over each batch can be independently bounded from below. Interestingly, due to the orthogonality of the optimal embeddings in every batch, the lower bound for each batch can be achieved by the common minimizer. The detailed proof is deferred to Sec.~\ref{app:proof_thm2} in the appendix.
 %
\begin{theorem}[Mini-batch SCL minimizers]\label{thm:scl_mini}
    Let $d\geq k$ and $\mathcal{B}$ be an arbitrary set of batches of examples. For any feasible $\Hb$ in \ref{eq:SCL_UFM_full}, the following lower bound holds,    
      \begin{align}\label{eq:scl_mini_lb}
      \Lc_\text{\emph{batch}}(\Hb) \geq \sum_{B \in \mathcal{B}}\sum_{c \in [k]}n_{B,c}\log\left(n_{B,c}-1+(n_{B}-n_{B,c})e^{-1}\right).
  \end{align}
  Equality holds if and only if for every $B\in\mathcal{B}$ and every pair of samples $i,j\in B$, $\h_i^\top\h_j=0$ if $y_i\neq y_j$ and $\h_i=\h_j$ if $y_i=y_j$.
\end{theorem}

We remark that the only previous study of SCL geometry with batches by \cite{graf2021dissecting} requires significantly more restrictive conditions on the batch set $\Bc$ to guarantee a common geometry among all batches $B\in\Bc$. Specifically, \cite{graf2021dissecting} assumes a batch set that includes all possible combinations of examples. Instead, thanks to the addition of ReLU, our requirements are significantly relaxed. This is detailed in the following discussion. \\

\begin{figure*}
        \centering
        \begin{subfigure}{0.99\textwidth}
            \centering
            \begin{tikzpicture}
                \node at (0.0,0) 
                {\includegraphics[clip, trim=1cm 13cm 1cm 13.3cm, width = \textwidth]{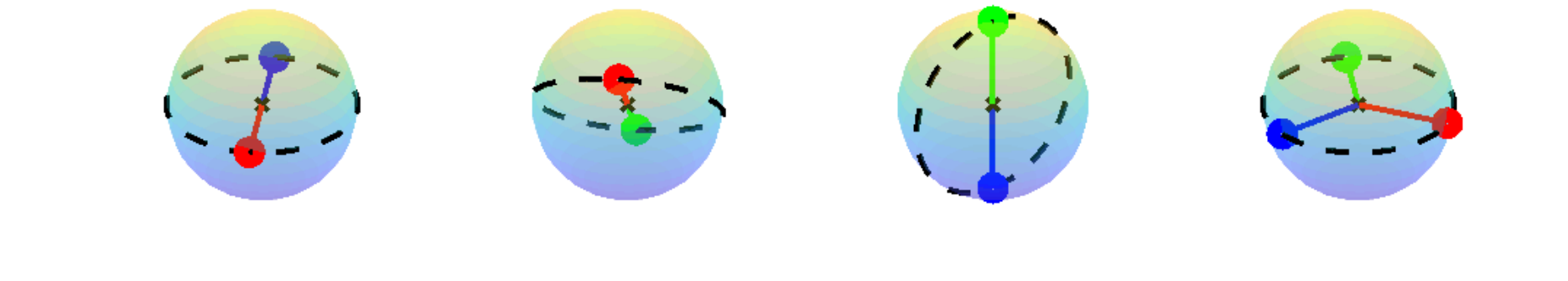}};
                \node at (0.0,-1.6) [scale=0.9]{\textbf{(a)} UFM optimal embeddings};
                \node at (-5.25,-1) [scale=0.9]{\textbf{(i)}};
                \node at (-1.55,-1) [scale=0.9]{\textbf{(ii)}};
                \node at (2.1,-1) [scale=0.9]{\textbf{(iii)}};
                \node at (5.75,-1) [scale=0.9]{\textbf{(iv)}};
            \end{tikzpicture}
        \end{subfigure}
        \hfill
        \begin{subfigure}{0.99\textwidth}
            \centering
            \begin{tikzpicture}
                \node at (0.0,0) 
                {\includegraphics[clip, trim=1cm 12cm 1cm 12.3cm, width = \textwidth]{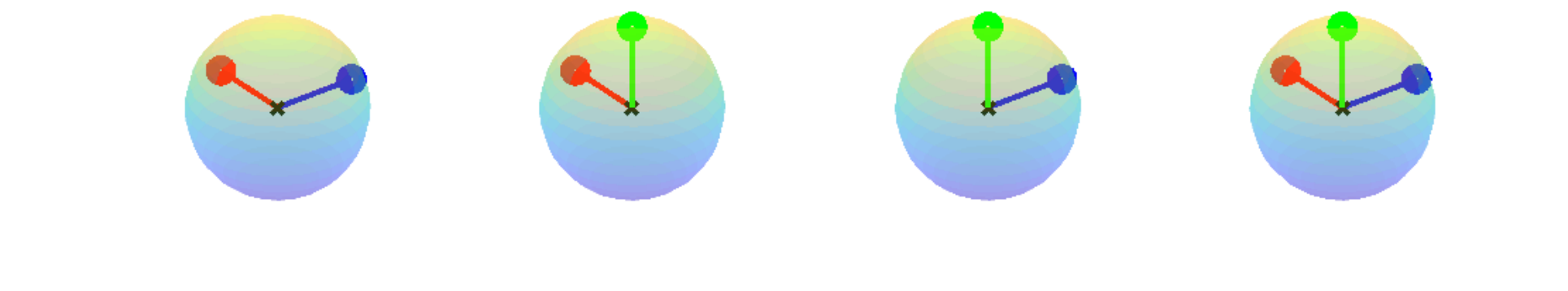}};
                \node at (0.0,-1.6) [scale=0.9]{\textbf{(b)} UFM$_+$ optimal embeddings};
                \node at (-5.25,-1) [scale=0.9]{\textbf{(i)}};
                \node at (-1.55,-1) [scale=0.9]{\textbf{(ii)}};
                \node at (2.1,-1) [scale=0.9]{\textbf{(iii)}};
                \node at (5.75,-1) [scale=0.9]{\textbf{(iv)}};
            \end{tikzpicture}
        \end{subfigure}
        \hfill
        \captionsetup{width= \linewidth}
        \caption[remark]
        {Rem.~\ref{rem:compare_graf_app} visualized, \textbf{(a)(i,ii,iii)} indicate the antipodal structure of the optimal embeddings under UFM in each of the 3 mini-batches respectively, whereas the overall optimal geometry is an ETF \textbf{(a)(iv)}. This contrasts the optimal embeddings under UFM$_+$ where each mini-batch \textbf{(b)(i,ii,iii)} is consistent with the overall optima \textbf{(b)(iv)}.}
        \label{fig:remark} 
    \end{figure*}
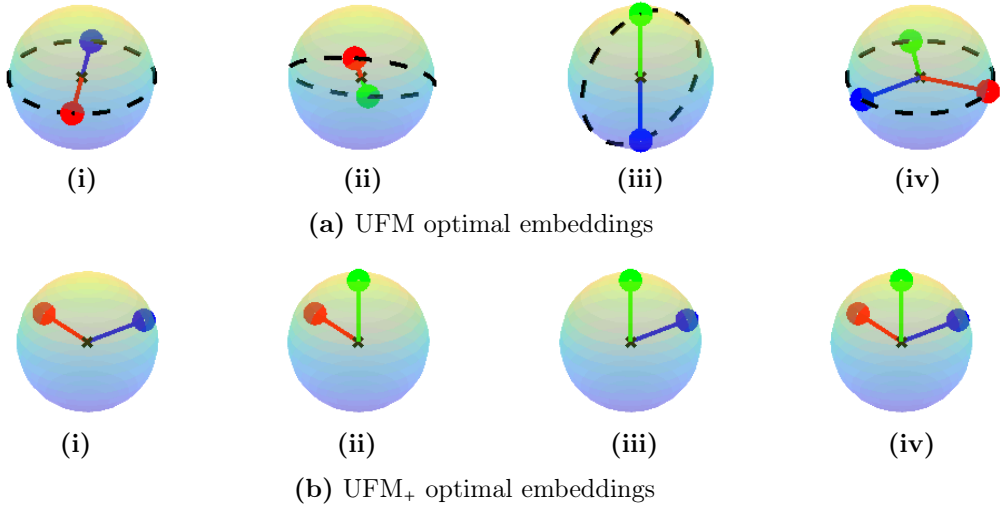


\begin{remark}[Comparison to analysis of \citet{graf2021dissecting}]\label{rem:compare_graf_app} {We elaborate on the comparison to \citet{graf2021dissecting} with an illustrative example below. When optimizing the loss decomposed over a set of mini-batches, it is crucial to consider whether the individual batch minimizers are also overall optimal solutions. }
Consider the following setting: $k= d = 3, y_1 = 1, y_2 = 2, y_3 = 3, B_1 = \{1,2\}, B_2 = \{1,3\}, B_3 = \{2,3\}, \mathcal{B} = \{B_1, B_2, B_3\}$. Let us identify the optimal embeddings for the two scenarios: (i) with and (ii) without non-negativity constraints on the embedding coordinates.
\\
    \noindent\textbf{(i)~Without non-negativity (UFM)}, the optimal embedding configurations can be shown to be ETF with $2$ vectors for every batch, which is simply an antipodal structure. In other words, the batch-wise optimal solutions are $\h_1 = -\h_2, \h_1 = -\h_3$ and $\h_2 = -\h_3$, for the batches $B_1, B_2$ and $B_3$, respectively. However, the batch-wise optimal solutions are infeasible due to their contradictory nature. From \cite{graf2021dissecting}, it can be deduced that the overall optimal configuration is instead an ETF with $3$ vectors. This example underscores the difficulty in optimizing the loss over every batch separately in case of SCL without non-negativity. See (a) in Fig.~\ref{fig:remark}.
\\    
    \noindent\textbf{(ii) With non-negativity \eqref{eq:SCL_UFM_full}}, our results imply that the optimal embeddings form a $2$-OF for every batch, i.e., $\h_1 \perp \h_2, \h_1 \perp\h_3$ and $\h_2 \perp\h_3$, for the batches $B_1, B_2$ and $B_3$, respectively. The three conditions are compatible with each other and the fact that the overall optimal solution is a $3$-OF in $\R^3$. Therefore, we were able to break down the overall optimization into individual batches. See (b) in Fig.~\ref{fig:remark}.
\end{remark} 

\begin{wrapfigure}[15]{h}{0.35\textwidth}
    \centering \vspace{-0.5cm}\includegraphics[width=0.35\textwidth]{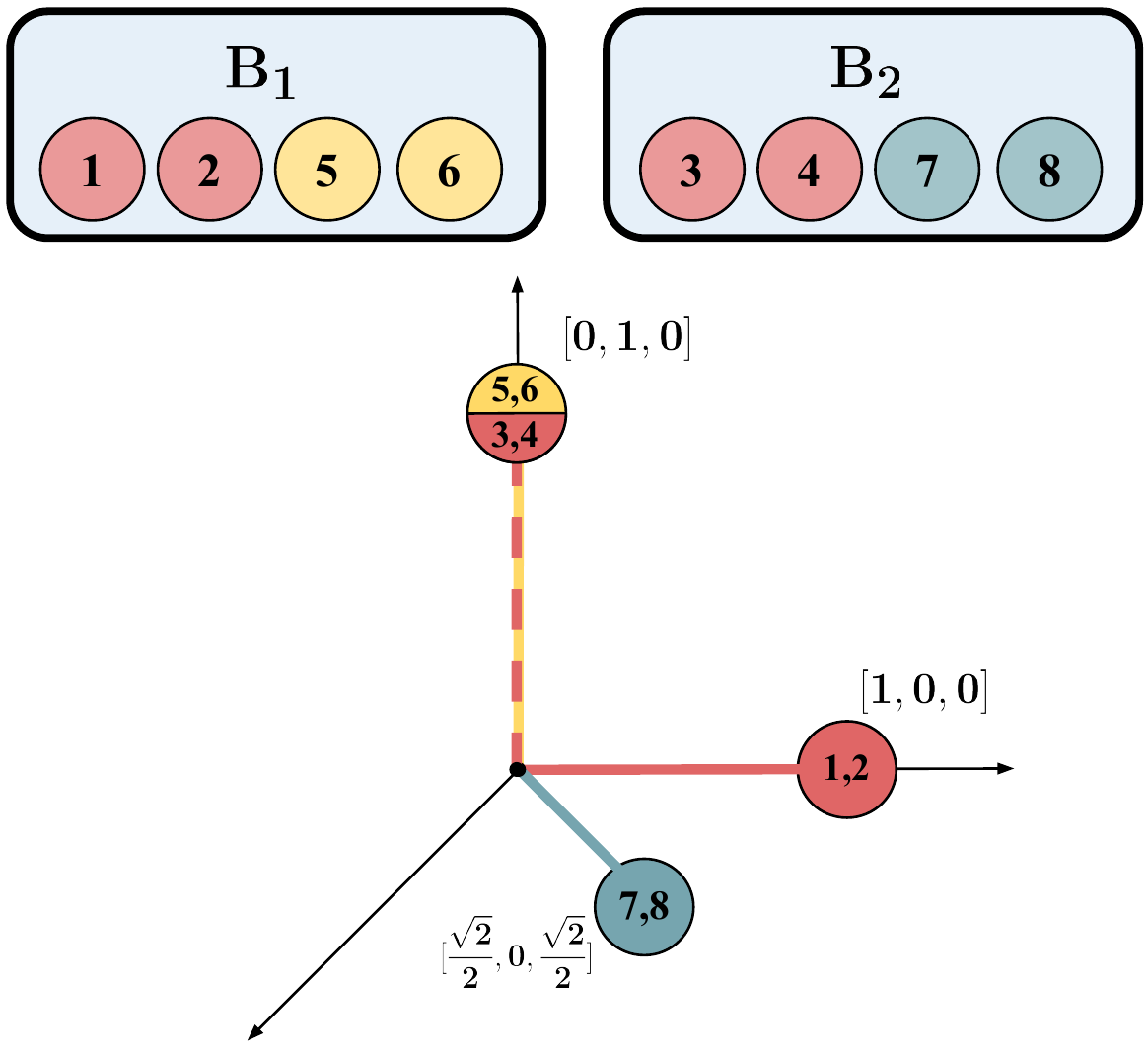} 
    \captionsetup{width=\linewidth}
    \caption{A non-OF geometry minimizing mini-batch SCL with $\mathcal{B}=\{B_1,B_2\}$. Samples with different labels are marked with different colors. 
    }
    \label{fig:counter_example}
\end{wrapfigure}
\noindent\textbf{Global minimizer geometry may not be unique.} It is easy to verify that any $\Hb$ following the OF geometry achieves the lower-bound of Thm.~\ref{thm:scl_mini} and it is also a global optimizer of the mini-batch SCL. However, depending on the choice of $\mathcal{B}$, the lower bound can possibly be attained by other optimal embeddings that do not necessarily satisfy NC or orthogonality. Hence the global optimizer may not have a unique geometry.

To highlight the importance of $\mathcal{B}$ in the characterization of the optimal geometry of \ref{eq:SCL_UFM_full} when using the batch-loss, consider the toy example in Fig.~\ref{fig:counter_example}. Suppose we have $k=3$ classes and we want to find the optimal normalized and non-negative features in $\R^3$ by minimizing \eqref{eq:scl_batch} for $\mathcal{B}=\{B_1, B_2\}$. Although the OF satisfies the global optimality condition in Thm.~\ref{thm:scl_mini}, the theorem requires milder conditions for the optimal $\widehat{\Hb}$: $\widehat{\h}_i$ and $\widehat{\h}_j$ need to be aligned ($y_i=y_j$) or orthogonal ($y_i\neq y_j$) only if they interact within one of the selected batches. Fig.~\ref{fig:counter_example} shows one such optimal geometry that does not satisfy either of NC or orthogonality conditions: Firstly, the samples $i=1$ and $i=3$ have significantly different features despite belonging to the same class. Second, ${\widehat{\h}}_{3}, \widehat{\h}_{4}$ align with $\widehat{\h}_{5}, \widehat{\h}_{6}$ although they have different labels, and  $\widehat{\h}_7, \widehat{\h}_8$ are not orthogonal to samples $\widehat{\h}_1, \widehat{\h}_2$. With this example, we are now ready to discuss the role of batching more formally in the next section.

\section{Batching matters} \label{sec:batching}

{Since the OF may not be the only solution for the \ref{eq:SCL_UFM_full} with mini-batch SCL \eqref{eq:scl_batch}}, here we study the role of mini-batches to avoid ambiguous solutions with different geometries. In Sec.~\ref{Sec:unique}, 
we identify necessary and sufficient conditions for a batching strategy to yield a unique global solution geometry. By this result, in Sec.~\ref{Sec:Binding_examples}, we propose a simple yet effective scheme that provably succeeds in improving the convergence to an OF.
\subsection{When is OF geometry the unique minimizer?}  \label{Sec:unique}
As discussed in Sec.~\ref{sec:mini_batch_theory} the uniqueness of the global minimizer's geometry when considering mini-batches depends on the interaction of samples in the loss function, {or, in other words, the choice of $\mathcal{B}$.}
Before specifying for which $\mathcal{B}$ the minimizer is unique, we need to define the Batch Interaction Graph{, a graph that captures the pairwise interactions of samples within the batches.}

    \begin{definition}[Batch Interaction Graph]\label{def:big}
        Consider an undirected graph $G = (V, E)$ where $V := [n]$. We define the {Batch Interaction Graph} for a given set of batches $\mathcal{B}$ as follows: vertices $u,v\in V$ are connected if and only if there exists a batch $B\in\mathcal{B}$ such that $u,v \in B$. Moreover, $G_c$ denotes the induced subgraph of $G$ with vertices $V_c=\{u:y_u=c\}$. 
    \end{definition}
%
    We state necessary and sufficient conditions on $\mathcal{B}$ for the minimizer of mini-batch SCL to be unique:
\begin{corollary}\label{cor:batch}
      Consider the Batch Interaction Graph $G$ corresponding to $\mathcal{B}$. The global optimizer geometry of \ref{eq:SCL_UFM_full} with mini-batch SCL \eqref{eq:scl_batch} is unique and forms an OF if and only if $G$ satisfies the following conditions: (1) For every class $c \in [k]$, $G_c$ is a connected graph. (2) For every pair $c_1,c_2\in[k]$, there exists at least one edge between $G_{c_1}$ and $G_{c_2}$.
  \end{corollary}
Corollary~\ref{cor:batch} serves as a check for whether a given batching scheme yields a unique global minimizer geometry or not. It also provides guidance for designing mini-batches. We elaborate on this below.

\subsection{\textit{Batch-binding} ensures unique OF minimizer and improves convergence}  \label{Sec:Binding_examples}
Corollary~\ref{cor:batch} shows that an arbitrary set of mini-batches is not guaranteed to have OF geometry as its minimizing configuration in the embedding space. To enforce the OF geometry as the unique minimizer, we introduce a simple scheme with low computational overhead as shown in Alg.~\ref{alg:binding}. 
See Fig.~\ref{fig:Binding_Examples_Visuals} in the appendix for a visual example.

 \begin{algorithm}
  \caption{Batch-binding}\label{alg:binding}
  \begin{algorithmic}[1]
    \Procedure{Batch-binding}{a set of mini-batches $\mathcal{B}$, number of classes $k$}
    \State \textbf{Binding examples}: A set of examples represents a set of binding examples if it includes exactly one example from every class. This set has, thus, $k$ elements. \label{def:binding}
\State To every batch $B\in\mathcal{B}$ add the chosen set of binding examples to create a new batch configuration $\mathcal{B}^\prime$
    \EndProcedure
  \end{algorithmic}
\end{algorithm}

Batch-binding adds a constant $k$ to the size of every mini-batch. $k$ is typically smaller than the batch sizes used to train SCL which could range from $2048$ to $6144$ \citep{khosla2020supervised}. While adding small computational overhead, this method guarantees that $\mathcal{B}^\prime$ satisfies the conditions of Cor.~\ref{cor:batch} and SCL learns an OF geometry. We conduct a series of experiments to illustrate the impact of batch binding on convergence to OF. 

To verify the role of batch selection on the learned features, we train ResNet-18 on CIFAR-10 under three batching scenarios: 
\begin{enumerate}
    \item \textbf{No batch-shuffling}: a mini-batch set $\mathcal{B}$ that partitions the training examples once at initialization and is held constant across training epochs. 
    \item \textbf{Batch-shuffling}: the examples are divided into mini-batch partition, with a random reshuffle of the examples at \textit{every} epoch.
    \item \textbf{No batch-shuffling + batch-binding}: we construct a fixed partition of examples into mini-batches; then, we perform batch-binding. In order to focus on the impact of batch selection strategies, we consider a relatively small batch-size of $128$.
\end{enumerate}

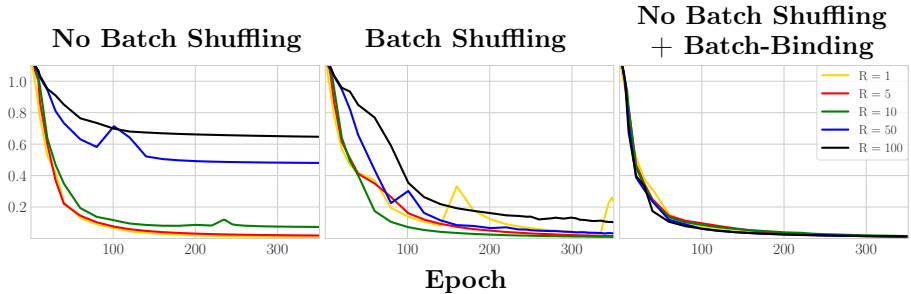
\begin{figure}[ht]
    \hspace*{-20pt}
    \begin{subfigure}[b]{\textwidth}
        \centering
        \begin{tikzpicture}
            \node at (0.0,0) 
            {\includegraphics[width=0.75\textwidth]{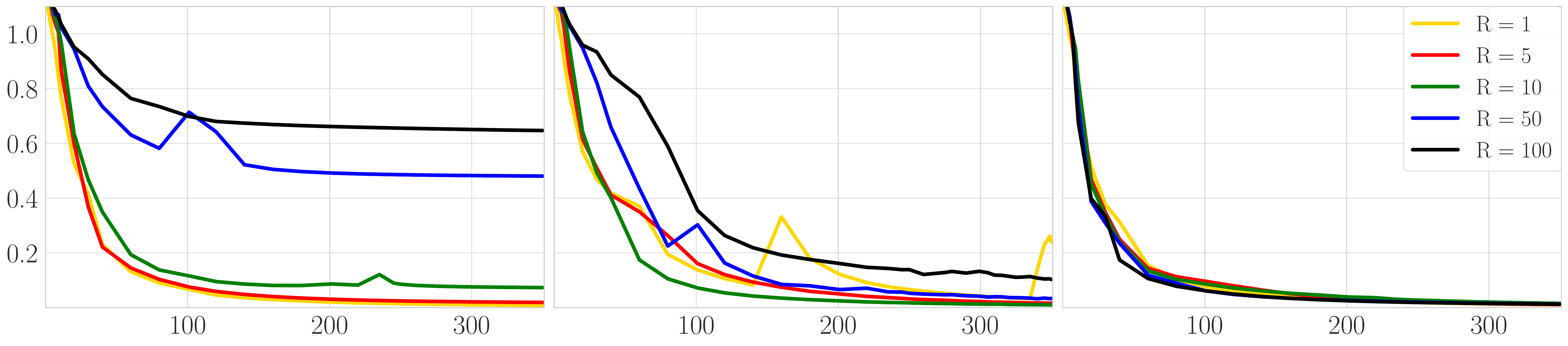}};
            \node at (-3.7,1.6) [scale=0.9]{\textbf{No Batch Shuffling}};
            \node at (0.1,1.6) [scale=0.9]{\textbf{Batch Shuffling}};
            \node at (4,1.9) [scale=0.9]{\textbf{No Batch Shuffling}};
            \node at (4,1.5) [scale=0.9]{\textbf{+ Batch-Binding}};
            \node at (0.1,-1.6) [scale=0.9]{\textbf{Epoch}};
        \end{tikzpicture}
    \end{subfigure}
    \hspace*{-20pt}
    \captionsetup{width=0.95\linewidth}
    \caption
    {{Convergence to the OF geometry for various batching schemes {including the analysis-inspired scheme ``No Batch Shuffling + Batch-Binding''}. ResNet-18 trained on CIFAR-10 with a small batch size of 128, see text. See also Fig.~\ref{Fig:Deepnet_SCL_Geom_NC_BindingBatch} in the appendix}} 
    \vspace{-15pt}
    \label{Fig:Deepnet_SCL_Geom_NC_BindingBatch_main}
\end{figure}

We remark that Thm.~\ref{thm:scl_mini} and Cor.~\ref{cor:batch} can explain the behaviors of all three schemes. Moreover, the batch-binding strategy guided by our analytical results is effective at ensuring fast and predictable convergence of deep-nets to a unique OF geometry. Fig.~\ref{Fig:Deepnet_SCL_Geom_NC_BindingBatch_main} shows the distance of learned embeddings to the OF geometry for the three batching scenarios mentioned above. In case of the fixed partition batching (left), the features do not converge to OF, especially with large imbalance ratios. This behavior is consistent with the conditions identified by Thm.~\ref{thm:scl_mini}, because in a typical random partition, the corresponding induced subgraphs are not necessarily connected. On the other hand, when the batching is performed with a randomly reshuffled data ordering (center), the geometry converges more closely to the OF, albeit with certain epochs deviating significantly from the convergence direction. We hypothesize that random reshuffling creates an opportunity for different examples to interact sufficiently, thus, when trained long enough, to converge close to an OF geometry. 
Finally, with the third batching strategy of batch-binding (right), even without shuffling the samples at every epoch, we observe a consistently fast convergence to the global optimizer, OF. These observations provide strong evidence supporting our analysis of batching predicting the behavior of SCL trained deep-nets. 

While the above studies the impact of batching itself, we perform a number of additional experiments in order to illustrate the impact of binding examples. In particular, we observe that batch binding helps improve convergence to OF geometry when training with less powerful models (see Fig.~\ref{Fig:DenseNet_BindingsComparison} for DenseNet experiments in the appendix), more complex dataset (see Fig.~\ref{Fig:Deepnet_SCL_CIFAR100_GMu} in the appendix and Fig.~\ref{Fig:Deepnet_SCL_Geom_GMu}  for CIFAR100 and TinyImageNet results) and even convergence to ETF in the absence of ReLU under balanced training (See Fig.~\ref{fig:binding_ETF} in the appendix). Such results emphasize the importance of batch binding and encourage further analysis of batching under contrastive learning.

\section{More Related Work}\label{sec:related_main}

The simple concept behind SCL, introduced as an extension of the contrastive loss \citep[e.g.,][]{chen2020simple, tian2019contrastive} to the fully supervised setting by \citet{khosla2020supervised},  involves pulling together the normalized features of examples belonging to the same class while pushing them away from examples of other classes.  Despite its simplicity, SCL offers a generalization of existing loss functions like the triplet loss \citep{weinberger2005distance} and the N-pair loss \citep{sohn2016improved}, and surpasses the performance of the CE loss on standard vision classification datasets,  while also being more robust to natural corruptions in the data and less sensitive to hyperparameters \citep{khosla2020supervised}. 


In recent years, there has been a notable interest regarding the properties of the embedding space and weights learned through unsupervised/supervised contrastive losses, as well as CE. 
Of particular importance is the endeavor to uncover the fundamental distinctions between these losses to effectively harness each of their strengths and formalize principled ways to combine them.
Specifically, starting by \cite{NC}, an increasing series of recent works have focused on uncovering the embedding geometry of CE, Mean-Squared Error (MSE) losses, and their variants. We highlighted some of these works in Sec.~\ref{sec:intro} while numerous other works contribute to these investigation \citep[e.g.,][]{zhou2022optimization,zhou2022all,yaras2022neural,gao2023study,han2021neural,sukenik2023deep}. In addition to the scientific curiosity surrounding such  studies, this exploration has paved the way for novel CE-based approaches to training DNNs on imbalanced data \citep{behnia2023implicit,liang2023inducing,xie2023neural,sharma2023learning,dang2023neural,yang2022inducing}. 

However, less attention has been devoted in the existing literature to a comparable set of results for the SCL. \citet{graf2021dissecting} is the first to study the embedding geometry of SCL for balanced data and comparing it to that of CE. To tackle the imbalances, and inspired by \citet{graf2021dissecting}
that suggests balanced data is crucial for obtaining symmetric embeddings, 
\citet{zhu2022balanced} have recently proposed and investigated a modification of the SCL that boosts performance under class imbalances.  
Our work extends  and complements these two studies by showing that SCL with ReLU can actually learn symmetric embedding structure even in the presence of imbalances. While preparing the manuscript of our paper, we became aware of a recent work \citep{cho2023mini}, where the authors consider the mini-batch optimization of the unsupervised contrastive loss and propose a loss-based choice of mini-batches to speed up convergence to ETF. Since their setting is different, it would be interesting to see how our ideas can benefit their findings and vice versa.

\section{Concluding remarks}\label{sec:conclusions}
\vspace{-0.3cm}
We have shown consistent empirical and theoretical evidence that SCL learns training embeddings that converge to the OF geometry irrespective of the level of class imbalance. We believe this finding contributes a unique result to the growing literature of neural-collapse / implicit-geometry phenomena. For balanced data, we extend the contributions of \citet{graf2021dissecting} to the case in presence of ReLU activations. Further, our results identify exact conditions on the mini-batch selection to achieve the OF geometry, allowing for a wider set of possibilities in batching than found by \cite{graf2021dissecting} for SCL without ReLU. For imbalanced data, our results are the first explicit characterization of the geometry, concluding that imbalances do \emph{not} alter the geometry if a ReLU activation is added at the last layer. In both cases, these advancements are achieved by analyzing a refined UFM model, \ref{eq:SCL_UFM_full}, that closely aligns with experimental conditions by constraining embeddings to the non-negative orthant. This model also leads to new findings about the intricate role of batching in SCL optimization, that may be of independent interest. A future investigation of exact characterization of the geometry with imbalances and without ReLU can help complete the understanding of the implicit-geometry of SCL. Although the majority of implicit geometry characterizations in the literature require $d>k$, it is interesting to extend our findings to settings where $d<k$ following the steps of \cite{gao2023study} for the CE loss. Likewise, our finding regarding the crucial role of batching in geometry opens the door to further investigations aimed at devising efficient batch strategies in cases with a large number of classes.
Finally, like most studies on neural-collapse phenomena, our results share a common limitation: they only provide insights into the behavior during training. On the other hand, there is a line of research that explores algorithmic modifications to SCL \citep{gunel2020supervised,jitkrittum2022elm,samuel2021distributional,li2022targeted,kang2021exploring,li2022targeted}, often rooted but not explicitly connected to geometric principles, to enhance generalization under imbalances. Ultimately, we envision the two research streams merging through the ongoing exchange of ideas.


\printbibliography
\newpage
\clearpage
\appendix
\onecolumn
\addtocontents{toc}{\protect\setcounter{tocdepth}{3}}
\tableofcontents
\vspace{30pt}
\noindent\textbf{Additional Notations.}~
While recalling the notation mentioned in Sec.~\ref{sec:intro}, we note a few more below. $\otimes$ denotes Kronecker products.
We use $\ones_m$ to denote an  $m$-dimensional vector of all ones. For vectors/matrices with all zero entries, we simply write $0$, as dimensions are easily understood from context. 
$\Vb^\dagger$ its Moore-Penrose pseudoinverse. $\nabla_\Vb\Lc \in\R^{m\times n}$ is the gradient of a scalar differentiable function $\Lc(.)$ with respect to $\Vb$.

\section{Proof of Thm.~\ref{thm:scl_full}}\label{app:proof_details}
 For $c\in[k]$, let $\mathcal{C}_c = \{i: y_i = c\}$ be the set of examples belonging to class $c$, and define $\Lc_i(\Hb)$ as follows,
\begin{align}\label{eq:L_i}
  \Lc_i(\Hb) &= \sum_{{j\in\mathcal{C}_{y_i},j\neq i}}
  \log\Big(\sum_{{\ell\in[n],\ell\neq i}}\exp{\left(\h_i^\top\h_\ell - \h_i^\top\h_j\right)}\Big)\nn\\
     &= \sum_{{j\in\mathcal{C}_{y_i},j\neq i}}\log\Big(\sum_{\ell\in\mathcal{C}_{y_i}, \ell\neq i}\exp{(\h_i^\top\h_\ell - \h_i^\top\h_j)}+\sum_{\ell\not\in \mathcal{C}_{y_i}}\exp{(\h_i^\top\h_\ell - \h_i^\top\h_j)}\Big).
\end{align}
Then, we can rewrite the full-batch SCL in \eqref{eq:SCL_full} as, 
\begin{align*}
    \Lc_\text{full}(\Hb)&= \sum_{i\in[n]}\frac{1}{n_{y_i}-1} \Lc_i(\Hb) = \sum_{c \in [k]}\frac{1}{n_{c}-1} \sum_{i\in\mathcal{C}_c}\Lc_i(\Hb).
\end{align*}
 Now for $c\in[k]$, consider example $i\in\mathcal{C}_c$, and define $$\mathbf{a}_i := \frac{1}{n_c-1}\sum_{\ell\in\mathcal{C}_{c}, \ell\neq i}
\h_\ell, \quad \mathbf{b}_i := \frac{1}{n - n_c}\sum_{\ell\not\in\mathcal{C}_{c}}
\h_\ell.$$ 
Note $\mathbf{a}_i$ is the mean-embedding of all examples in class $c$ but $i$, and $\mathbf{b}_i$ is the mean-embedding of all examples not in class $c$. Starting from \eqref{eq:L_i}, by applying Jensen's inequality to the strictly convex function $f_1(x) := \exp(x - \h_i^\top\h_j)$ we have,
 %
 \begin{align}\label{eq:jensen_1}
 \Lc_i(\Hb) 
 &\geq \sum_{{j\in\mathcal{C}_{c},j\neq i}}\log\Big( (n_c-1) \exp{(\h_i^\top\mathbf{a}_i - \h_i^\top\h_j)}  +  (n-n_c) \exp{(\h_i^\top\mathbf{b}_i - \h_i^\top\h_j)}\Big)\nn\\
&=\sum_{{j\in\mathcal{C}_{c},j\neq i}}\log\Big( \big((n_c-1)\exp{(\h_i^\top\mathbf{a}_i)}  +  (n-n_c) \exp{(\h_i^\top\mathbf{b}_i})\big) \exp{(- \h_i^\top\h_j)}\Big)\nn\\
&\stackrel{(i)}{=}\sum_{{j\in\mathcal{C}_{c},j\neq i}}\log\Big(\alpha_i \exp{(- \h_i^\top\h_j)}\Big)\nn\\
&\stackrel{(ii)}{=}(n_c-1)\log\Big(\alpha_i \exp{(- \h_i^\top(\frac{1}{n_c-1}\sum_{{j\in\mathcal{C}_{c},j\neq i}}\h_j))}\Big)\nn\\
&= (n_c-1)\log(\alpha_i \exp(-\h_i^\top\mathbf{a}_i))\nn\\
&= (n_c-1)\log\Big((n_c-1) + (n-n_c)\exp\big(\frac{1}{n-n_c}\h_i^\top \sum_{\ell \not\in \mathcal{C}_c} \h_\ell - \frac{1}{n_c-1} \h_i^\top \sum_{\ell \in \mathcal{C}_c, \ell\neq i} \h_\ell\big)\Big),
 \end{align}
 where in $(i)$, we define $\alpha_i := (n_c-1)\exp{(\h_i^\top\mathbf{a}_i)}  +  (n-n_c) \exp{(\h_i^\top\mathbf{b}_i})$, and in $(ii),$ we use the fact that the function $f_2(x):=\log(\alpha_i\exp(x))$ is an affine function. Now consider all samples $i\in\mathcal{C}_c$. From \eqref{eq:jensen_1}, \begin{align}\label{eq:Jensen_L_i}
 \frac{1}{n_c-1}\sum_{i\in\mathcal{C}_c} &\Lc_i(\Hb) \geq \sum_{i\in\mathcal{C}_c} \log\Big((n_c-1) + (n-n_c)\exp\big(\frac{1}{n-n_c}\h_i^\top \sum_{\ell \not\in \mathcal{C}_c} \h_\ell - \frac{1}{n_c-1} \h_i^\top \sum_{\ell \in \mathcal{C}_c, \ell\neq i} \h_\ell\big)\Big)\nn\\
 &\geq n_c\log\Big((n_c-1) + (n-n_c)\exp\big(\frac{1}{n_c(n-n_c)} \sum_{\substack{i\in\mathcal{C}_c \\ \ell \not\in \mathcal{C}_c}} \h_i^\top \h_\ell - \frac{1}{n_c(n_c-1)} \sum_{\substack{i\in\mathcal{C}_c \\ \ell \in \mathcal{C}_c, \ell\neq i}} \h_i^\top \h_\ell\big)\Big).
 \end{align}
 In the last line we use Jensen's inequality on  $f_3(x):=\log((n_c-1)+(n-n_c)\exp(x))$ which is strictly convex since 
 $n_c>1$ and $n-n_c>0$. 

 By the ReLU constraints we have $\h_i \geq 0,\, i\in[n]$, which implies $\h_i^\top\h_j\geq 0,\,\forall i,j\in[n]$. Further, by Cauchy-Schwarz inequality $\h_i^\top\h_j \leq \norm{\h_i}\norm{\h_j}=1$, with equality achieved if and only if $\h_i=\h_j$. Since $f_3$ is non-decreasing, we can simplify the bound in \eqref{eq:Jensen_L_i} as follows,
 \begin{align}\label{eq:cauchy}
     \frac{1}{n_c-1}\sum_{i\in\mathcal{C}_c} &\Lc_i(\Hb) \geq n_c\log\Big((n_c-1) + (n-n_c)e^{- 1}\Big).
 \end{align}
 Summing over all classes $c\in[k]$, we get the final bound on the full-batch SCL:
\begin{align*}
    \Lc_\text{full}(\Hb) = \sum_{c\in[k]}\frac{1}{n_c-1}\sum_{i\in\mathcal{C}_c} \Lc_i(\Hb) \geq \sum_{c\in[k]} n_c\log\Big((n_c-1) + (n-n_c)e^{- 1}\Big).
\end{align*}
To achieve the lower-bound, from \eqref{eq:cauchy}, we require $\h_i=\h_j$ if $y_i=y_j$ and $\h_i^\top\h_j=0$ otherwise. This requirement also satisfies the equality condition for the Jensen inequalities applied in \eqref{eq:jensen_1} and \eqref{eq:Jensen_L_i}. Thus, any $\Hb$ achieving the lower-bound follows an OF geometry (Def.~\ref{def:ncof}), which exists as long as $d\geq k$. 




\section{Proof of Thm.~\ref{thm:scl_mini}}\label{app:proof_thm2}

    Consider the mini-batch SCL in \eqref{eq:scl_batch}, repeated below for convenience:
 \begin{align*}
     \Lc_{\text{batch}}(\Hb):=\sum_{B \in \mathcal{B}}\sum_{i\in B}\frac{1}{n_{B,y_i}-1}\sum_{\substack{j\in B\\j \neq i, y_j = y_i}}\log\bigg(\sum_{\substack{\ell \in B \\ \ell \neq i}}\exp{\left(\h_i^\top\h_\ell - \h_i^\top\h_j\right)}\bigg),
 \end{align*}
 Denoting the loss over a batch $B$ by 
 \begin{align*}
     \Lc_{B}(\Hb) := \sum_{i\in B}\frac{1}{n_{B,y_i}-1}\sum_{\substack{j\in B\\j \neq i, y_j = y_i}}\log\bigg(\sum_{\substack{\ell \in B \\ \ell \neq i}}\exp{\left(\h_i^\top\h_\ell - \h_i^\top\h_j\right)}\bigg),
 \end{align*}
 we have $\Lc_{\text{batch}}(\Hb)=\sum_{B \in \mathcal{B}}\Lc_{B}(\Hb)$.
 Now, for a given batch $B$, we apply Thm.~\ref{thm:scl_full} to get the lower bound and conditions for equality:
 \begin{align*}
      \Lc_B(\Hb) \geq \sum_{c \in [k]}n_{B,c}\log\left(n_{B,c}-1+(n_{B}-n_{B,c})e^{-1}\right).
  \end{align*}
  Moreover, equality holds if and only if for every pair of samples $i,j\in B$: $\h_i^\top\h_j=0$ if $y_i\neq y_j$ and $\h_i=\h_j$ if $y_i=y_j$. With this, the overall loss can be bounded from below by summing the individual lower bounds over different batches:
  \begin{align*}
      \Lc_\text{{batch}}(\Hb) \geq \sum_{B \in \mathcal{B}}\sum_{c \in [k]}n_{B,c}\log\left(n_{B,c}-1+(n_{B}-n_{B,c})e^{-1}\right),
  \end{align*}
  with equality achieved if and only if for every $B\in\mathcal{B}$ and every pair of samples $i,j\in B$, $\h_i^\top\h_j=0$ if $y_i\neq y_j$ and $\h_i=\h_j$ if $y_i=y_j$. As long as $d \geq k$, the equality conditions of every individual batch can be satisfied simultaneously.

\subsection{Proof of Cor.~\ref{cor:batch}}

{From Thm.~\ref{thm:scl_mini}, the optimal configuration of embeddings in \ref{eq:SCL_UFM_full} under the mini-batch SCL depends on how batch conditions interact with each other. Specifically, recall that the equality in \eqref{eq:scl_mini_lb} is achieved if and only if for any batch $B\in\mathcal{B}$, and each pair of samples $(i,j)\in B$, $\h_i=\h_j$ if $y_i=y_j$, and $\h_i^\top\h_j = 0$ otherwise. The OF geometry clearly satisfies all these conditions and achieves the minimal cost of \ref{eq:SCL_UFM_full}. However, as discussed in Sec.~\ref{sec:mini_batch_theory}, these equality conditions may be satisfied by configurations other than OF for an arbitrary batching scheme. So, in Cor.~\ref{cor:batch}, we specify the requirements of a batching scheme in order for the global optimal of the mini-batch SCL to be unique (up to global rotations), and match the optimal of the full-batch SCL, which is the OF geometry. }

%
%
To prove the corollary, we separately address the `IF' and `ONLY IF' parts of the  Cor.~\ref{cor:batch}. 

\noindent$\bullet$~\textbf{`IF' direction.}~{Assume the batching scheme satisfies both conditions of Cor.~\ref{cor:batch}: {1) $\forall c\in[k]$, the induced subgraph $G_c$ is connected, and 2) $\forall c\neq c'\in[k]$, there exists an edge between $G_c$ and $G_{c'}$.} We show below that under these two conditions the optimum of \ref{eq:SCL_UFM_full} under mini-batch SCL follows an OF geometry. {In other words, we show that the optimal embeddings align if they belong to the same class (NC) and are orthogonal if they have different labels (mean-embeddings follow k-OF). (See Def.~\ref{def:ncof}.)}}

   {
   \textit{{NC:}} Consider a class $c$. From our assumption, the induced subgraph $G_c$ is connected. Thus, a path exists from any node (representing corresponding example) to any other node in $G_c$. 
   Consider three nodes\footnote{{The same arguments and considerations apply when $G_c$ consists of only two nodes.}} along a path, indexed by $i,j,l$ that belong to class $c$. Let there be an edge between $i,j$ and $j,l$ each. By Def.~\ref{def:big} of the Batch Interaction Graph, examples $i$ and $j$ are present in a batch, say $B_1$, and examples $j,l$ belong to a batch $B_2$ that is possibly different from $B_1$. From Thm.~\ref{thm:scl_mini}, we can infer that at the optimal solution, we have $\h_i = \h_j$, due to equality conditions for batch $B_1$. Also, $\h_j = \h_l$, due to equality conditions for batch $B_2$. Thus, by transitivity, we have $\h_i = \h_j = \h_l$. Repeating this argument along any path in $G_c$ for every $c \in [k]$, we obtain the NC property: $\h_i = \h_j, \forall i,j:y_i = y_j=c$ for every class $c \in [k]$.}
    
    \textit{$k$-OF:} From our assumption, for any pair of classes $c_1$ and $c_2$, there exists an edge between $G_{c_1}$ and $G_{c_2}$ {connecting}
    at least one pair of nodes $i,j: i\in G_{c_1}, j \in G_{c_2}$. By definition of the Batch Interaction Graph, we know that examples $i$ and $j$ belong in at least one batch. {Thus,} by Thm.~\ref{thm:scl_mini}, at the optimal solution, we have $\h_i \perp \h_j$ since examples $y_i=c_1\neq c_2=y_j$. Now, {by NC,  $\h_i=\mub_{c_1}$, $\h_j=\mub_{c_2}$, and thus, $\mub_{c_1}\perp \mub_{c_2}$. This holds for every pair of classes $c_1\neq c_2 \in [k]$. Therefore, the matrix $\M=[\mub_1,...,\mub_k]$ of class-mean embeddings forms a $k-$OF.}
     

    {Combining the two statements above, we conclude that at the global optimal solution, the embeddings follow the OF geometry, as desired.}

\noindent$\bullet$~\textbf{`ONLY IF' direction.} For the other direction, it suffices to show that if either of the two conditions in Cor.~\ref{cor:batch} does not hold, then there exists an optimizer that does not follow the OF geometry. 
Specifically, we show that when either of the two conditions is violated {and $d\geq k+1$}, there exists an embeddings matrix $\widetilde{\Hb}$ attaining the loss lower-bound that does not satisfy one of the two requirements of the OF geometry: 1) $\widetilde{\Hb}$ does not follow NC, or 2) the corresponding mean-embeddings $\widetilde{\M}$ do not arrange as a $k$-OF.

\textit{Case 1.} Suppose for a $c\in[k]$, the induced subgraph $G_c$ is not connected, and without loss of generality, assume $c=1$. This means that $G_1$ has at least two separate components. Denote the nodes in each of the two component by $V_1^1$ and $V_1^2$ respectively, and recall for $c\geq 2$, $V_c=\{i:y_i=c\}$. As $d\geq k+1$, we can choose a set of $k+1$ vectors $[\widetilde{\mub}_1^1, \widetilde{\mub}_1^2, \widetilde{\mub}_2, \widetilde{\mub}_3,...,\widetilde{\mub}_k]$ such that they form a $(k+1)-$OF. Define $\widetilde{\Hb}=[\widetilde{\h}_1,...,\widetilde{\h}_n]$ as follows: 
\begin{align*}
    \forall i \in V_1^1, &\quad \widetilde{\h}_i = \widetilde{\mub}_1^1\\
    \forall i \in V_1^2, &\quad \widetilde{\h}_i = \widetilde{\mub}_1^2\\
    \forall i \in V_c,\,c\in[k], &\quad \widetilde{\h}_i = \widetilde{\mub}_c\,.
\end{align*}
Then, $\widetilde{\Hb}$ satisfies the equality conditions in Thm.~\ref{thm:scl_mini} since there is no edge between the nodes in $V_1^1$ and $V_1^2$ by the assumption. However, the embeddings in class $y=1$ do not align, since $\widetilde{\mub}_1^1$ is orthogonal to $\widetilde{\mub}_1^2$. Thus, $\widetilde{\Hb}$ optimizes \ref{eq:SCL_UFM_full} while it does not satisfy NC and differs from the OF geometry.

\textit{Case 2.} Suppose there exists $c_1\neq c_2\in[k]$ for which there is no edges between $G_{c_1}$ and $G_{c_2}$, and without loss of generality, assume $c_1=1, c_2=2$. Consider an embedding matrix $\widetilde{\Hb}$ that satisfies NC, and the corresponding mean-embedding matrix $\widetilde{\M}=[\widetilde{\mub}_1,\widetilde{\mub}_2,...,\widetilde{\mub}_k]$ is such that $\forall c\neq c'\in\{2,...,k\}$, $\widetilde{\mub}_c^\top\widetilde{\mub}_{c'}=0$ and $\widetilde{\mub}_1=\widetilde{\mub}_2$. Such an $\widetilde{\M}$ exists as $d\geq k$ and all we need is to have $[\widetilde{\mub}_2,...,\widetilde{\mub}_k]$ be a $(k-1)-$OF in the non-negative orthant. Since, there is no edge between $G_1$ and $G_2$, there is no $B\in\mathcal{B}$ that includes samples from classes $y=1$ and $y=2$ simultaneously. Equivalently, to achieve the lower-bound of Thm.~\ref{thm:scl_mini}, we do not require orthogonality between any pairs of samples $i\in\mathcal{C}_1$ and $j\in\mathcal{C}_2$. Therefore, $\widetilde{\Hb}$ is an optimal solution of the \ref{eq:SCL_UFM_full}. However, the mean-embeddings do not follow a $k-$OF. In fact, this optimal geometry, does not distinguish the samples in classes $y=1$ and $y=2$.

Therefore, the global minimizer will be uniquely an OF if and only if the batching scheme satisfies both the conditions stated in Cor.~\ref{cor:batch}.


\subsection{Batch analysis for \ref{eq:SCL_UFM_full} vs UFM}
\label{rem:compare_graf}

{\citet[Thm.~2]{graf2021dissecting} studies the UFM without ReLU constraints and with balanced labels, and proves that the global solution is a simplex ETF. The authors note that their proof relies on the batch construction as the set of all combinations of a given size.
In contrast, we have proved that for \ref{eq:SCL_UFM_full}  the global solution is an OF for a wider range of batching scenarios (specifically, as long as they satisfy the interaction properties characterized in Cor.~\ref{cor:batch}). Without ReLU, as noted by \citet[Fig.~5]{graf2021dissecting}, the optimal configuration of examples in each batch can have a different geometry, depending on the label distribution of the examples within the batch. However, \emph{with} ReLU, the optimal configuration of every batch is an OF over the classes whose examples are present in the batch. In particular, there is no contradiction between two mini-batches having both overlapping and mutually exclusive classes, since the optimal configuration of one batch does not violate the optimal configuration of another. Furthermore, the overall batch construction can have a unique OF as the optimal configuration provided the conditions in Cor.~\ref{cor:batch} are satisfied. The conditions include the batching assumed by \citet{graf2021dissecting} as a special case, while being applicable in less restrictive scenarios. {Sec.~\ref{Sec:Binding_examples} explored the implications of this finding, by studying the criteria for a batching scheme to lead to a unique minimizer geometry and further suggesting a simple scheme to convert an arbitrary batching to one satisfying these criteria.} }

\section{Proofs for Section \ref{sec:UFMvsUFM+}}

\subsection{Proof of Lemma \ref{lem:OF2ETF}}

Let $\Vb=\begin{bmatrix}
    \vb_1&\ldots&\vb_k
\end{bmatrix}$    and $\bar\Vb=\begin{bmatrix}
    \bar\vb_1&\ldots&\bar\vb_k
\end{bmatrix}$. 
By definition of a $k$-OF, we know that $\Vb^\top\Vb\propto\Id$. Without loss of generality suppose the constant of proportionality is $1$ so that  $\Vb^\top\Vb=\Id\,.$ Since all $\vb_i$s are orthonormal, this then implies that $\Vb^\top\vb_G=\frac{1}{k}\ones_k$ and $\vb_G^\top\vb_G=\frac{1}{k}\,.$ These put together gives the desired as follows:
\begin{align*}
    \bar\Vb^\top\bar\Vb &= \left(\Vb-\vb_G\ones_k^\top\right)^\top\left(\Vb-\vb_G\ones_k^\top\right) = \Vb^\top\Vb - \ones_k\vb_G^\top\Vb-\Vb^\top\vb_G\ones_k^\top+(\vb_G^\top\vb_G)\ones_k\ones_k^\top=\Id_k - \frac{1}{k}\ones_k\ones_k^\top\,.
\end{align*}

\subsection{Proof of Lemma \ref{lem:counterexample}}

    To rule out the optimality of ETF in imbalanced setups, we show ETF does not minimize the loss in a $k=3$ class example. Specifically, consider a STEP-imbalanced training set with 3 classes of sizes $[R n_\text{min}, n_\text{min}, n_\text{min}]$ with $n_\text{min}\geq 2$ and $R \geq 10$. Suppose $\Hb_\text{ETF}$ follows the ETF geometry. Then, from Def.~\ref{def:etf} and the feasibility condition, for the mean-embeddings $\M_\text{ETF}$ we have $\M_\text{ETF}^\top\M_\text{ETF} = \frac{k}{k-1}(\Id_k - \frac{1}{k}\ones_k\ones_k^\top).$ Thus, the value of the loss function at ETF is 
    \begin{align*}
        \Lc_{\text{{full}}}(\Hb_\text{ETF}) = n_\text{min} \Big(R \log\big(R n_\text{min}-1+2n_\text{min}e^{-\frac{3}{2}}\big) + 2\log\big(n_\text{min}-1+n_\text{min}(R+1)e^{-\frac{3}{2}}\big)\Big).
    \end{align*}
    Now, consider $\widetilde{\Hb} = [\widetilde{\mub}\otimes \ones_{R n_\text{min}}^\top, -\widetilde{\mub}\otimes \ones_{ n_\text{min}}^\top, -\widetilde{\mub}\otimes \ones_{n_\text{min}}^\top]$, where {$\widetilde{\mub}$} is an arbitrary unit-norm vector. Since $\norm{\widetilde{\mub}}=1$, it follows that  $\widetilde{\Hb}$ is in the feasible set. With this choice of $\widetilde{\Hb}$, we have,
    \begin{align*}
        \Lc_{\text{{full}}}(\widetilde{\Hb}) = n_\text{min} \Big(R \log\big(R n_\text{min}-1+2n_\text{min}e^{-2}\big) + 2\log\big(2n_\text{min}-1+R n_\text{min}e^{-2}\big)\Big).
    \end{align*}
    It is easy to check that, for imbalance levels higher than $R\geq10$, $$\Lc_{\text{{full}}}(\widetilde{\Hb}) < \Lc_{\text{{full}}}(\Hb_\text{ETF}).$$ In other words, ETF does not always attain the optimal cost in \ref{eq:ufm}.



\section{Additional Discussion  
}\label{sec:discussion}

\subsection{Detailed comparison between UFM and UFM$_+$}\label{sec:UFMvsUFM+}
Within this section, we provide an in-depth analysis
of the disparities between the predictions generated by the UFM, commonly utilized in previous studies, and the refined version \ref{eq:SCL_UFM_full} that we employ in this work to study the embedding geometry of SCL.

The majority of previous works \cite[e.g.,][]{NC,zhu2021geometric,fang2021exploring} study the geometry of learned embeddings by investigating a proxy unconstrained features model (UFM). Specifically, the UFM drops the dependence of the learned embeddings $\Hb$ on the network parameters $\thetab$, and finds optimal $\widehat{\Hb}$ that minimizes the loss function. In case of SCL, as studied in e.g., \cite{graf2021dissecting}, the corresponding UFM is as follows,
\begin{align}\label{eq:ufm}
    \arg\min_{\Hb} \Lc_{\text{{SCL}}}
    (\Hb)~~\text{{subj. to}}~~\|\h_i\|^2 = 1,\, \forall i\in[n].
    \tag{$\text{UFM}$}
\end{align}
As explained in Sec.~\ref{sec:theo_results}, this paper employs a refined version of the UFM, namely \ref{eq:SCL_UFM_full}, to characterize the representations learned by SCL. Specifically, \ref{eq:SCL_UFM_full} further constrains the embeddings $\Hb$ to be non-negative, in this way accounting for the ReLU activation commonly used in several deep-net architectures. Thus the search for the optimal $\Hb$ is restricted to the non-negative orthant.


\subsubsection{Centering heuristic}\label{sec:centering}
Many deep neural network models incorporate ReLU activations, resulting in nonnegative feature embeddings at the last layer. In contrast, the UFM does not impose any constraints on the optimal $\Hb$, allowing it to contain negative entries.  To bridge this gap, previous works \cite[e.g.,][]{NC,zhu2021geometric,fang2021exploring,zhou2022optimization,seli,zhou2022all} apply a heuristic approach called \emph{global-mean centering} on the learned embeddings before comparing their arrangement with the theoretical prediction given by the UFM. Specifically, the centering heuristic is applied as follows: Prior to comparing the geometries, we subtract the global-mean embedding $\mub_G = \frac{1}{k}\sum_{c \in [k]}\mub_c$ from all class-mean embeddings $\mub_c$ to form the \emph{centered class-mean embeddings}  $\Bar{\mub}_c$:
\begin{align*}
    \Bar{\mub}_c = \mub_c - \mub_G,\ \qquad\ \Bar{\M} = \begin{bmatrix} \Bar{\mub}_1 & \cdots & \Bar{\mub}_k \end{bmatrix}\,.
\end{align*}
Indeed, this heuristic centering operation effectively ensures that the mean-embedding matrix $\bar{\M}$, after the necessary shifting, is centered at the origin, satisfying $\bar{\M}\ones_k=0$. This property also holds for the global optimizers of the UFM found in the previous work, which served as a motivation for the heuristic centering. 

Unlike the approach mentioned above, our findings do not necessitate any heuristic embedding processing for comparing the geometries. This is because our model directly provides the geometry of the embeddings in their original form.

\subsubsection{Centered OF is simplex ETF 
}\label{ref:OFvsETF}
Focusing on the SCL, \citet{graf2021dissecting} have used the \ref{eq:ufm} to characterize the learned embeddings and have found that, for balanced classes, the simplex ETF geometry is the global optimizer of \ref{eq:ufm}. In other words, the optimal embeddings $\Hb$ adhere to the NC (Defn.~\ref{def:nc}), and the corresponding mean embeddings $\M$ form a simplex ETF.  {For the reader's convenience, we recall below the definition of simplex ETF from \cite{NC}.}
{
\begin{definition}[Simplex ETF]\label{def:etf}
    We say that $k$ vectors $\Vb = [\vb_1,\cdots,\vb_k] \in \R^{d\times k}$ form a simplex-ETF {if $\Vb^\top\Vb \propto \Id_k-\frac{1}{k}\ones_k\ones_k^\top$, i.e., for each pair of $(i,j)\in[k],$ $\|\vb_i\| = \|\vb_j\|$ and $\frac{\vb_i^T\vb_j}{\|\vb_i\|\|\vb_j\|} = \frac{-1}{k-1}$.} 
\end{definition}
}
\begin{wrapfigure}[20]{h}{0.38\textwidth} 
     \centering
            \begin{tikzpicture}
                \node at (0.0,0.0) 
                {\includegraphics[width=0.35\textwidth]{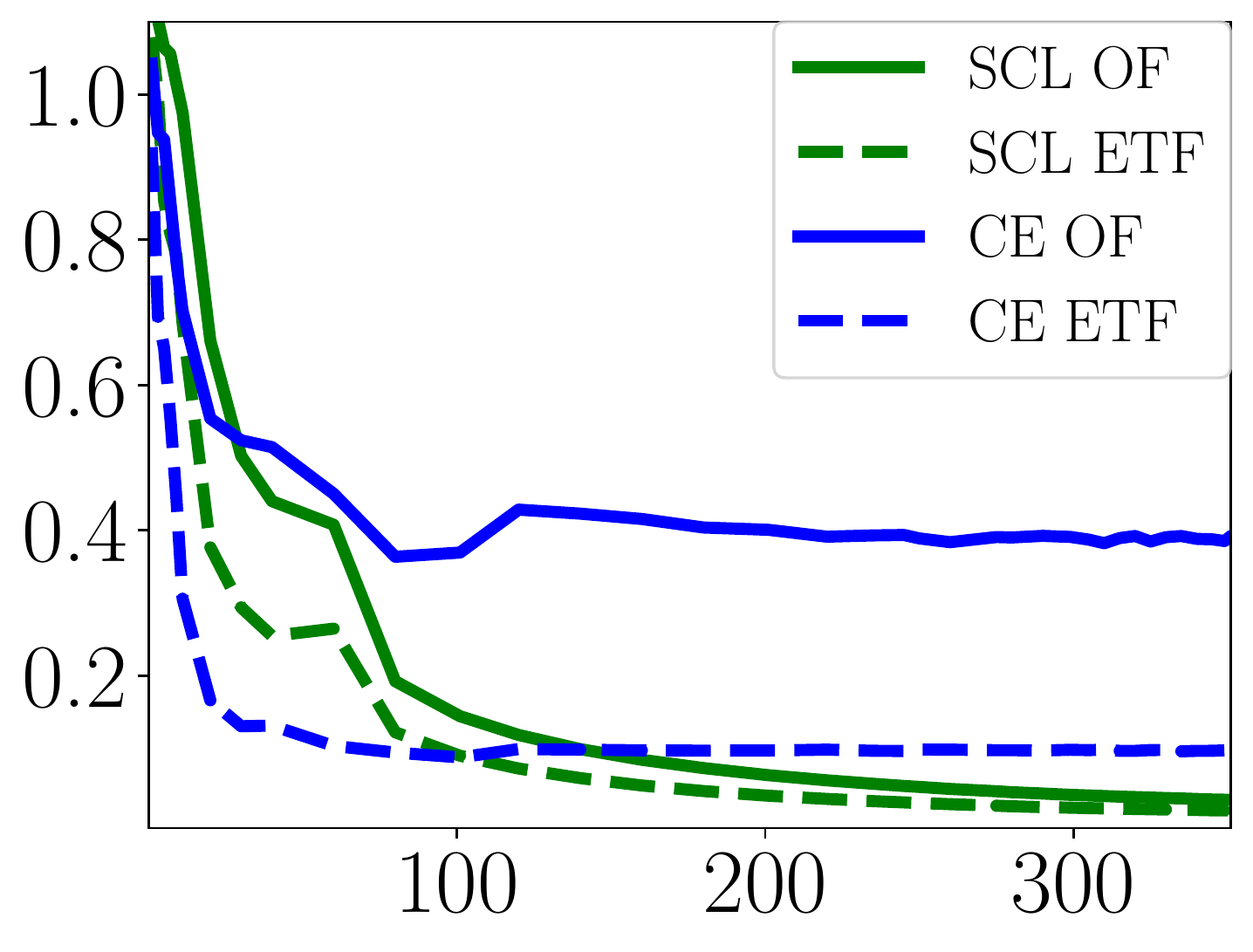}};
                 \node at (0.1,-2.3) [scale=0.9]{Epoch};
    			\node at (0.2,2.3) [scale=0.9]{$\G_\M$ \textbf{convergence}};
            \end{tikzpicture}
    \captionsetup{width=0.95\linewidth}
    \vspace{-0.1in}
    \caption{Convergence of embeddings geometries for SCL (in green) and CE (in blue)  on balanced CIFAR10 with ResNet-18. Solid lines track the distance of \emph{uncentered} embeddings to OF and  dashed lines track the distance of \emph{centered} embeddings to simplex ETF.  For CE, the learning rate and weight decay and batch size are set according to \cite{NC}. Note the convergence of SCL (solid green) to its implicit geometry is noticeably better compared to CE (dashed blue).}
    \label{fig:CE_vs_SCL_OF_ETF}
\end{wrapfigure}
In this paper, we show instead that the global optimizer of \ref{eq:SCL_UFM_full} is an OF. Remarkably, this result holds true regardless of the label-imbalance profile. However, for the purpose of comparison with the findings of \cite{graf2021dissecting}, let us specifically consider the scenario of balanced data. In this case, an intriguing question arises: 
\begin{center}
\emph{Specifically for balanced data, how does our discovery that embeddings with ReLU converge to an OF compare to the previous finding by \cite{graf2021dissecting} that embeddings without ReLU converge to an ETF?}    
\end{center}

The answer to this question lies on the centering trick discussed in Section \ref{sec:centering}. Specifically, stating that embeddings form an OF implies that \emph{centered} embeddings form an ETF. This straightforward fact is formalized in the following lemma.


\begin{lemma}\label{lem:OF2ETF}
Suppose that a set of vectors $\{\vb_1,...,\vb_k\}$ form a $k$-OF in $\R^d$ with $d \geq k$, and their mean-centered counterparts are $\Bar{\vb}_c = \vb_c - \vb_G, \forall c \in [k]$, where $\vb_G = \frac{1}{k}\sum_{c\in[k]}\vb_c$.  Then, the centered embeddings $\{\Bar{\vb}_1,...,\Bar{\vb}_k\}$ form a simplex ETF spanning a $k-1$-dimensional subspace. 
\end{lemma}
%
%

Lemma \ref{lem:OF2ETF} states that based on our finding that embeddings form an OF, we can deduce that centered embeddings form an ETF. This conclusion aligns with the analysis conducted by \cite{graf2021dissecting} on the UFM (See Fig.~\ref{fig:CE_vs_SCL_OF_ETF} for a representative comparison). However, the UFM analysis itself does not provide information regarding the necessity of the centering technique, which remains heuristic. Additionally, it is important to note that it is unclear how to establish that (uncentered) embeddings form an OF if we initially know that centered embeddings form an ETF. This is due to the unknown vector used for centering, which cannot be determined. 



\subsubsection{UFM can fail to predict the true geometry}
In the previous section, we demonstrated that for balanced data, the solution found by the \ref{eq:ufm} does not predict the true geometry of the embeddings (i.e., OF) in presence of ReLU, but it can predict the geometry of the embeddings after the application of a global-mean centering heuristic (i.e., ETF). Naturally this raises the following question: 
\begin{center}
\emph{Does the \ref{eq:ufm} consistently provide accurate predictions for the geometry of the \emph{centered} embeddings? In other words, does the OF geometry predicted by our refined \ref{eq:SCL_UFM_full} always align with being a global optimizer of the UFM after centering?}    \end{center}

The lemma presented below demonstrates that the answer to this question is negative. Specifically, it shows that the global optimizer of \ref{eq:ufm} is sensitive to the label distribution of the training set, thus ETF is not necessarily a global optimizer in the presence of imbalances.  This is in contrast to \ref{eq:SCL_UFM_full}, for which we showed that the global optimizer  is consistently an OF, irrespective of the labels distribution. In other words, the difference between \ref{eq:ufm} and \ref{eq:SCL_UFM_full} cannot be addressed by only considering the centering of the optimal embeddings in general.



\begin{lemma}\label{lem:counterexample}
    If classes in the training set are not balanced, 
    the global solution of \ref{eq:ufm} is not necessarily an ETF.  Thus, the solution of \ref{eq:SCL_UFM_full} (which according to Thm.~\ref{thm:scl_full} is an OF) is in general different to that of \ref{eq:ufm} even after centering is applied.
\end{lemma}

\subsection{Comparing implicit Geometries: SCL vs CE -- A summary}
{We conclude this discussion by providing a summary of the observations on the contrasting implicit geometries of embeddings between the two loss functions: SCL and CE.
\\
\noindent~~\textbf{1. The SCL geometry with ReLU is robust to label-imbalances.} Specifically, the geometry of SCL embeddings exhibits symmetry consistently, whereas imbalances introduce distortions in the implicit geometry of SCL without ReLU and also for CE \citep{fang2021exploring,seli}.
\\
\noindent~~\textbf{2. The convergence of SCL to its implicit geometry is notably superior.} The measured geometry of SCL exhibits a faster decrease in distance from its analytic formula (i.e., OF) with each epoch, eventually reaching lower values. This holds true not only for label-imbalanced data, where \cite{seli} finds that the convergence of embeddings deteriorates as the imbalance ratio increases, but also for balanced data, where the convergence of embeddings with CE is inferior compared to classifiers \citep{NC}; see Fig. \ref{fig:CE_vs_SCL_OF_ETF} for visualization.
\\
\noindent~~\textbf{3. For SCL, batching  matters.}}~During CE training, when using any arbitrary batching method, there exists an implicit interaction among all the examples, that is facilitated by the inclusion of the classifier vectors in the loss objective. On the other hand, with SCL, the interactions between examples are explicitly controlled by the mini-batches. Specifically, our analysis of the role of batches in Secs.~\ref{sec:mini_batch_theory} and \ref{Sec:unique}, along with the batch-binding scheme presented in Sec. \ref{Sec:Binding_examples} reveal that the structure of mini-batches critically affects the geometry of the learnt embeddings. 

\section{Additional experimental results and discussion}\label{app:exp_appex}
\subsection{Details on the main experimental setup}\label{sec:exp_details_SM}
In our deep-net experiments, we focus on two common network architectures, ResNet and DenseNet. Specifically, we use ResNet-18, ResNet-34 and DenseNet-40 with approximately $63$, $11$ million and $200$ thousand trainable parameters, respectively. For all models, we replace the last layer linear classifier with a normalization layer (normalizing such that $\| \h_i\|=1$ for $i \in [n]$) of feature dimension $d=512$. Specifically, following  \cite{graf2021dissecting}, we  directly optimize the normalized features that are then used for inference. (We note this is slightly different compared to \cite{khosla2020supervised}, where the authors train the output features of a projection head, then discard this projection head at inference time). We use this projection head when studying NCC test accuracy. In particular, we add a 2 layer non-linear MLP with with and without ReLU to report the test accuracies provided in Tab.~\ref{tab:cifar100_main}, Tab.~\ref{tab:relu_noRelu_all_test} and , Tab.~\ref{Tab:NCC_Bindings}. Following \cite{khosla2020supervised}, in addition to the NCC test accuracy on post-projection feature (final output), we evaluate the accuracies on the pre-projection features (without input to projector) as well. We remark that both ResNet and DenseNet architectures include ReLU activations before the final output, which enforces a non-negativity on the learned embeddings. Resnet-18 models with CIFAR10, MNIST have been trained for 350 epochs with a constant learning rate of $0.1$, no weight decay, batch size of $1024$, and SCL temperature parameter $\tau = 0.1$ (consistent with the choice of $\tau$ made in \cite{khosla2020supervised,graf2021dissecting}). For CIFAR100 and Tiny-ImageNet experiments, we train a ResNet-34 under a similar setup, for 500 epochs, (with batch binding in Fig.~\ref{Fig:Deepnet_SCL_Geom_GMu}). For all experiments with other models or datasets, we provide experimental details in the following sections. All ResNet-18 and DenseNet models have been trained on a single Tesla V100 GPU machine and all ResNet-34 models were trained on a single machine with 2-4 Tesla V100 GPU, depending on experiment.

\subsection{Details on Fig.~\ref{Fig:Deepnet_SCL_Geom_GMu} Heatmaps}\label{sec:exp_fig1_detail}
Since this particular experiment provides the most insight into the geometric behaviour of features, we have provided further explanation of the experimental setup and method. ResNet-18 models were trained for 350 epochs on MNIST using \textbf{CE}, \textbf{SCL} and \textbf{SCL + ReLU} under difference imbalance ratios. For SCL and SCL + ReLU we extract the features, calculate the class means and plot a heatmap of the gram matrices $G_{M}$. For CE, following previous work by \cite{papyan2018full, seli, zhu2021geometric} we center the class means by the global mean, ie $\overline{\mu_{c}} = \mu_{c} - \frac{1}{K} \sum_{i=1}^K \mu_{i}$ and then proceed to calculate $G_{M}$. In order to have a fair comparison, we normalize all gram matrices by dividing by the matrix maximum $\overline{G_{M}} = \frac{G_{M}}{\text{max}_{i,j} |G_{\mu}^{i,j}| }$. No projection layer was implemented for these results. The heatmaps were included to provide a clear visual, emphasizing the impact of ReLU on achieving symmetric geometry under imbalance. Each cell $[i,j]$ of the $G_{M}$ matrix represents the inner product $\mu_i^T \mu_j$, rather a normalized value representing the angle between the two class centers. In addition, as SCL includes a normalization layer, the diagonals $G_{M}[i,i]$ would be closer to zero as the features exhibit stronger neural collapse properties.






\subsection{Additional geometric analysis}

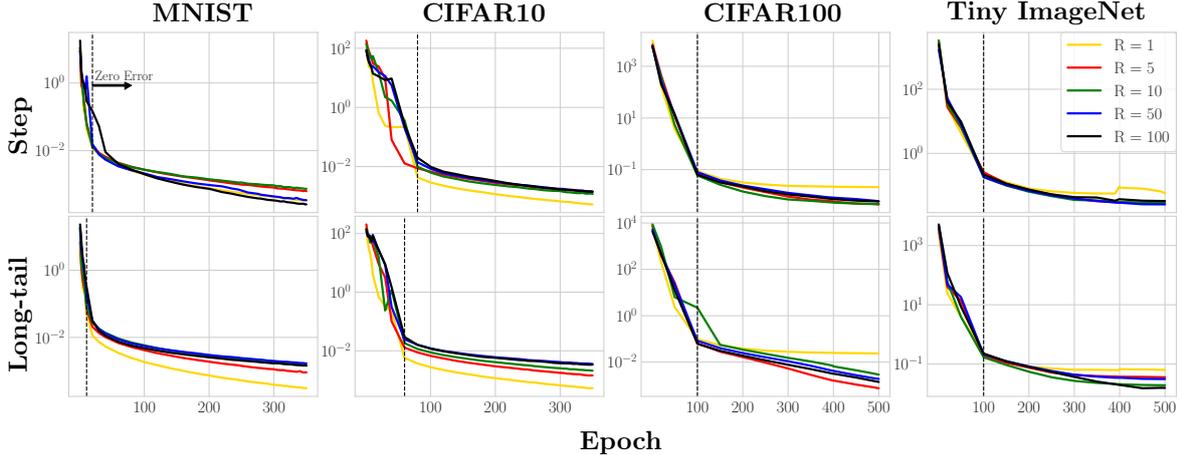
\begin{figure}[t]
\begin{subfigure}[b]{0.95\textwidth}
    \centering
    \begin{tikzpicture}
        \node at (0.0,0) 
        {\includegraphics[width=\textwidth]{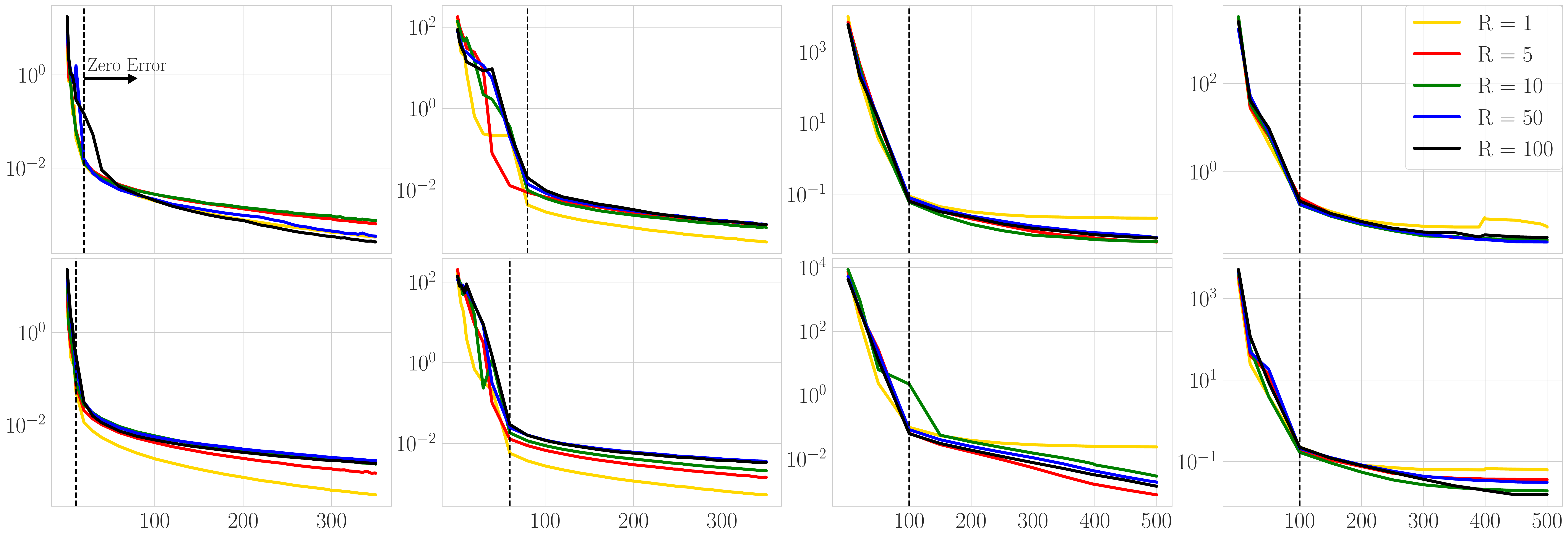}};
        \node at (-5.3,2.8) [scale=0.9]{\textbf{MNIST}};
        \node at (-1.6,2.8) [scale=0.9]{\textbf{CIFAR10}};
        \node at (2.2,2.8) [scale=0.9]{\textbf{CIFAR100}};
        \node at (5.8,2.8) [scale=0.9]{\textbf{Tiny ImageNet}};
        \node at (-7.7,1.3)  [scale=0.9, rotate=90]{\textbf{Step}};
        \node at (-7.7,-1.2)  [scale=0.9, rotate=90]{\textbf{Long-tail}};
        \node at (0.2,-2.9) [scale=0.9]{\textbf{Epoch}};
    \end{tikzpicture}
\end{subfigure}
\captionsetup{width=0.95\linewidth}
\caption
{Neural Collapse metric $\beta_\text{NC}:=\tr(\Sigma_W\Sigma_B^\dagger)/k$ for the corresponding experiments in Fig.~\ref{Fig:Deepnet_SCL_Geom_GMu}. Values are usually on the order of $10^{-2}$ suggesting strong convergence of embeddings to their class means.}
\label{fig:Deepnet_SCL_Geom_NC}
\end{figure}

     In addition to analyzing the convergence of the Gram-matrix of class-mean embeddings $\G_\M$ to the OF geometry (as provided in Fig.~\ref{Fig:Deepnet_SCL_Geom_GMu}), we also keep track of Neural Collapse (Defn.~\ref{def:nc}) of individual embeddings and orthogonality of their class-means (Defn.~\ref{def:of}) separately. Furthermore, we qualitatively study the Gram matrices $\G_\M$ and $\G_\Hb$  and compare them to corresponding matrices for the CE loss under imbalances. 
     
     \subsubsection{Neural Collapse}~
     In order to quantify the collapse (NC) of embeddings, $\h_i, i \in [n]$, to their class-means,  we measure $\beta_\text{NC}:=\tr(\Sigma_W\Sigma_B^\dagger)/k$ \citep{NC}. Here, $\Sigma_B= \sum_{c\in [k]}(\boldsymbol{\mu}_c - \boldsymbol{\mu}_{G})(\boldsymbol{\mu}_c - \boldsymbol{\mu}_{G})^\top$ is the between class covariance matrix, $\boldsymbol{\mu}_G = \frac{1}{k}\sum_{c\in[k]}\boldsymbol{\mu}_c$ is the  global mean, and $\Sigma_W  = \sum_{i\in [n]}(\h_i - \boldsymbol{\mu}_{y_i})(\h_i - \boldsymbol{\mu}_{y_i})^\top$ is the within class covariance matrix.

     For the experiments shown in Fig.~\ref{Fig:Deepnet_SCL_Geom_GMu} and Fig.~\ref{fig:loss_convergence} in the main body, the corresponding values of $\beta_\text{NC}$ can be found in Fig.~\ref{fig:Deepnet_SCL_Geom_NC} and Fig.~\ref{fig:loss NC and angle convergence SM}(b) respectively.

    \subsubsection{Angular convergence}
\begin{figure}[h]
    \begin{subfigure}[b]{\textwidth}
        \centering
        \begin{tikzpicture}
            \node at (0.0,0) 
            {\includegraphics[width=0.85\textwidth]{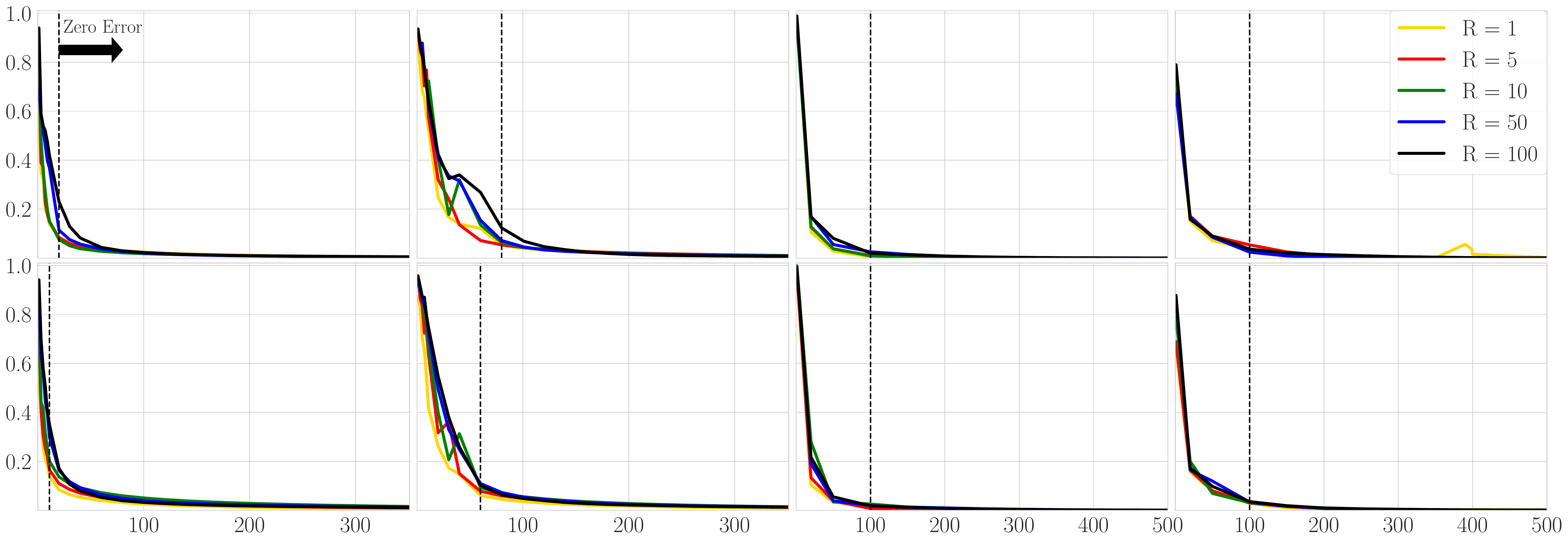}};
            \node at (-4.7,2.6) [scale=0.9]{\textbf{MNIST}};
            \node at (-1.4,2.6) [scale=0.9]{\textbf{CIFAR10}};
            \node at (1.8,2.6) [scale=0.9]{\textbf{CIFAR100}};
            \node at (5.0,2.6) [scale=0.9]{\textbf{Tiny ImageNet}};
            \node at (-7,1.2)  [scale=0.9, rotate=90]{\textbf{Step}};
            \node at (-7,-1)  [scale=0.9, rotate=90]{\textbf{Long-tail}};
            \node at (0.0,-2.6) [scale=0.9]{\textbf{Epoch}};
        \end{tikzpicture}
    \end{subfigure}
    \caption
    {Average cosine similarity between different class means $\alpha_{\text{sim}}(c,c') := \nicefrac{\mub_c^\top\mub_{c'}}{\|\mub_c\|\|\mub_{c'}\|}$ for corresponding results from Fig.~\ref{Fig:Deepnet_SCL_Geom_GMu}. Final values are mostly on the order of $10^{-2}$, indicating strong orthogonality between class-mean embeddings.}
    \label{fig:Deepnet_SCL_Geom_angles}
\end{figure}
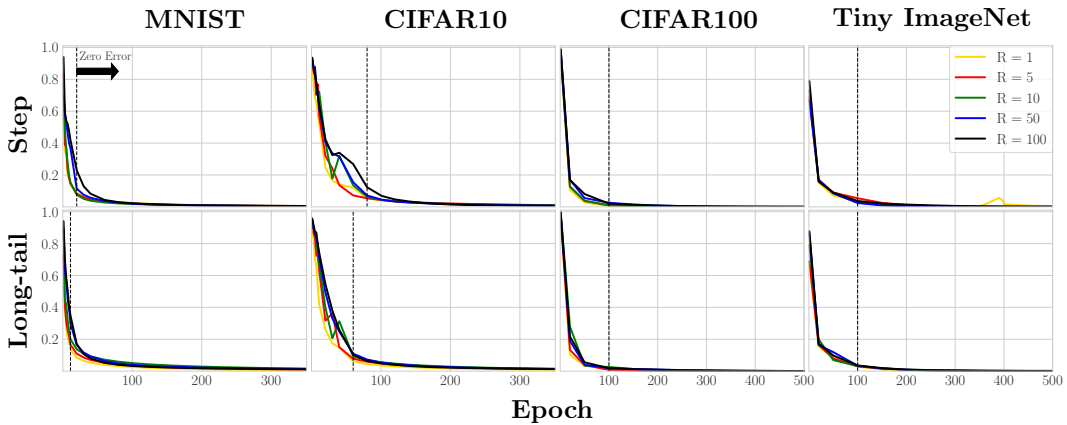

~
    Having confirmed convergence of embeddings to their respective class-means via NC ($\h_i \approx \mub_c $ for $i:y_i = c$), we can now compare the feature geometry to the OF geometry by calculating the average $\alpha_{\text{sim}}(c,c') := \nicefrac{\mub_c^\top\mub_{c'}}{\|\mub_c\|\|\mub_{c'}\|}$ between each pair of classes. Fig.~\ref{fig:Deepnet_SCL_Geom_angles} plots the average cosine similarity $\text{Ave}_{c\neq c'}\alpha_{\text{sim}}(c,c')$ between class means for the same experiment as that 
 of Fig.~\ref{Fig:Deepnet_SCL_Geom_GMu}. The graphs indicate strong convergence to orthogonality between feature representations of different classes. Similarly, results for angular convergence corresponding to Fig.~\ref{fig:loss_convergence} are provided in Fig.~\ref{fig:loss NC and angle convergence SM}(c), indicating similar convergence to OF for the full-batch experiments.

     \subsubsection{Embedding heatmaps}~
     As a qualitative measure, we have generated heatmaps  that visually represent the learned embedding geometries; see Figs.~\ref{fig:heatmap}, \ref{fig:GM matrices comparison_old}, \ref{fig:GH matrices}. Specifically, we generate heatmaps of the Gram-matrices $\G_\M = \M^\top\M$ and $\G_\Hb = \Hb^\top\Hb$. In Fig.~\ref{fig:heatmap} we train ResNet-18 with the full MNIST dataset. In Fig.~\ref{fig:GM matrices comparison_old} we run on a subset of the dataset ($n = 10000$ total examples) with a batch size of 1000. Specifically for Fig.~\ref{fig:GM matrices comparison_old}, when optimizing with CE loss we modify the network (ResNet-18) such that it has a normalization layer before the linear classifier head. We consider this additional setting to allow for a comparison where CE features are constrained to the unit sphere akin to our SCL experiments. Lastly, in Fig.~\ref{fig:GH matrices} we plot the learned features Gram-matrix $\G_\Hb$ for a ResNet-18 trained on CIFAR10 ($n = 10000$ total examples) with a batch size of $1000$.\footnote{In both Fig.~\ref{fig:GH matrices} and Fig.~\ref{fig:GM matrices comparison_old} no batch duplication is used as described in Sec.~\ref{sec:Deepnet_SCLGeom_exp}.} This heatmap qualitatively shows a more complete picture as we are plotting $\G_\Hb = \Hb^\top \Hb$ rather than $\G_\M$, thus simultaneously illustrating both Neural Collapse and convergence to the k-OF structure.  

\subsubsection{Experiments with MLPs}~
{In Fig.~\ref{fig:MLP convergence} we run experiments with a simple 6 layer multilayer perceptron (MLP) to further explore the impact of model complexity on geometric convergence. The MLP includes batch normalization and ReLU activation between each layer. Each layer has  512 hidden units. We train the model with a batch size of 1000 with random reshuffling at each epoch. Furthermore, we train under $R$-STEP imbalanced MNIST. No batch duplication was used. All other aspects of the implementation are as described in Sec.~\ref{sec:exp_details_SM}. As shown in Fig.~\ref{fig:MLP convergence} all metrics $\Delta_{\G_\M}$, $\beta_{\text{NC}}$, and $\text{Ave}_{c \neq c'}\text{ }\alpha_{\text{sim}(c,c')}$ indicate strong convergence to the OF geometry, irrespective of imbalance ratio $R$.}

    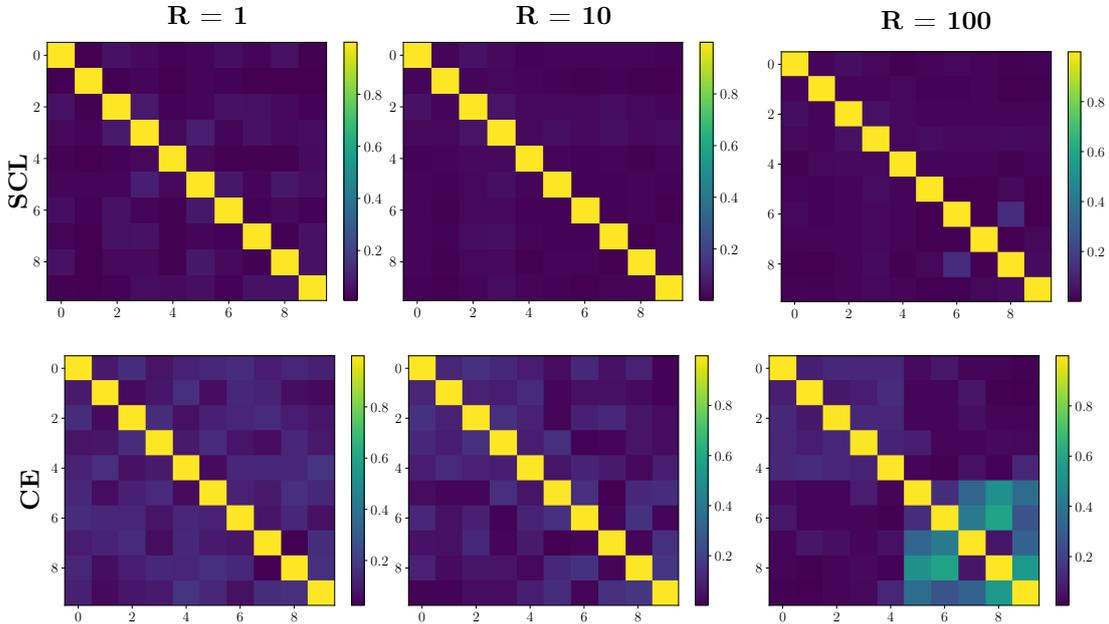
\begin{figure*}[h]
        \centering
        \begin{subfigure}[b]{0.3\textwidth}
            \centering
            \begin{tikzpicture}
                \node at (0.0,0) 
                {\includegraphics[width=\textwidth]{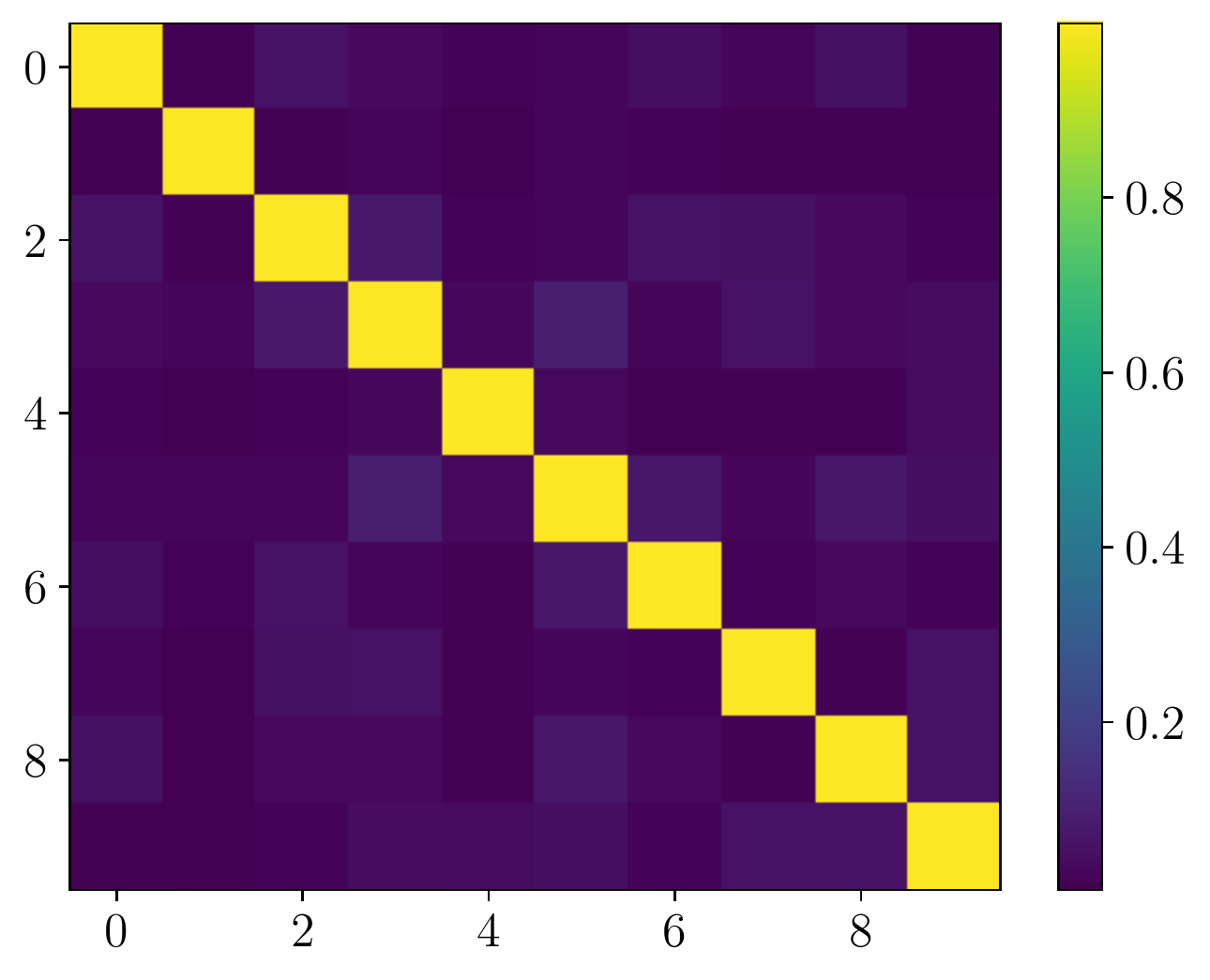}};
    			\node at (0.0,2.2) [scale=0.9]{\textbf{R = 1}};
                \node at (-2.5,0.0)  [scale=0.9, rotate=90]{\textbf{SCL}};
            \end{tikzpicture}
        \end{subfigure}
        \begin{subfigure}[b]{0.3\textwidth}
            \centering
            \begin{tikzpicture}
                \node at (0,0) 
                {\includegraphics[width=\textwidth]{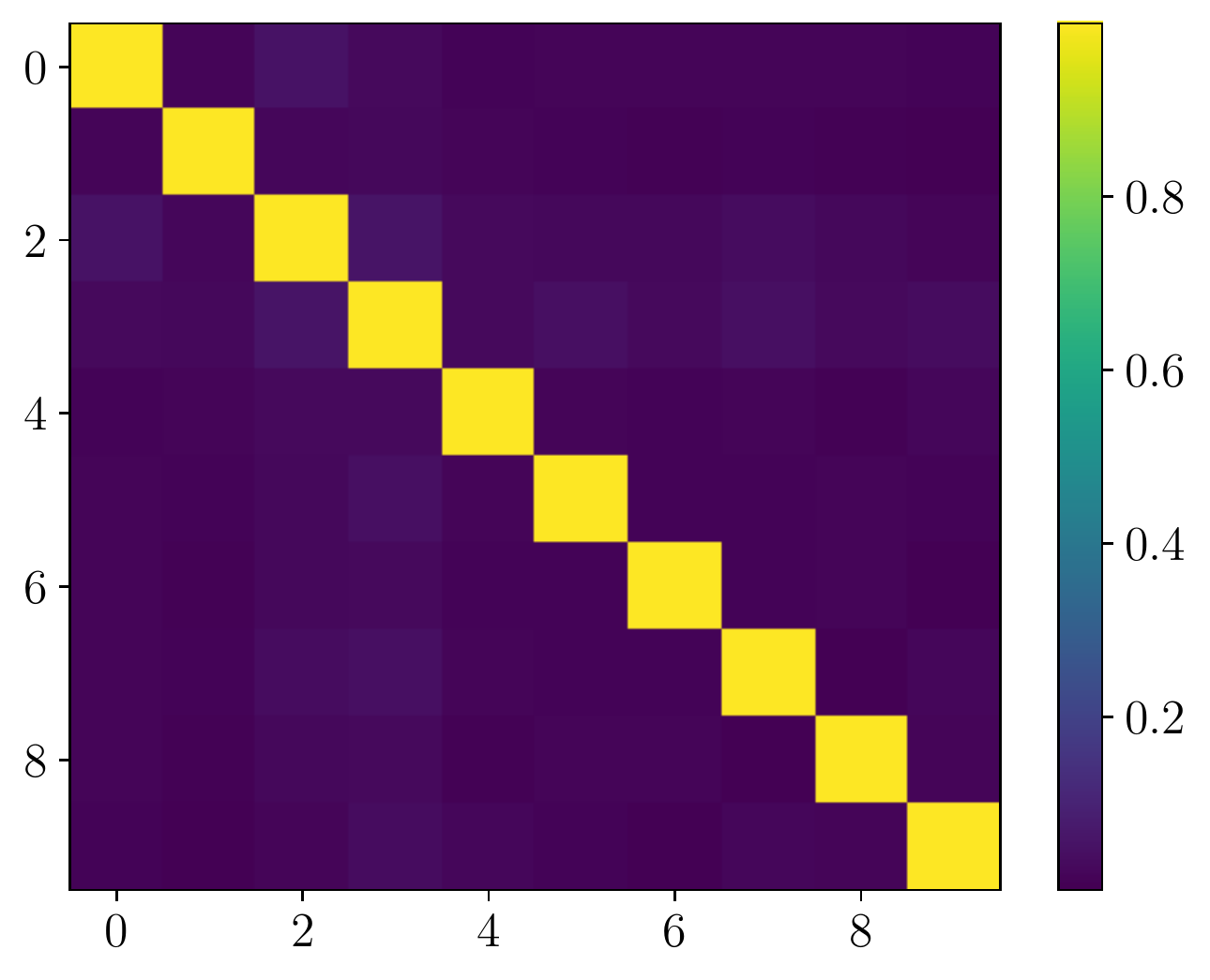}};
    			\node at (0.0,2.2) [scale=0.9]{\textbf{R = 10}};
               \node at (-2.4,0.0)  [scale=0.9, rotate=90]{};
            \end{tikzpicture}
        \end{subfigure}
        \begin{subfigure}[b]{0.29\textwidth}
            \centering
            \begin{tikzpicture}
                \node at (0,0) 
                {\includegraphics[width=\textwidth]{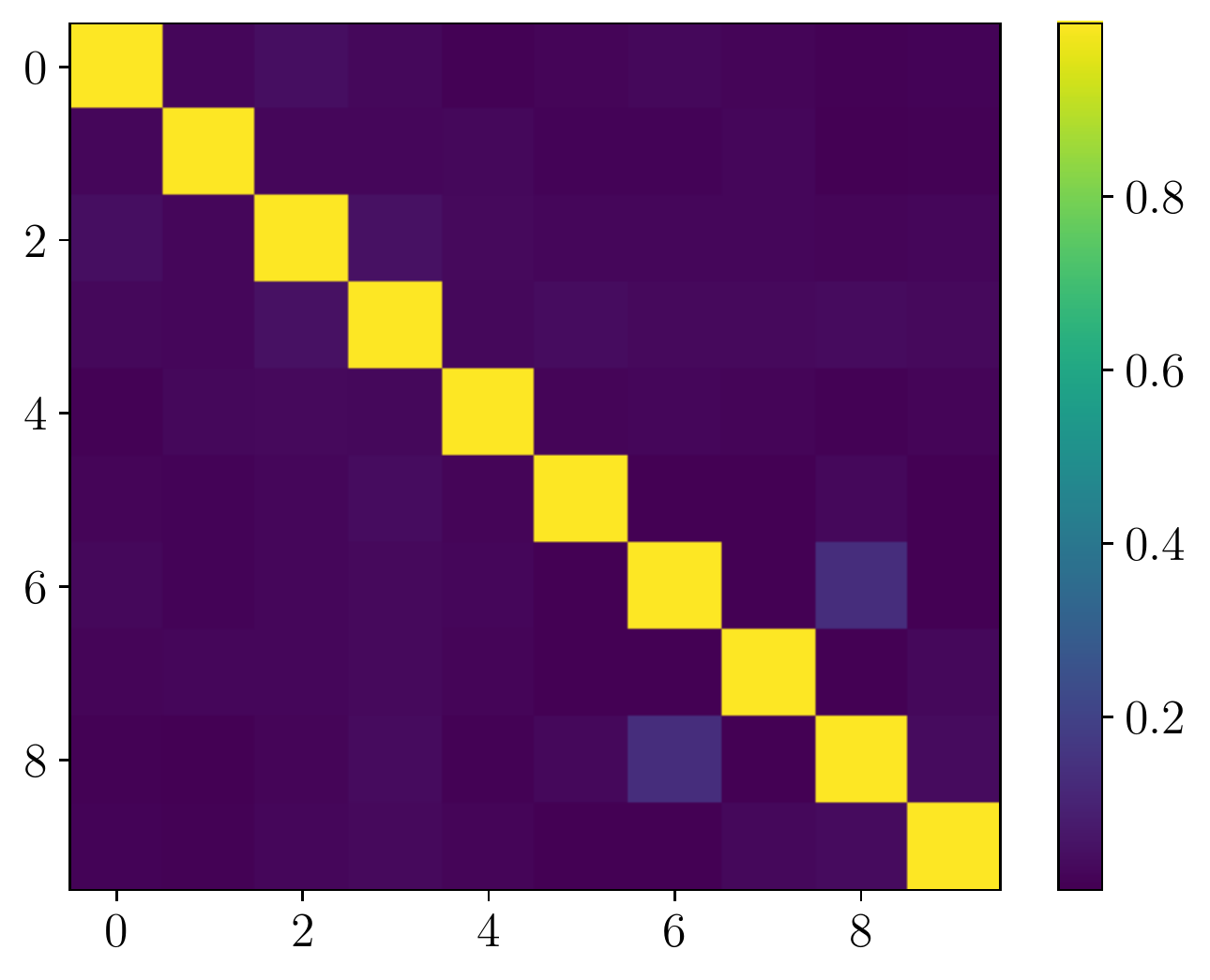}};
    			\node at (0.0,2.2) [scale=0.9]{\textbf{R = 100}};
                \node at (-2.4,0.0)  [scale=0.9, rotate=90]{};
            \end{tikzpicture}
        \end{subfigure}
        \begin{subfigure}[b]{0.29\textwidth}
            \centering
            \begin{tikzpicture}
                \node at (0.0,0) 
                {\includegraphics[width=\textwidth]{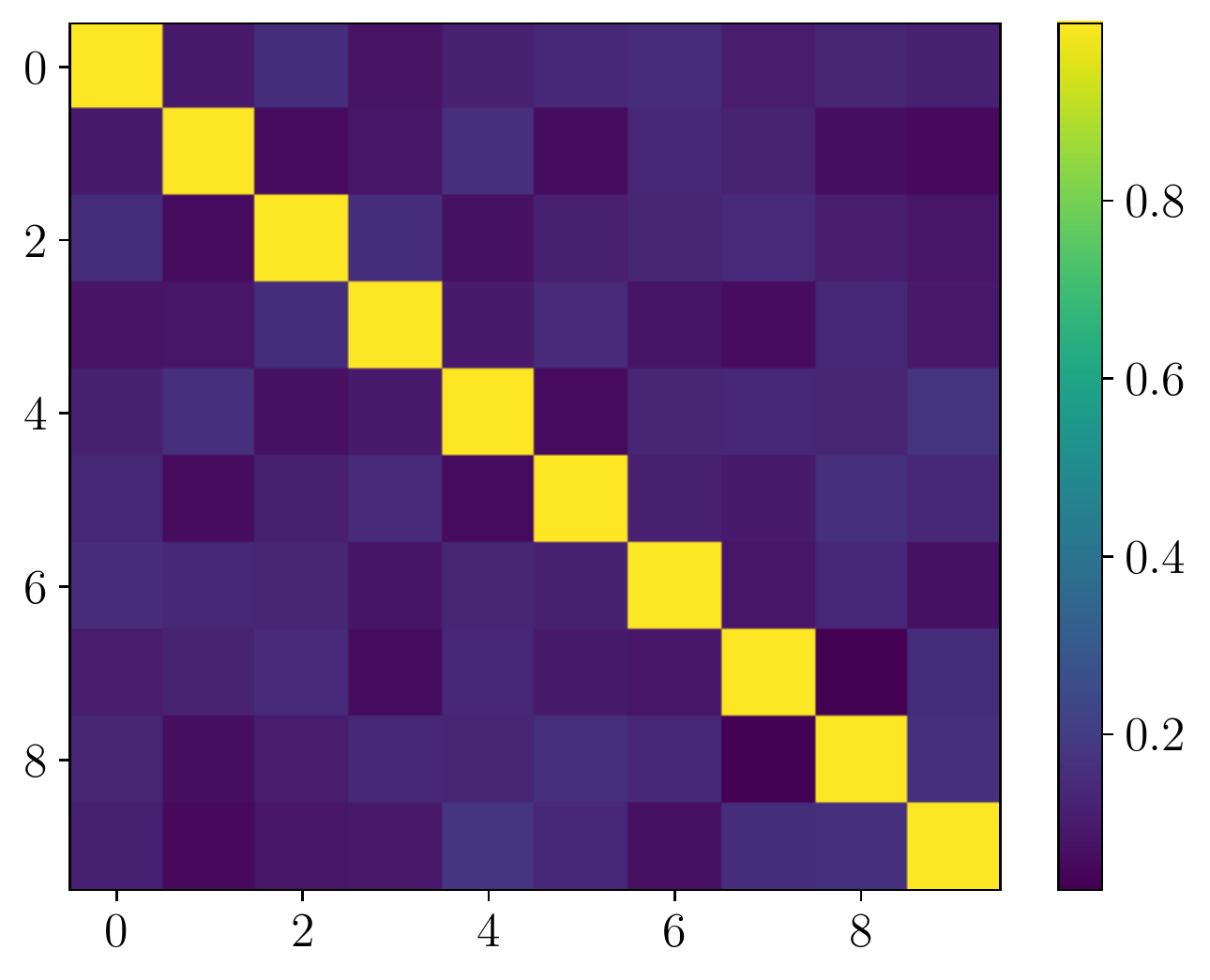}};
    			\node at (0.0,1.8) [scale=0.9]{\textbf{}};
                \node at (-2.5,0.0)  [scale=0.9, rotate=90]{\textbf{CE}};
            \end{tikzpicture}
        \end{subfigure}
        \begin{subfigure}[b]{0.29\textwidth}
            \centering
            \begin{tikzpicture}
                \node at (0,0) 
                {\includegraphics[width=\textwidth]{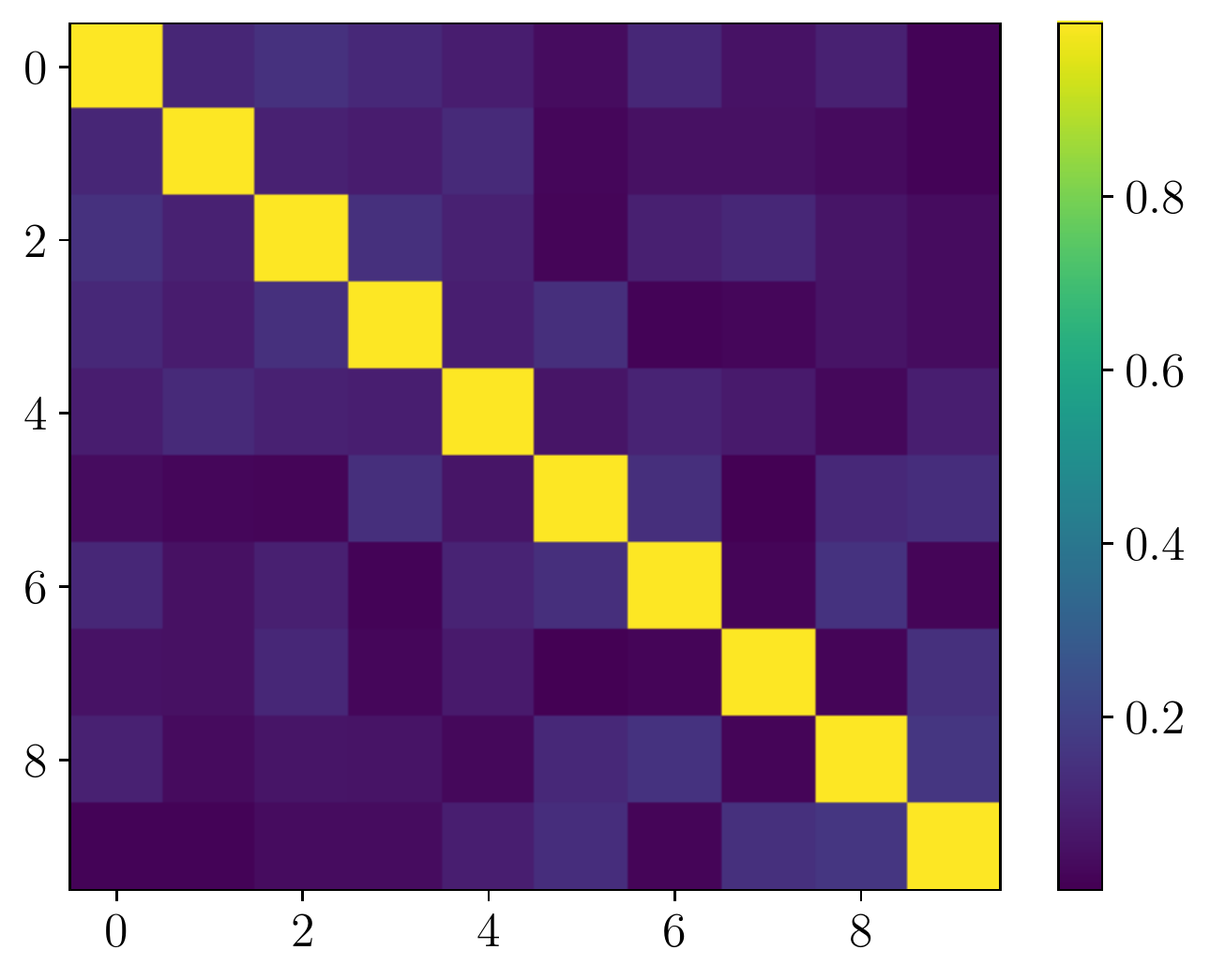}};
    			\node at (0.0,1.8) [scale=0.9]{\textbf{}};
               \node at (-2.4,0.0)  [scale=0.9, rotate=90]{};
            \end{tikzpicture}
        \end{subfigure}
        \begin{subfigure}[b]{0.29\textwidth}
            \centering
            \begin{tikzpicture}
                \node at (0,0) 
                {\includegraphics[width=\textwidth]{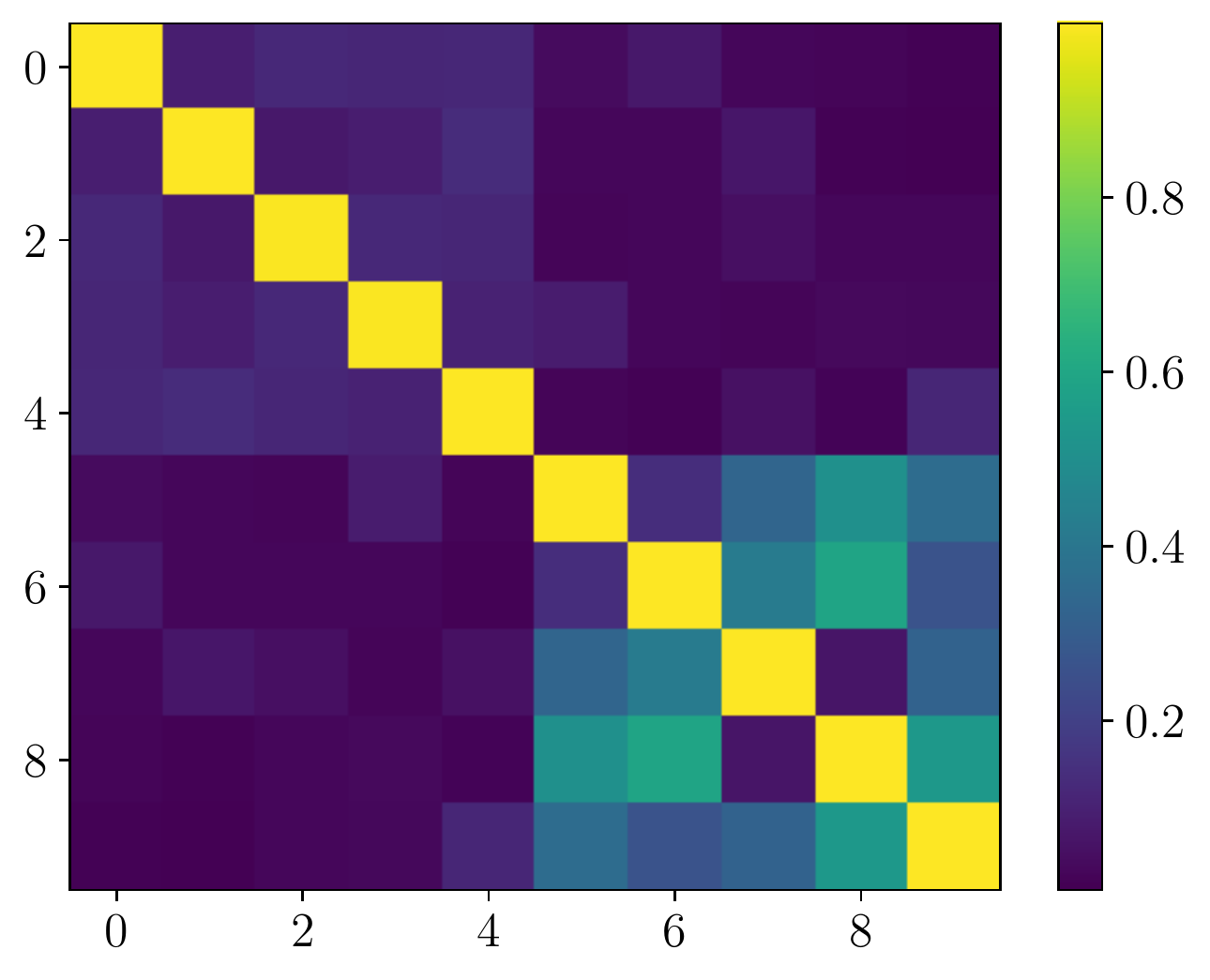}};
    			\node at (0.0,1.8) [scale=0.9]{\textbf{}};
                \node at (-2.4,0.0)  [scale=0.9, rotate=90]{};
            \end{tikzpicture}
        \end{subfigure}
        \caption[GM]
        {$\G_\M$ Gram-matrices of mean-embeddings for various $R$-STEP imbalances at last epoch (350) of training with ResNet-18 on MNIST with $n = 10000$. To allow for fair comparison, the CE features are normalized before the classifier head akin to the SCL experiments.} 
        \label{fig:GM matrices comparison_old}
    \end{figure*}
  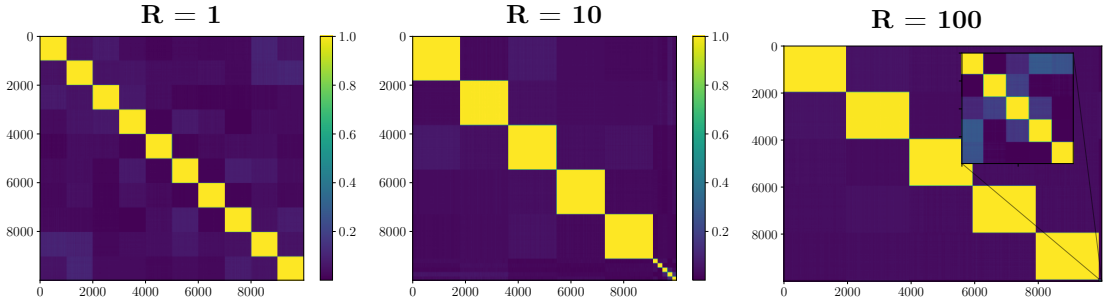
\begin{figure*}[h]
        \centering
        \hspace{-0.5in}
        \begin{subfigure}[b]{0.3\textwidth}
            \centering
            \begin{tikzpicture}
                \node at (0.0,0) 
                {\includegraphics[width=\textwidth]{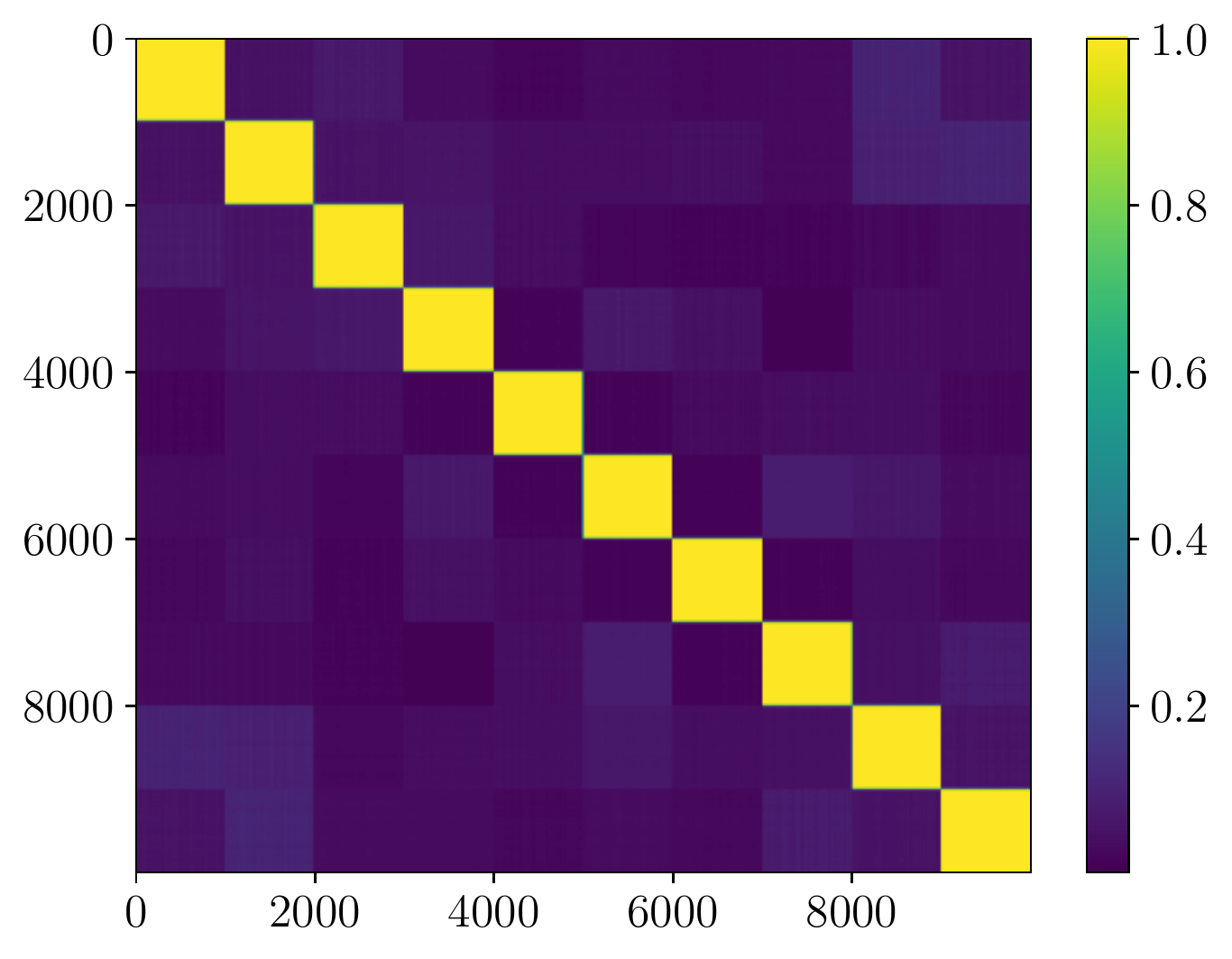}};
    			\node at (0.0,2) [scale=0.9]{\textbf{R = 1}};
                \node at (-2.7,0.0)  [scale=0.7, rotate=90]{};
            \end{tikzpicture}
        \end{subfigure}
        \begin{subfigure}[b]{0.3\textwidth}
            \centering
            \begin{tikzpicture}
                \node at (0,0) 
                {\includegraphics[width=\textwidth]{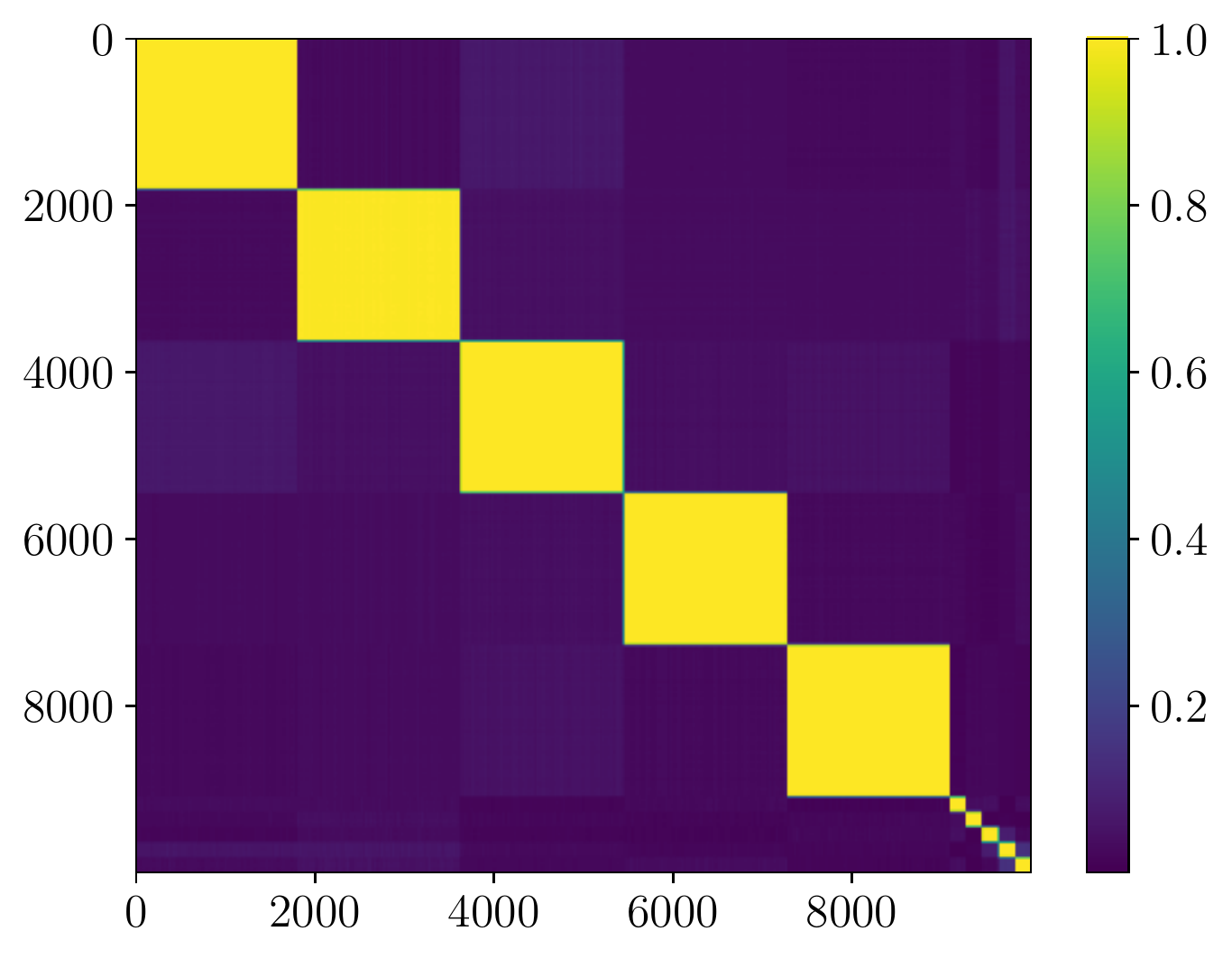}};
    			\node at (0.0,2) [scale=0.9]{\textbf{R = 10}};
               \node at (-2.7,0.0)  [scale=0.7, rotate=90]{};
            \end{tikzpicture}
        \end{subfigure}
        \begin{subfigure}[b]{0.3\textwidth}
            \centering
            \begin{tikzpicture}
                \node at (0,0) 
                {\includegraphics[width=\textwidth]{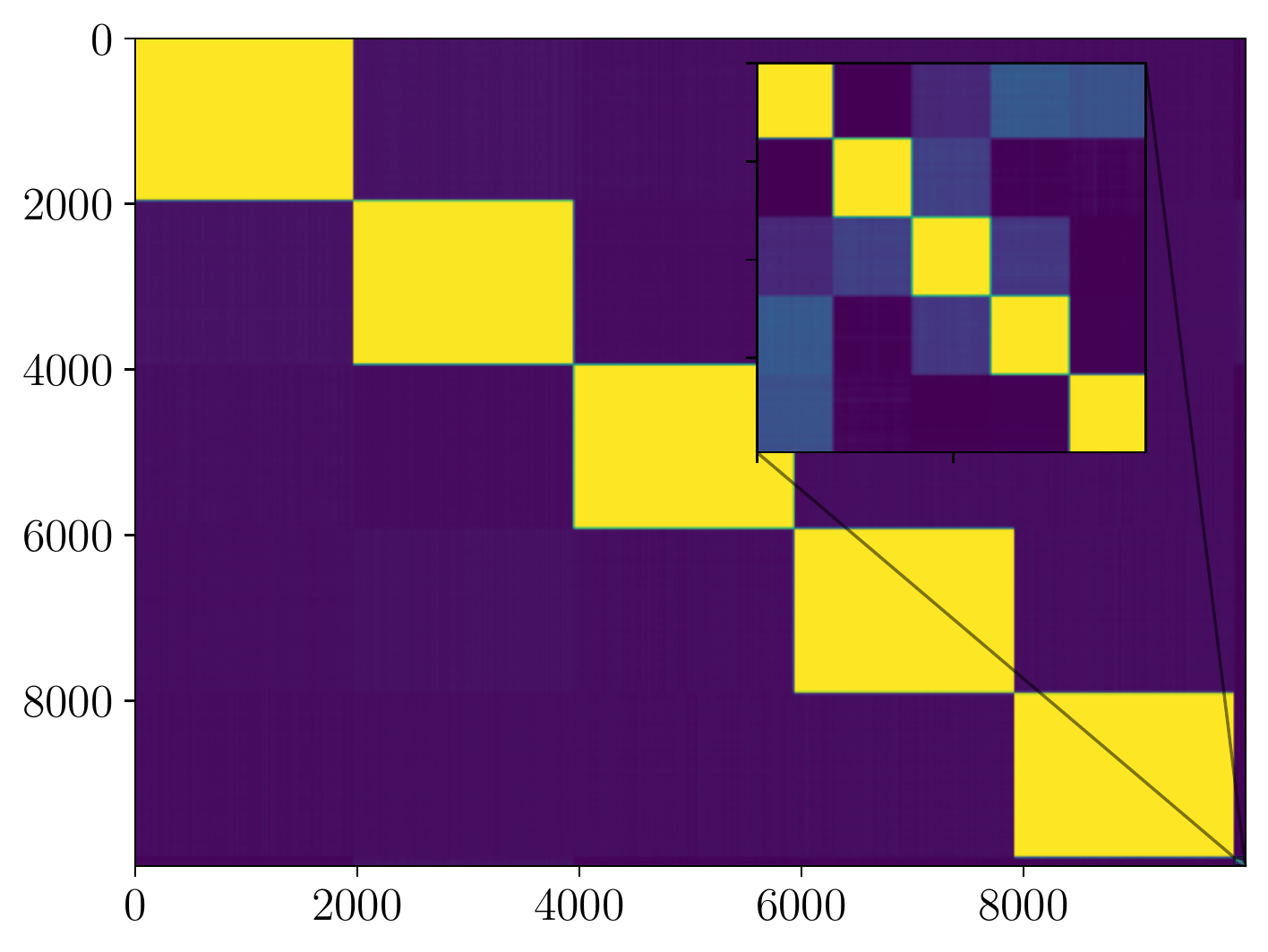}};
    			\node at (0.0,2) [scale=0.9]{\textbf{R = 100}};
                \node at (-2.7,0.0)  [scale=0.7, rotate=90]{};
            \end{tikzpicture}
        \end{subfigure}
        \caption[GH]
        {$\G_\Hb$ Gram-matrices of feature embeddings for various imbalances at last epoch (350) of training SCL with CIFAR10 and ResNet.} 
        \label{fig:GH matrices}
    \end{figure*}

\subsection{Optimization dynamics}

\subsubsection{Loss convergence}
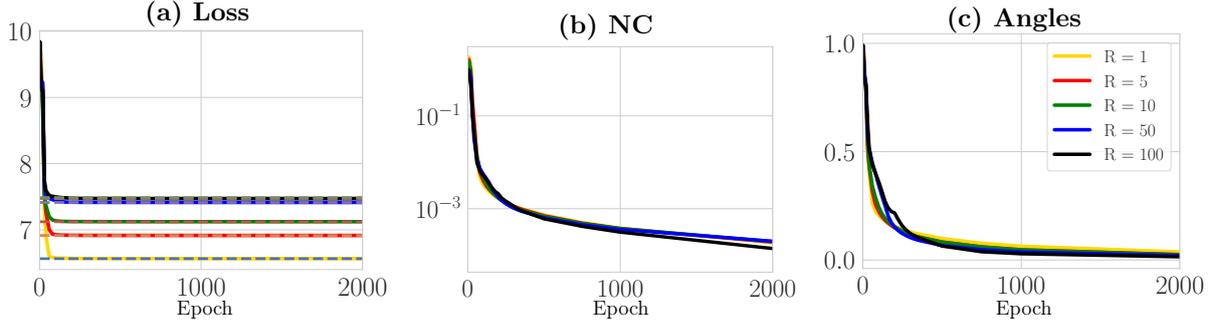
\begin{figure*}
        \centering
        \begin{subfigure}{0.325\textwidth}
            \centering
            \begin{tikzpicture}
                \node at (0.0,0) 
                {\includegraphics[width=\textwidth]{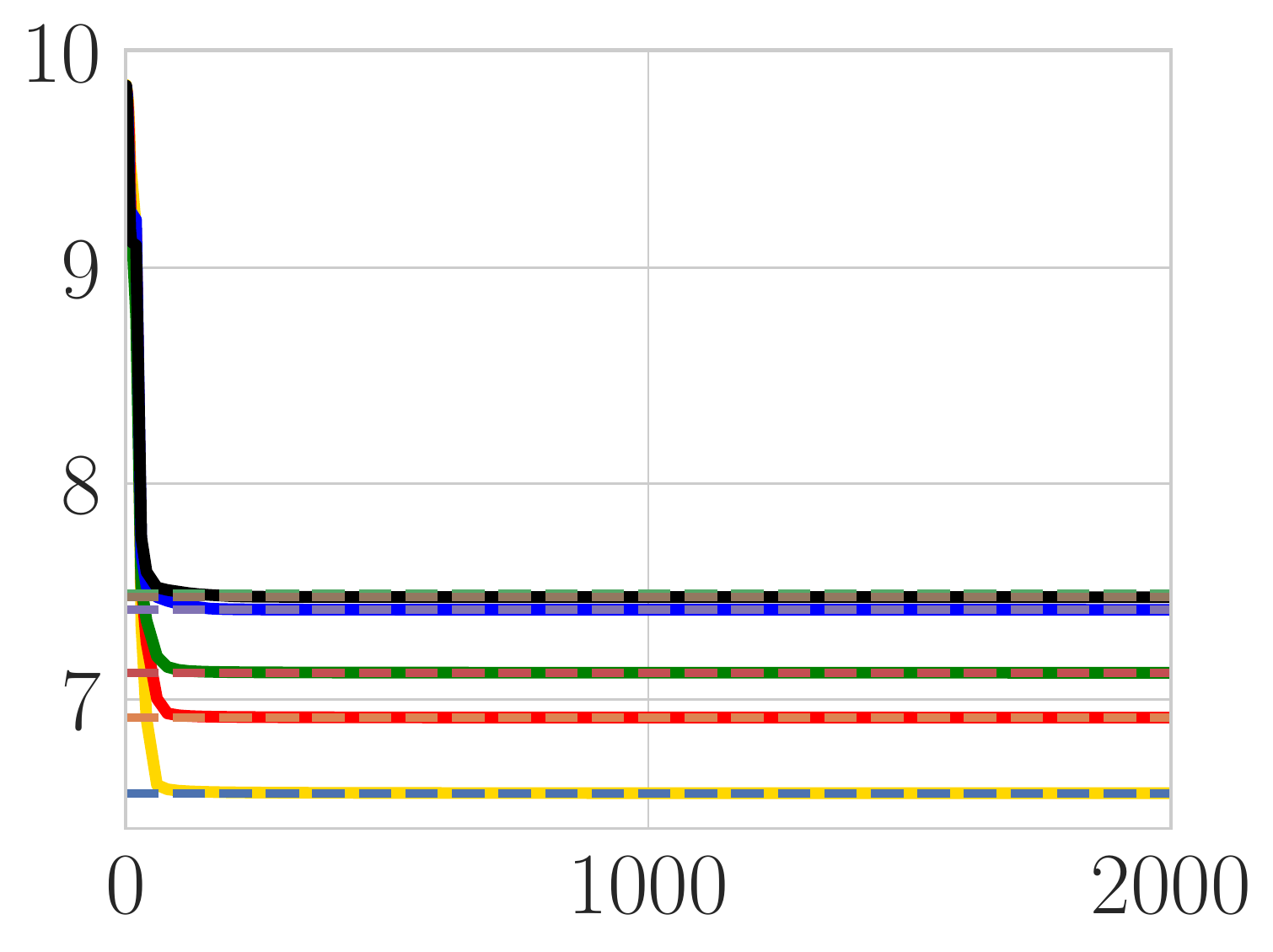}};
                 \node at (0.1,-1.97) [scale=0.7]{Epoch};
    			\node at (0.0,1.95) [scale=0.9]{\textbf{(a) Loss}};
            \end{tikzpicture}
        \end{subfigure}
        \hfill
        \begin{subfigure}{0.325\textwidth}
            \centering
            \begin{tikzpicture}
                \node at (0,0) 
                {\includegraphics[width=\textwidth]{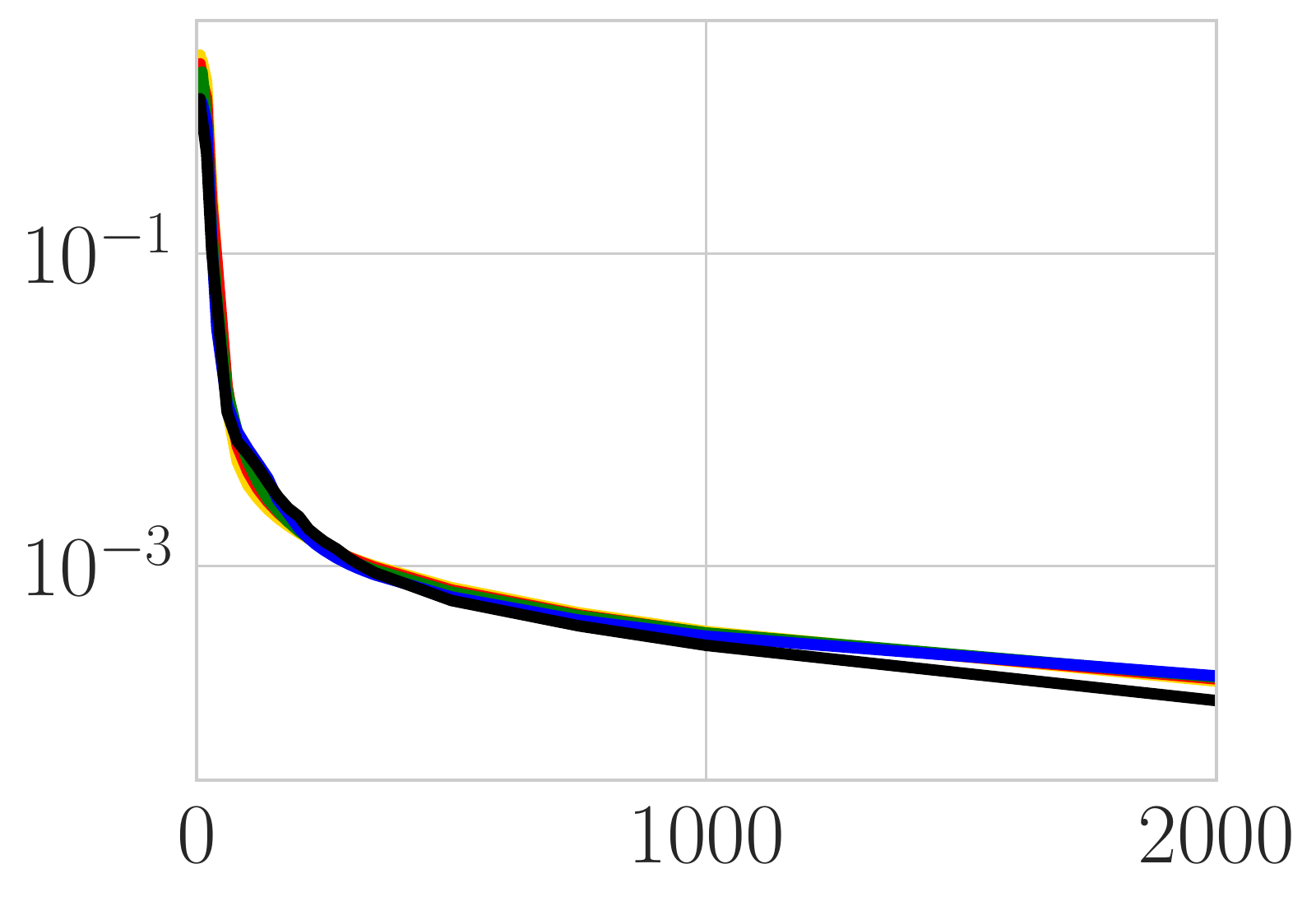}};
                \node at (0.2,-1.8) [scale=0.7]{Epoch};
    			\node at (0.0,1.95) [scale=0.9]{\textbf{(b) NC}};
            \end{tikzpicture}
        \end{subfigure}
        \hfill
        \begin{subfigure}{0.325\textwidth}
            \centering
            \begin{tikzpicture}
                \node at (0,0) 
                {\includegraphics[width=\textwidth]{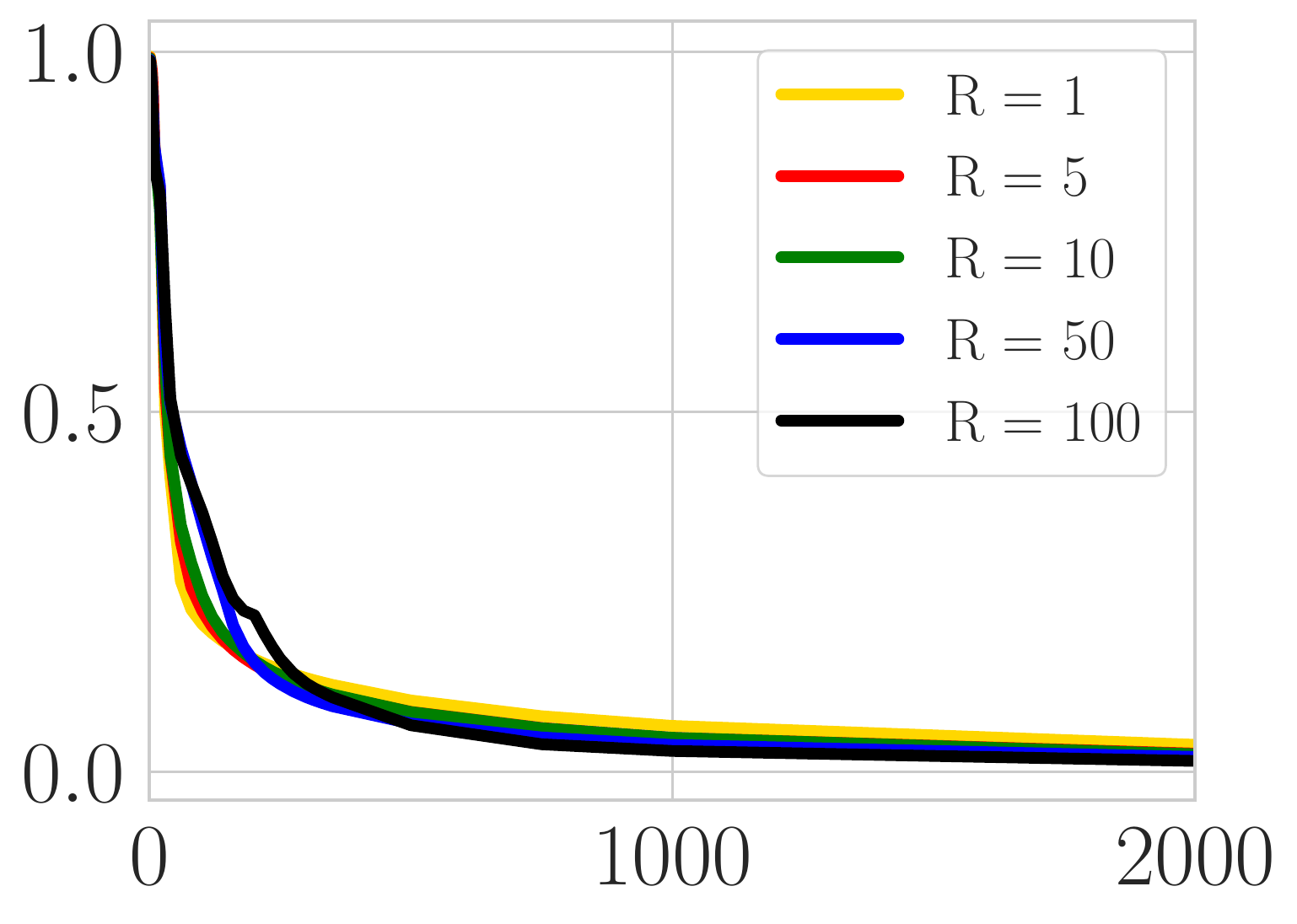}};
                \node at (0.13,-1.9) [scale=0.7]{Epoch};
    			\node at (0.0,1.95) [scale=0.9]{\textbf{(c) Angles}};
            \end{tikzpicture}
        \end{subfigure}
        \captionsetup{width= \linewidth}
        \caption[loss]
        {Full-batch SCL: ResNet-18 trained on a $R$-STEP imbalanced subset of MNIST of size $n=1000$. \textbf{(a)} Loss converges to the lower bound (dashed lines) computed in Thm.~\ref{thm:scl_full}. 
        \textbf{(b)} Within-class feature variation becomes negligible (NC). \textbf{(c)} The average pairwise cosine similarity of class-means approaches zero. Each epoch is equivalent to one iteration of gradient descent.}
        \label{fig:loss NC and angle convergence SM} 
    \end{figure*}

     In order to compute the lower bounds shown in Fig.~\ref{fig:loss_convergence},  we use Thm.~\ref{thm:scl_full}, substituting $e^{-1}$ with $e^{-1/\tau}$ using $\tau = 0.1$ as employed in our experiments; this substitution is allowed thanks to  Remark \ref{rem:tau}. Furthermore, we compute the lower bounds and $\Lc(\Hb)$ on a per sample basis, thus we divide by $n = 1000$ which corresponds to the number of datapoints in our single batch. Lastly, as a method of maintaining numerical stability (as implemented in \cite{khosla2020supervised}) we apply a global scaling of the loss by a factor $\nicefrac{\tau}{\tau_b}$, where $\tau_b = 0.07$ is the base temperature \cite{khosla2020supervised}. The complete experiment, conducted over $2000$ epochs (with axes limited to $500$ epochs for clarity in Figure~\ref{fig:loss_convergence}), is available in Figure~\ref{fig:loss NC and angle convergence SM}.

\subsubsection{Effect of $\tau$}~
\begin{remark}\label{rem:tau}
    The normalization of the embeddings in \ref{eq:SCL_UFM_full} corresponds to SCL with temperature $\tau=1$. More generally, the normalization becomes $\norm{\h_i}^2=1/\tau$. The conclusion of Thm.~\ref{thm:scl_full} is insensitive to the choice of $\tau$, thus stated above for $\tau=1$ without loss of generality. Although the value of $\tau$ does not affect the global optimizers of \ref{eq:SCL_UFM_full}, we have empirically observed that it impacts the speed of convergence during training.
\end{remark}

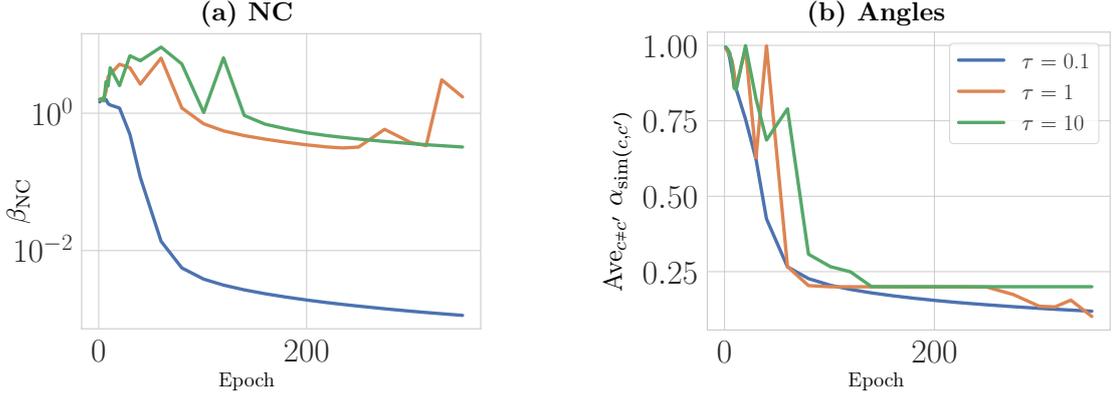
\begin{figure*}[h]
        \centering
        \begin{subfigure}[b]{0.4\textwidth}
            \centering
            \begin{tikzpicture}
                \node at (0.0,0) 
                {\includegraphics[width=\textwidth]{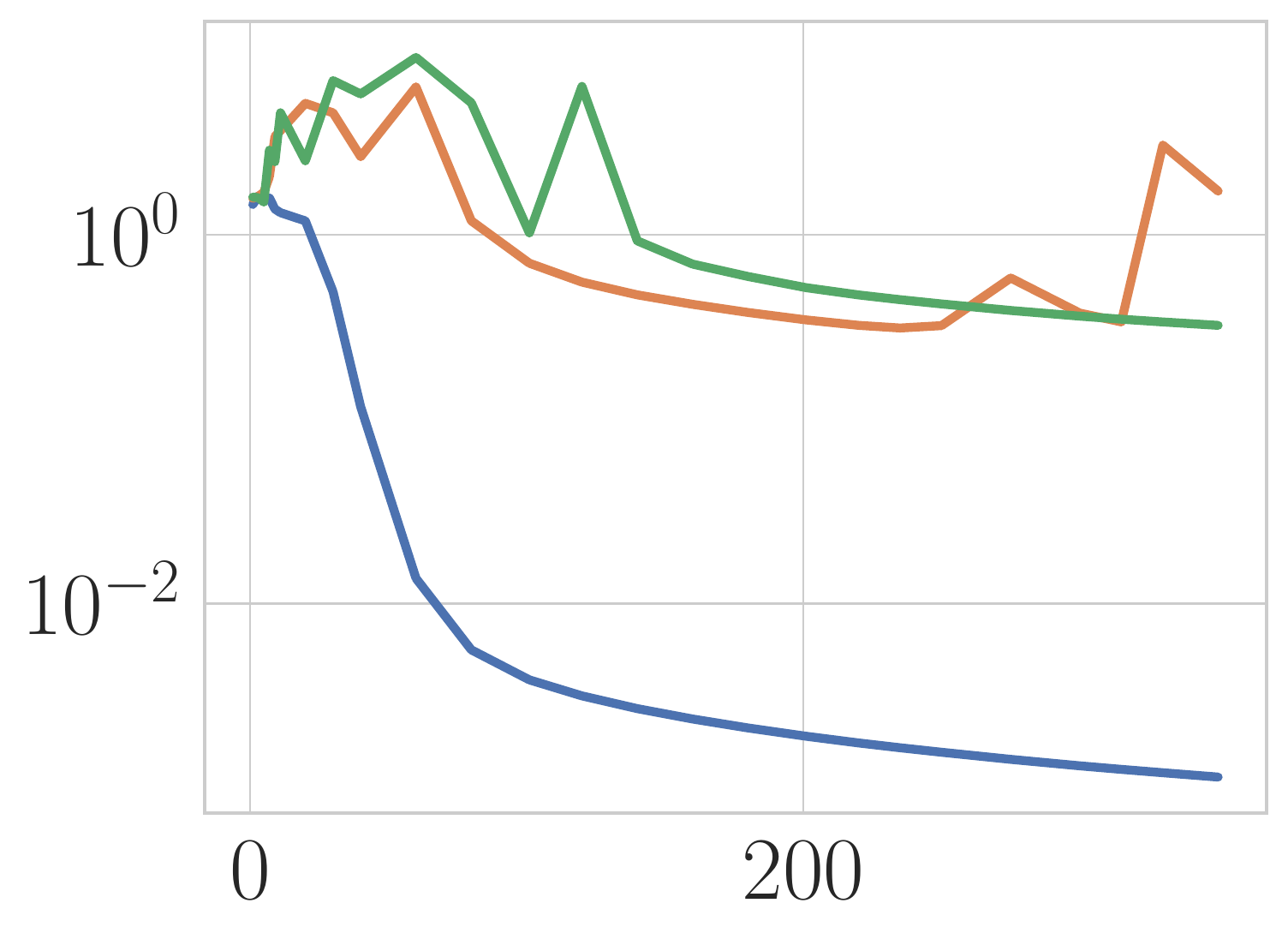}};
                 \node at (0.0,-2.4) [scale=0.7]{Epoch};
    			\node at (0.0,2.47) [scale=0.9]{\textbf{(a) NC}};
                 \node at (-3,0.0)  [scale=0.9, rotate=90]{$\beta_\text{NC}
                 $};
            \end{tikzpicture}
        \end{subfigure}
       \hspace{0.5in}
        \begin{subfigure}[b]{0.4\textwidth}
            \centering
            \begin{tikzpicture}
                \node at (0,0) 
                {\includegraphics[width=\textwidth]{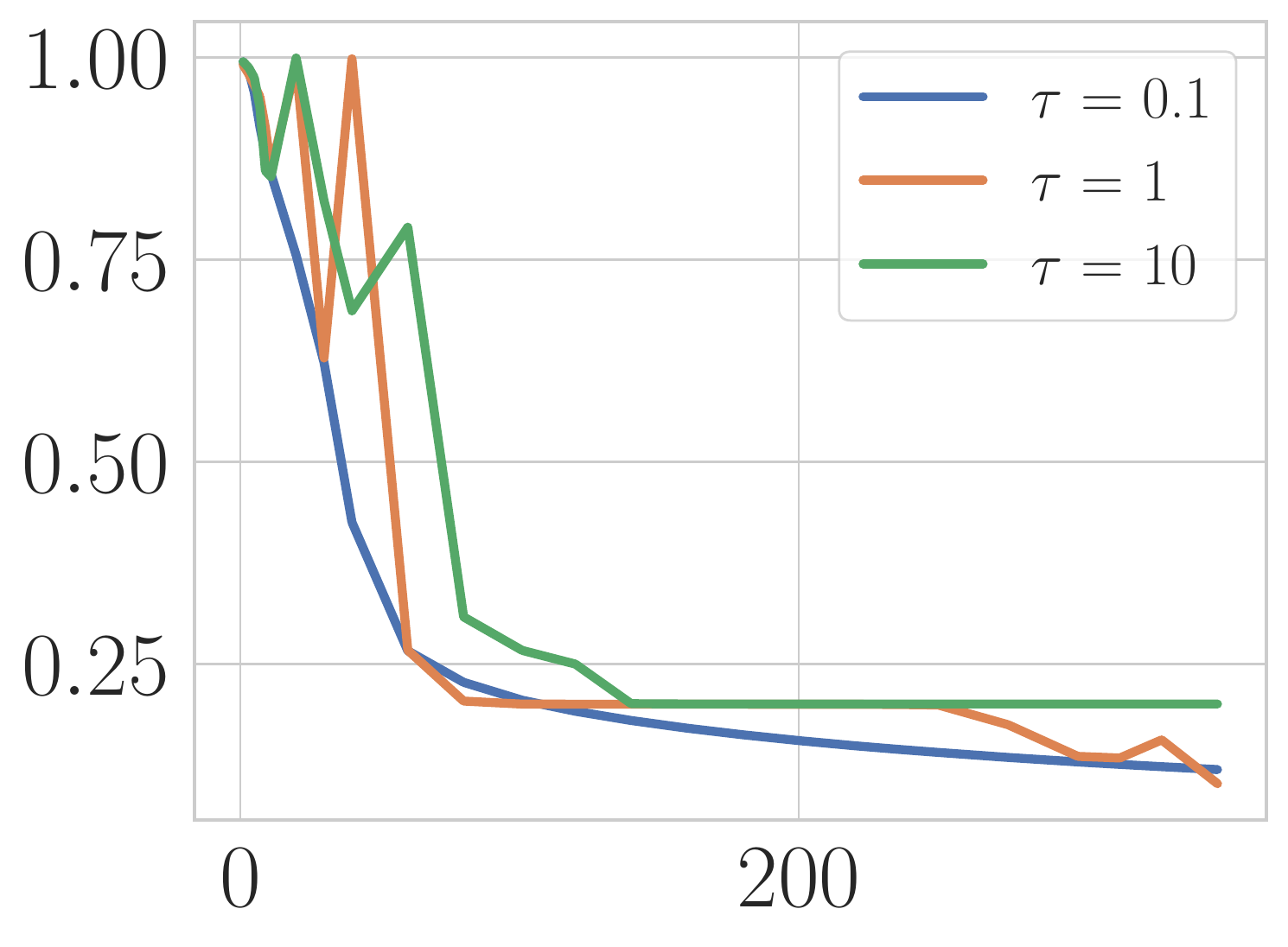}};
                \node at (0.0,-2.4) [scale=0.7]{Epoch};
    			\node at (0.0,2.47) [scale=0.9]{\textbf{(b) Angles}};
               \node at (-3.4,0.0)  [scale=0.9, rotate=90]{$\text{Ave}_{c \neq c'}\text{ }\alpha_{\text{sim}(c,c')}$};
            \end{tikzpicture}
        \end{subfigure}
        \captionsetup{width= \linewidth}
        \caption[loss]
        {Full-batch SCL: ResNet-18 trained on a subset of MNIST of size $n=1000$ with different temperature parameters $\tau$. \textbf{(a)} Within-class feature variation (NC). \textbf{(b)} Average pairwise cosine similarity of class-means. Each epoch is equivalent to one iteration of gradient descent.}
        \label{fig:effect of tau full batch} 
    \end{figure*}

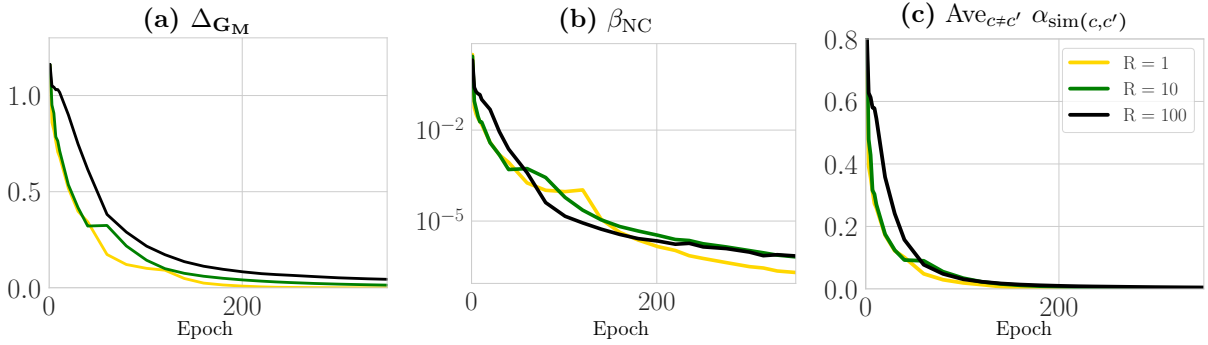
\begin{figure*}
        \centering
        \begin{subfigure}{0.325\textwidth}
            \centering
            \begin{tikzpicture}
                \node at (0.0,0) 
                {\includegraphics[width=\textwidth]{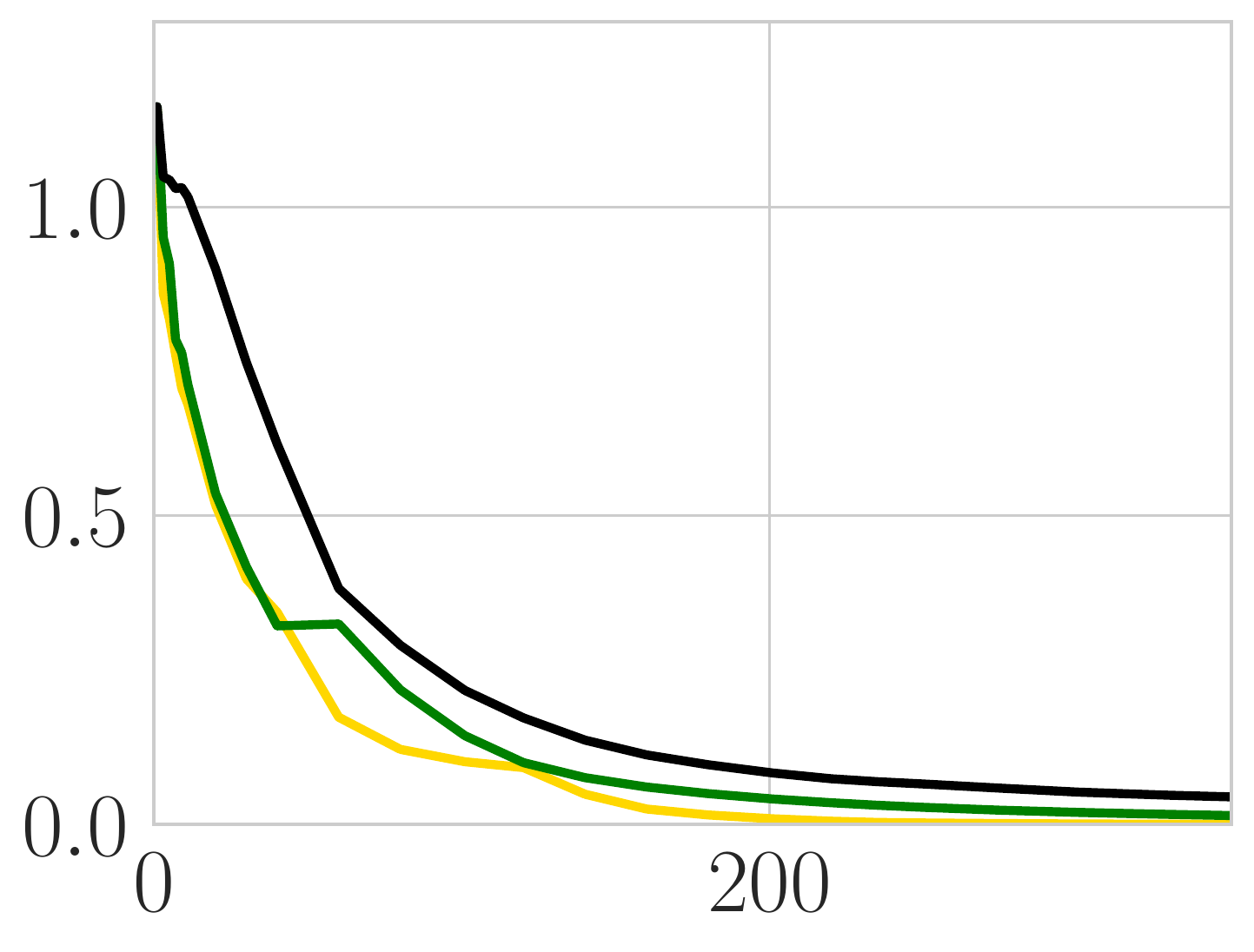}};
                 \node at (0.1,-2) [scale=0.7]{Epoch};
    			\node at (0.0,2.1) [scale=0.9]{\textbf{(a) $\Delta_{\G_\M}$}};
            \end{tikzpicture}
        \end{subfigure}
        \hfill
        \begin{subfigure}{0.325\textwidth}
            \centering
            \begin{tikzpicture}
                \node at (0,0) 
                {\includegraphics[width=\textwidth]{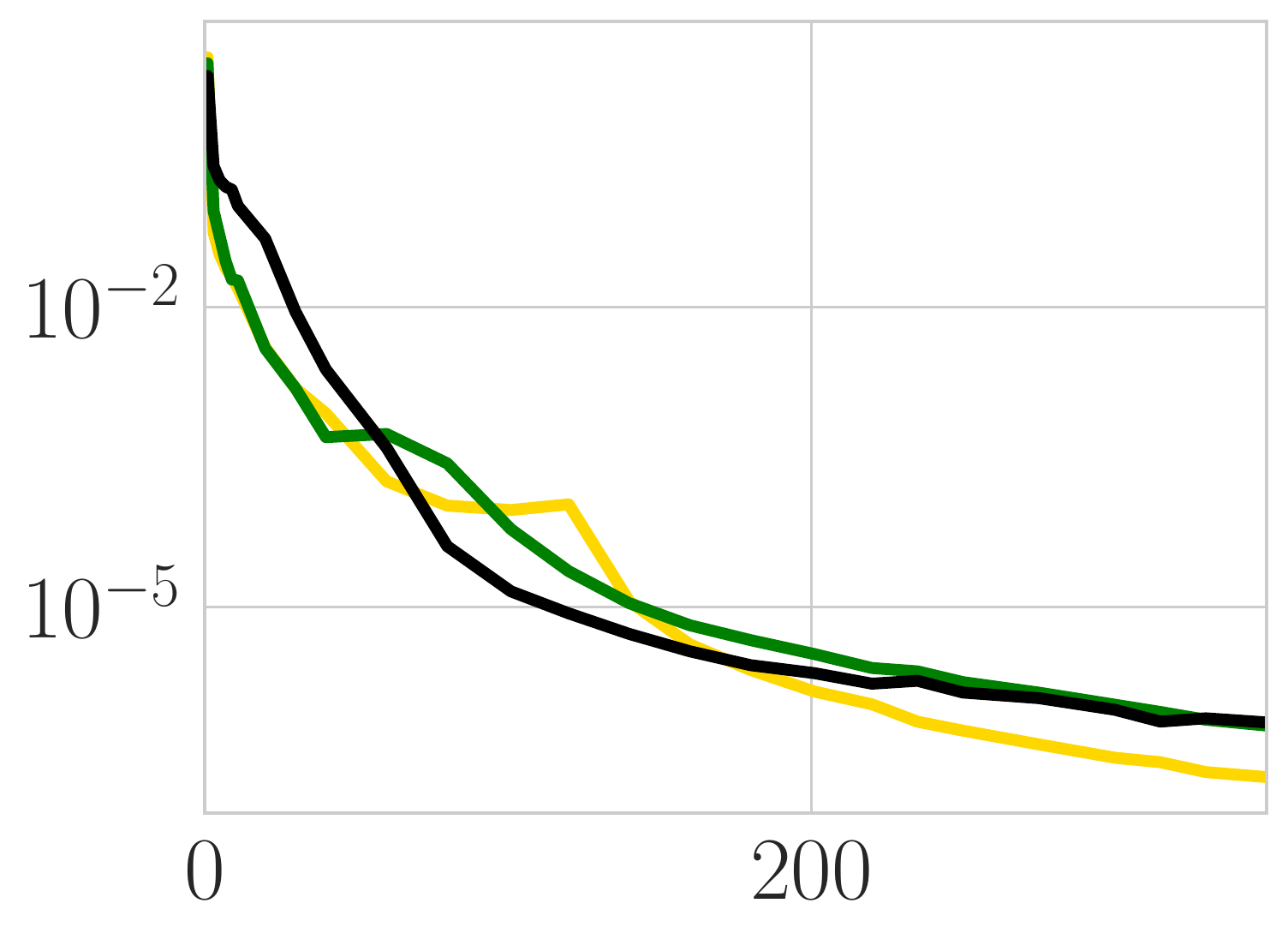}};
                \node at (0.2,-2) [scale=0.7]{Epoch};
    			\node at (0.0,2.1) [scale=0.9]{\textbf{(b)} $\beta_{\text{NC}}$};
            \end{tikzpicture}
        \end{subfigure}
        \hfill
        \begin{subfigure}{0.325\textwidth}
            \centering
            \begin{tikzpicture}
                \node at (0,0) 
                {\includegraphics[width=\textwidth]{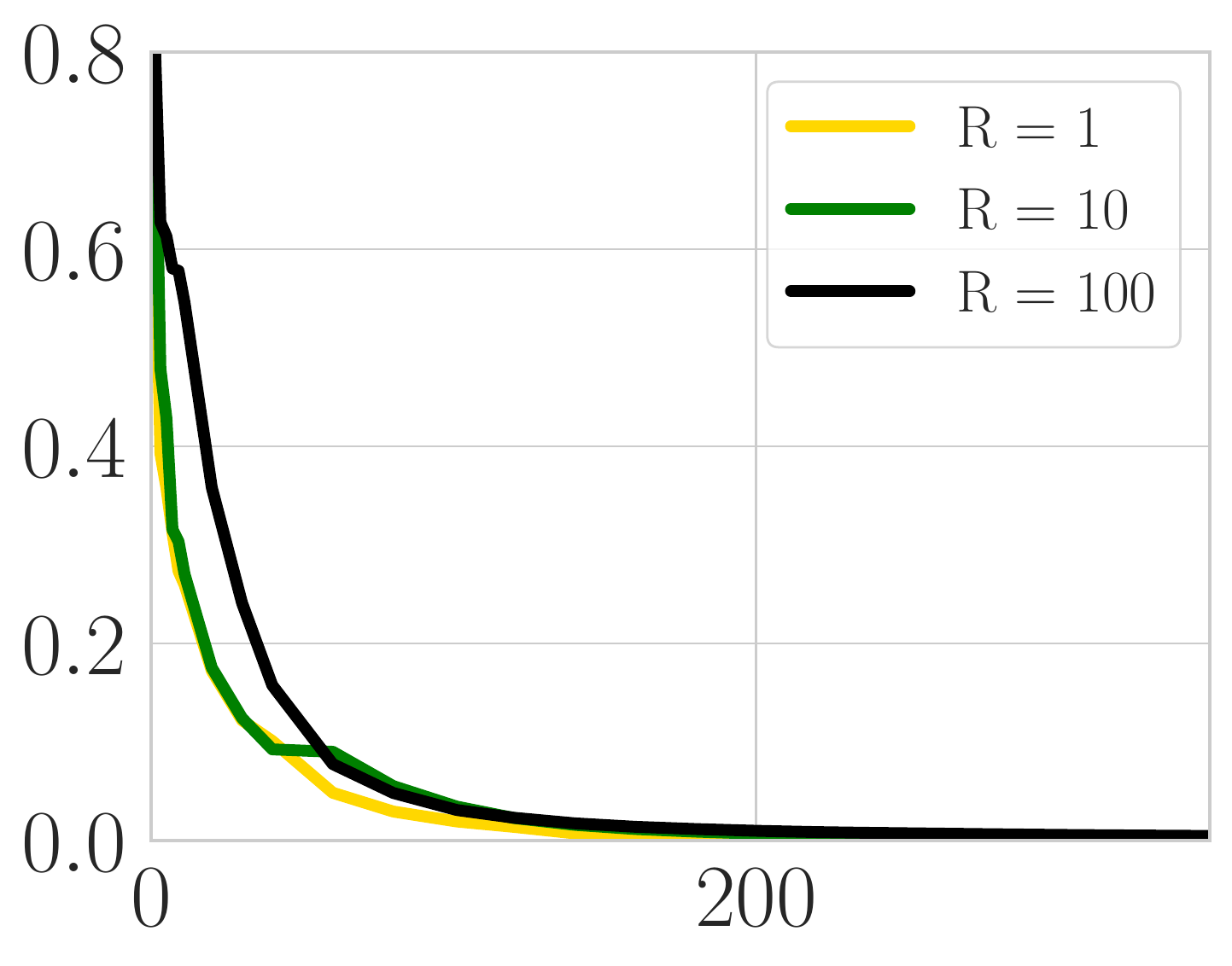}};
                \node at (0.13,-2.05) [scale=0.7]{Epoch};
    			\node at (0.0,2.1) [scale=0.9]{\textbf{(c)} $\text{Ave}_{c \neq c'}\text{ }\alpha_{\text{sim}(c,c')}$};
            \end{tikzpicture}
        \end{subfigure}
        \captionsetup{width= \linewidth}
        \caption[loss]
        {Geometry convergence metrics \textbf{(a)} $\Delta_{\G_\M}$, \textbf{(b)} $\beta_{\text{NC}}$, and \textbf{(c)} $\text{Ave}_{c \neq c'}\text{ }\alpha_{\text{sim}(c,c')}$  for a 6 layer multilayer perceptron (MLP) model with ReLU activations trained with SCL and MNIST under $R$-STEP imbalance.}
        \label{fig:MLP convergence} 
    \end{figure*}

As described in Sec.~\ref{sec:Deepnet_SCLGeom_exp} and Sec.~\ref{sec:full_batch_theory} the optimality of the OF geometry for the \ref{eq:SCL_UFM_full} is invariant to the choice of the temperature parameter $\tau$. However, we have found that the speed of convergence to the OF geometry is dependent on the choice of $\tau$. Shown in Fig.~\ref{fig:effect of tau full batch} is a full batch SCL experiment on $n=1000$ samples of MNIST trained on ResNet-18 with $\tau = 0.1, 1, 10$. It is clear from Fig.~\ref{fig:effect of tau full batch} that: (a) the within-class feature variation converges significantly faster for smaller $\tau$; (b) the angles converge to k-OF faster and smoother for smaller $\tau$ as well (see Fig.~\ref{fig:effect of tau full batch} (b)). In all cases, values of $\beta_\text{NC}$ and  $\alpha_{\text{sim}}(c,c')$ continue to decrease, and we anticipate further convergence if the networks are trained for longer. These results qualitatively agree with the findings of  \cite{khosla2020supervised} which suggest that smaller $\tau$ improves training speed. 

\subsection{Complementary results and discussions on batch-binding}\label{app:comp_results}
In this section, we describe examples where batching methods and batch-binding help improve the convergence speed of embeddings geometries to OF. 

\subsubsection{How batch-binding ensures a unique OF geometry}
\begin{figure}%
    \centering
    \subfloat[\centering]{{\includegraphics[width=0.65\linewidth]{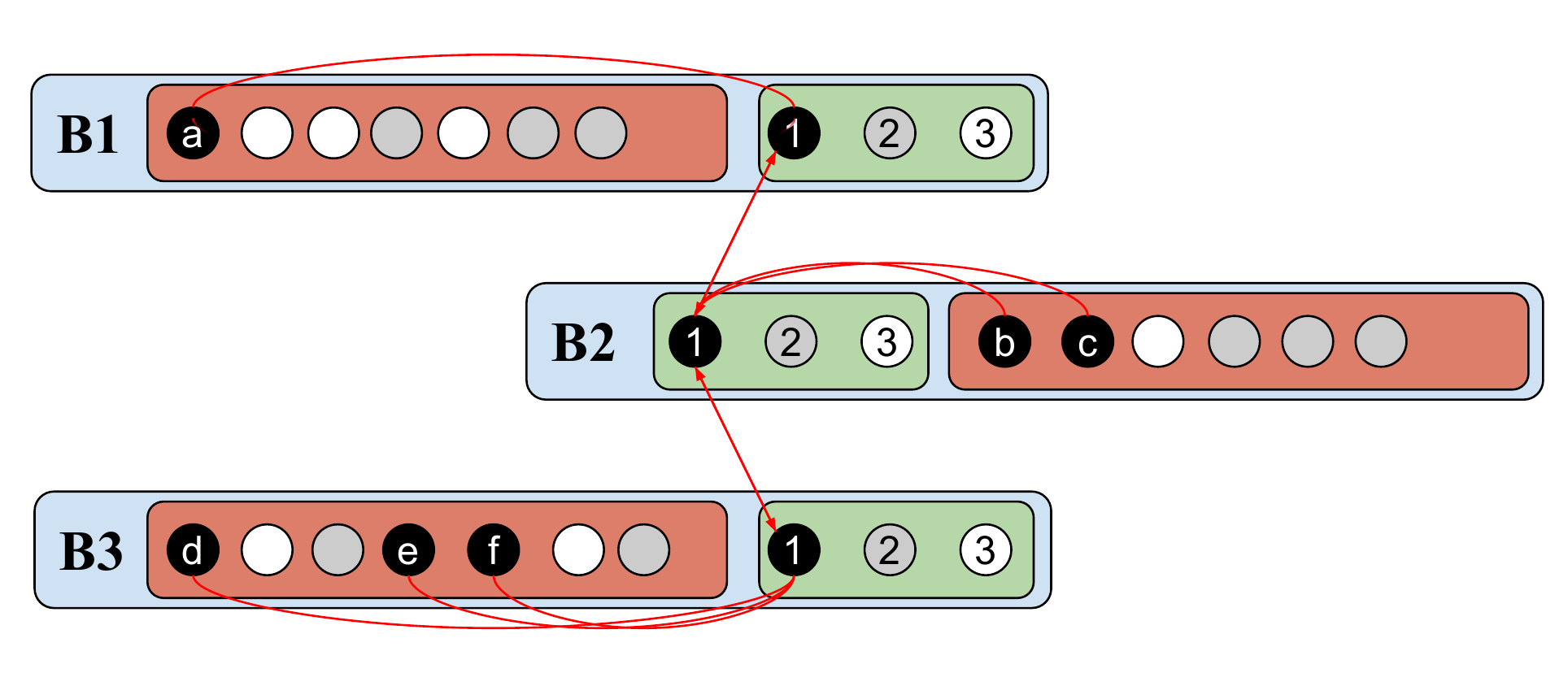} }}%
    \newline
    \centering
    \subfloat[\centering]{{\includegraphics[width=0.36\linewidth]{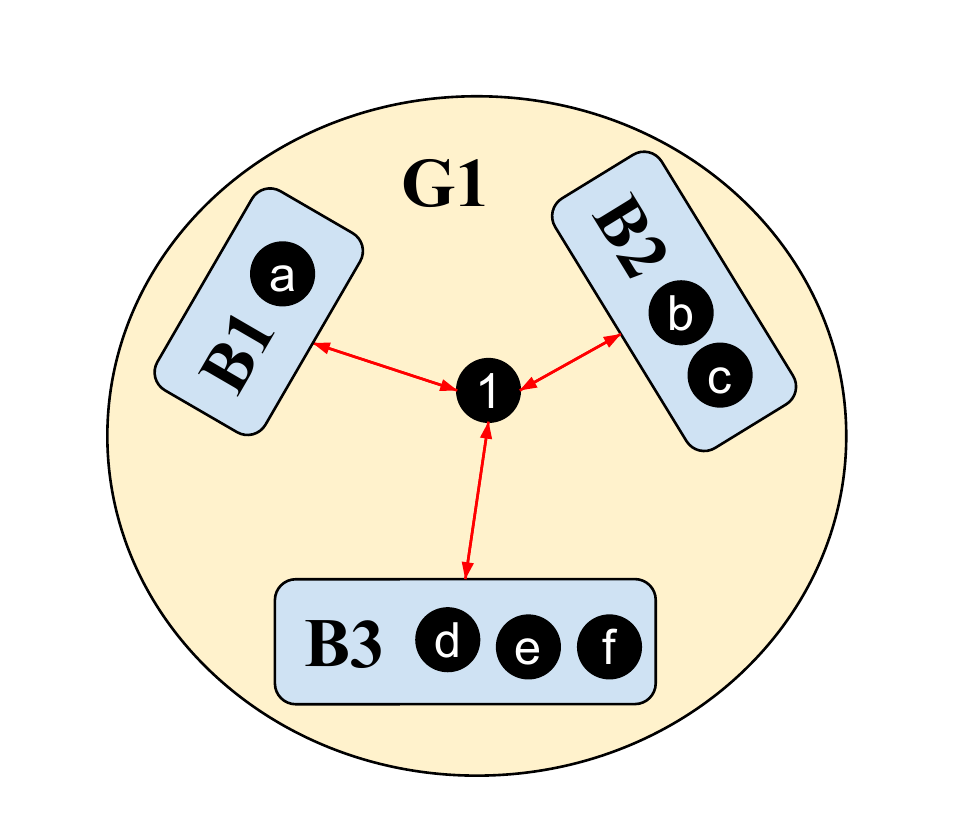} }}%
    \qquad
    \subfloat[\centering]{{\includegraphics[width=0.32\linewidth]{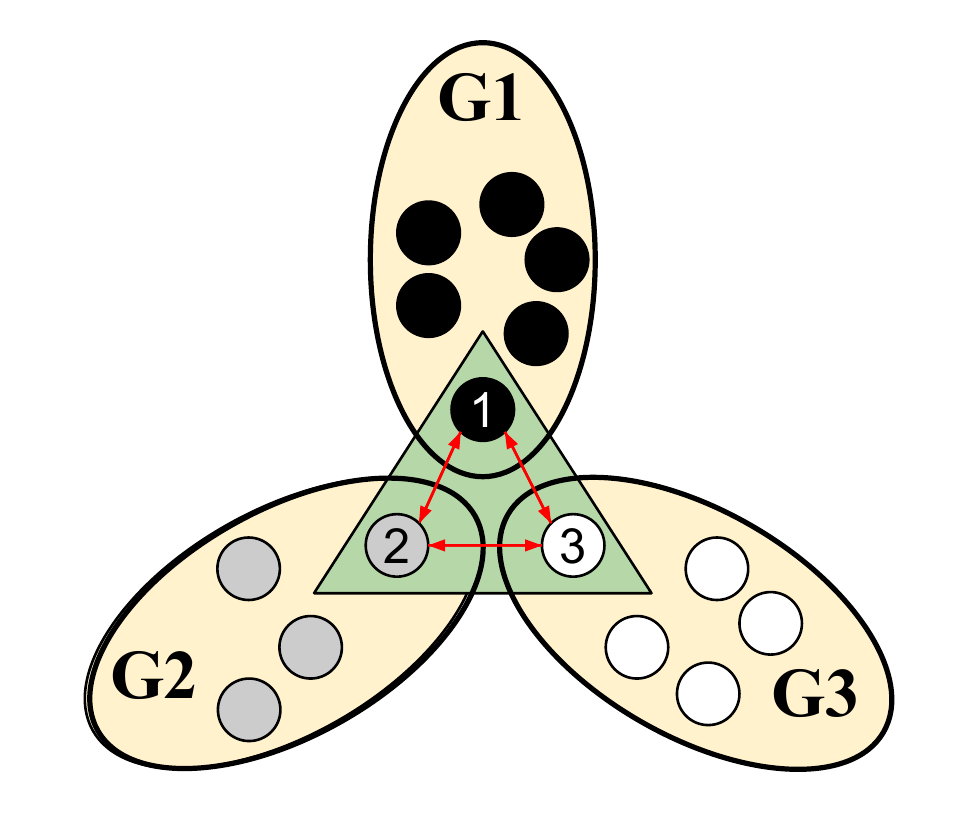} }}%
    \caption{A simple illustration to explain how adding binding examples to each batch satisfies the requirements of Cor.~\ref{cor:batch}, thus leads to unique OF geometry. \textbf{(a)} Gives a $3$ class (black, grey and white) classification  example with 3 batches. In addition to the data for each batch (included in red), the binding examples $1,2,3$ are added to each batch. \textbf{(b)} Gives the Batch Interaction Graph for the induced subgraph $G_1$ of the class (black) of example $1$
    and illustrates how all batches are connected through examples $1$, satisfying the first condition of Cor.~\ref{cor:batch}. \textbf{(c)} Elaborates on how all three class  graphs $G_1,G_2,G_3$ are connected through the interactions of data points $1,2,3$, satisfying the second condition of Cor.~\ref{cor:batch}.}
    \label{fig:Binding_Examples_Visuals}%
\end{figure}

~ 
Fig.~\ref{fig:Binding_Examples_Visuals} provides a simple illustration demonstrating how adding binding examples can satisfy the requirements of Cor.~\ref{cor:batch}.
While there are alternative approaches to satisfy the graph conditions stated in Cor.~\ref{cor:batch} for ensuring a unique OF geometry, the method of adding the same $k$ examples to each batch is a straightforward technique that is often computationally efficient, considering that batch sizes typically exceed the number of classes $k$.

 \begin{figure}[ht]
    \hspace*{-5pt}
    \begin{subfigure}[b]{\textwidth}
        \centering
        \begin{tikzpicture}
            \node at (0.0,0) 
            {\includegraphics[width=0.75\textwidth]{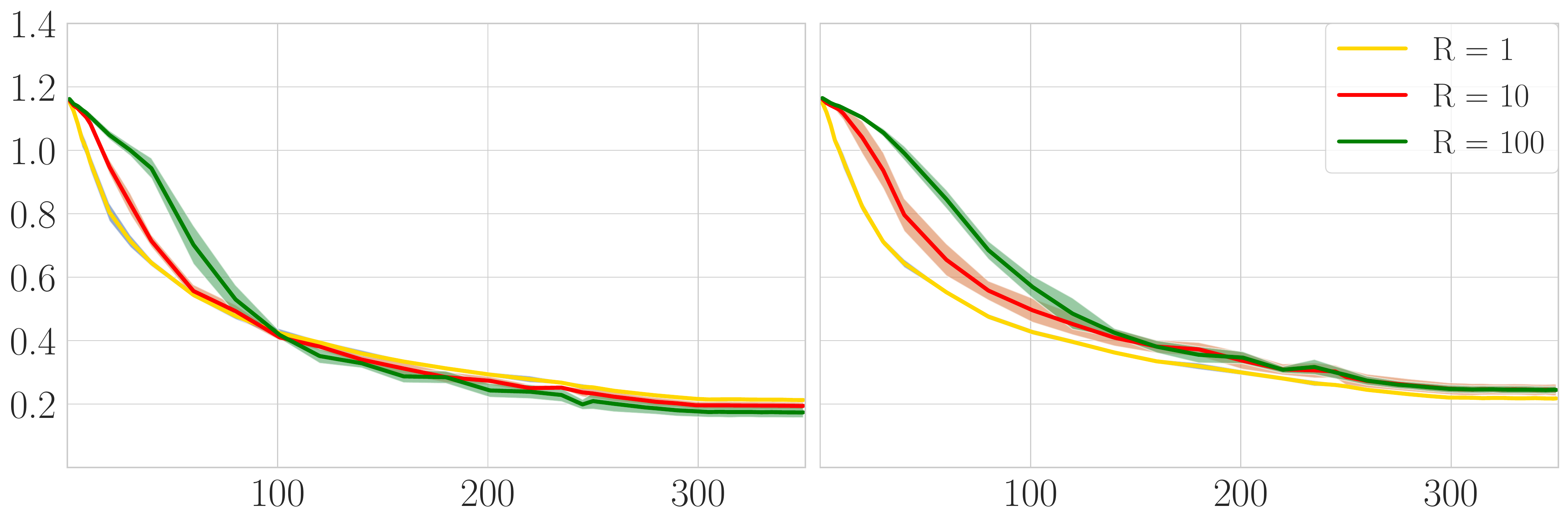}};
            \node at (-2.4,2.07) [scale=0.9]{\textbf{Step}};
            \node at (2.6,2.07) [scale=0.9]{\textbf{Long-tail}};
            \node at (-6.25,0.0)  [scale=0.9, rotate=90]{$\Delta_{\G_\M}$};
            \node at (0.2,-2.07) [scale=0.9]{\textbf{Epoch}};
        \end{tikzpicture}
    \end{subfigure}
    \caption
    {Convergence of $\G_\M$ to the OF geometry for a ResNet-18 model trained on CIFAR10 under STEP imbalance, with data augmentation. Results represent the average run results over 5 versions of the experiments with randomly chosen binding examples from each class. 
    }
    \label{Fig:error_bars_plotGM}
\end{figure}

\noindent\textbf{Geometry convergence for different batching schemes.}~ 
A comparison of OF convergence with three batch selection schemes was introduced in Sec.~\ref{sec:batching}. We considered three strategies: 1) \textbf{No batch-shuffling}, 2) \textbf{Batch-shuffling}, and 3) \textbf{No batch-shuffling + batch-binding}. We noted that optimization with batch-binding, even in absence of shuffling is very effective at guiding the solution to the unique OF geometry. 

\begin{figure}[t]
    \hspace*{-16pt}
    \begin{subfigure}[b]{\textwidth}
        \centering
        \begin{tikzpicture}
            \node at (0.0,0) 
            {\includegraphics[width=0.75\textwidth]{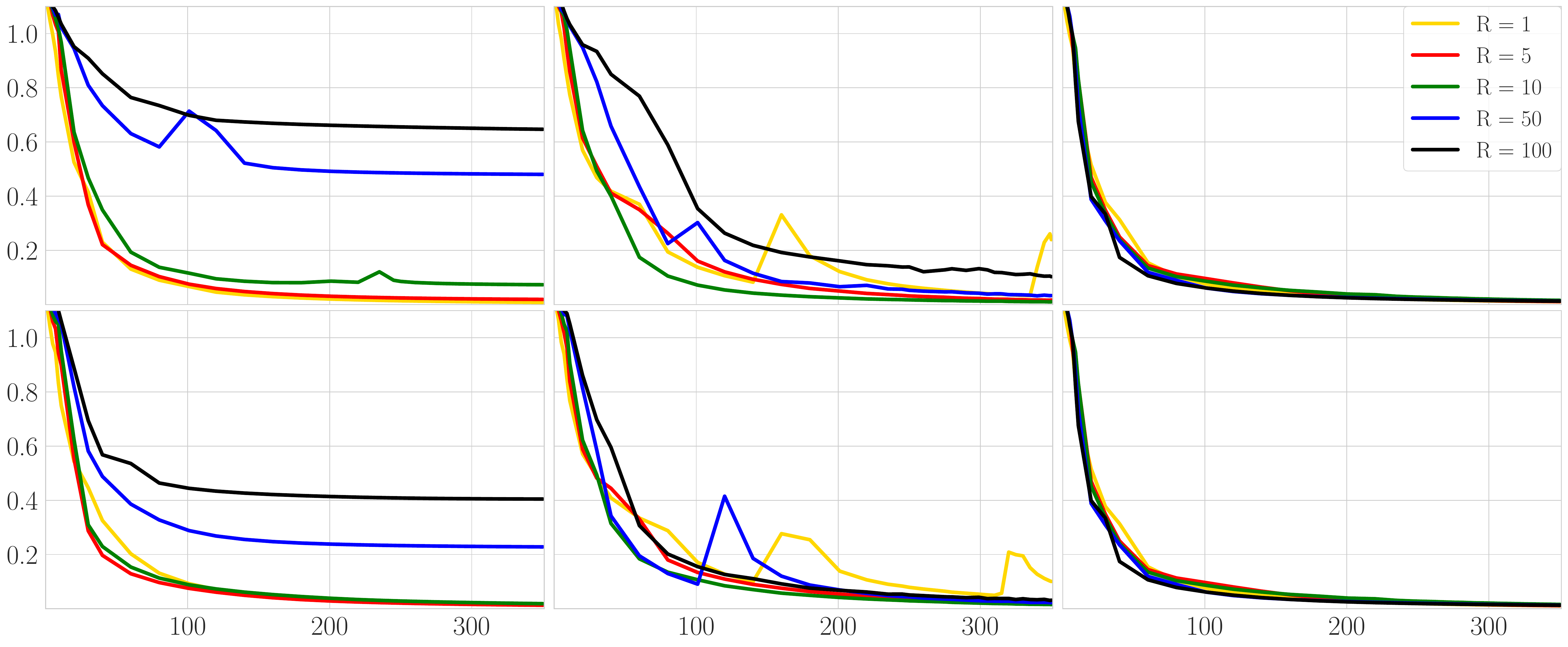}};
            \node at (-3.5,2.7) [scale=0.9]{\textbf{No Batch Shuffling}};
            \node at (0.1,2.7) [scale=0.9]{\textbf{Batch Shuffling}};
            \node at (3.8,3.1) [scale=0.9]{\textbf{No Batch Shuffling}};
            \node at (3.7,2.7) [scale=0.9]{\textbf{+ Batch-Binding}};
            \node at (-6.2,1)  [scale=0.9, rotate=90]{\textbf{Step}};
            \node at (-6.2,-0.8)  [scale=0.9, rotate=90]{\textbf{Long-tail}};
            \node at (0.1,-2.7) [scale=0.9]{\textbf{Epoch}};
        \end{tikzpicture}
    \end{subfigure}
    \caption
    {{Convergence to the OF geometry for various batching schemes {including the analysis-inspired scheme ``No Batch Shuffling + Batch-Binding''}. See text for details. Experiments conducted with CIFAR-10 and ResNet-18.}}
    \vspace{-8pt}
    \label{Fig:Deepnet_SCL_Geom_NC_BindingBatch}
\end{figure}

Further, in Table~\ref{Tab:NCC_Bindings} in the appendix, we present experimental evidence in  suggesting that batch binding does not negatively effect the balanced test accuracy of \textit{Nearest Class Center} (NCC) classifier. This observation, paired with the impact of binding examples on OF convergence, motivates further analysis of the relationship between OF geometry and test accuracy.

\begin{figure}[t]
    \hspace*{-5pt}
    \begin{subfigure}[b]{\textwidth}
        \centering
        \begin{tikzpicture}
            \node at (0.0,0) 
            {\includegraphics[width=0.75\textwidth]{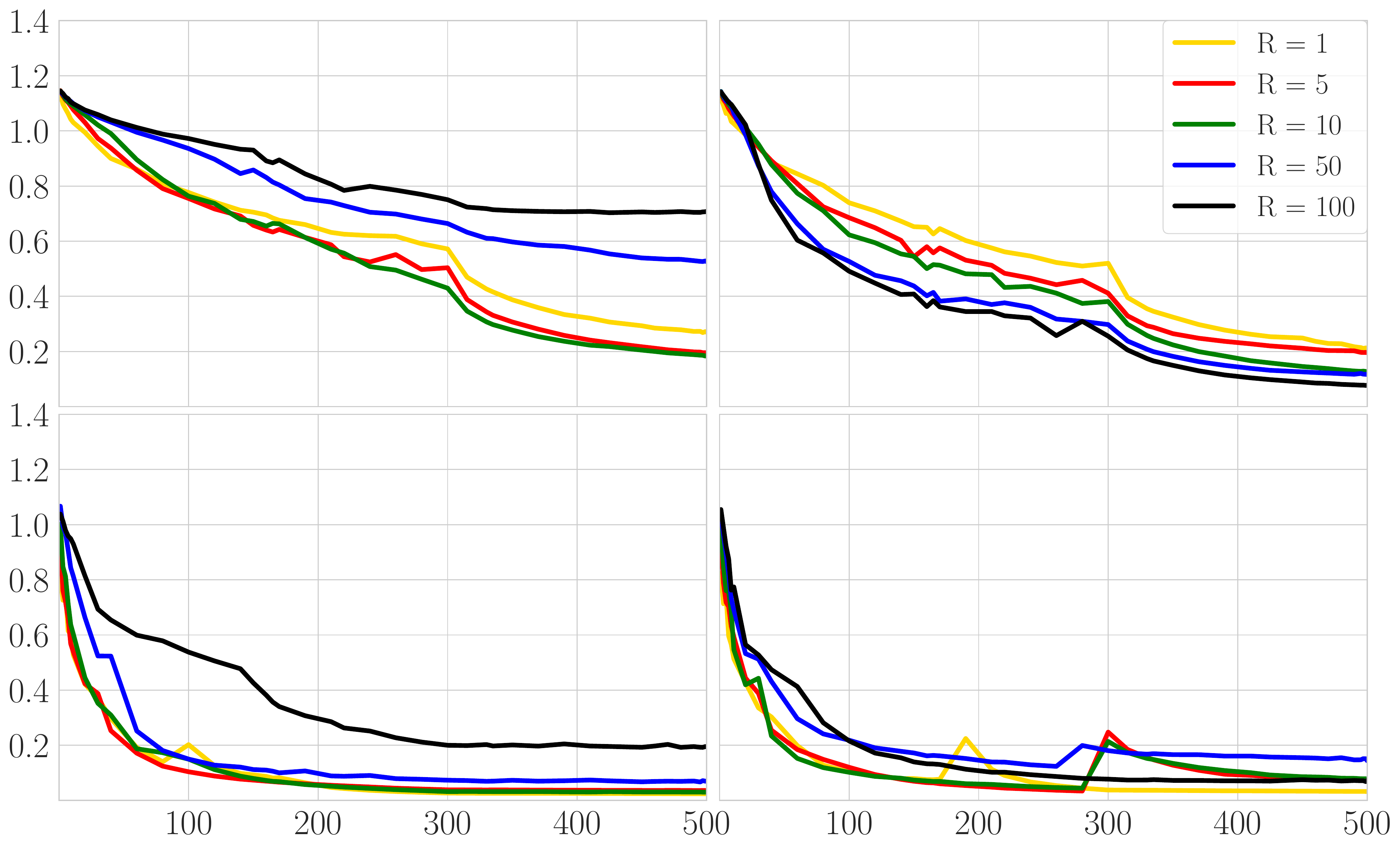}};
            \node at (-2.4,3.8) [scale=0.9]{\textbf{Batch Shuffling}};
            \node at (2.6,4.3) [scale=0.9]{\textbf{Batch Shuffling}};
            \node at (2.6,3.8) [scale=0.9]{\textbf{+ Batch-Binding}};
            \node at (-6.2,1.6)  [scale=0.9, rotate=90]{\textbf{CIFAR10}};
            \node at (-6.2,-1.2)  [scale=0.9, rotate=90]{\textbf{MNIST}};
            \node at (0.2,-3.8) [scale=0.9]{\textbf{Epoch}};
        \end{tikzpicture}
    \end{subfigure}
    \caption
    {Convergence of $\G_\M$ to the OF geometry for a DenseNet-40 model trained on CIFAR10 and MNIST under STEP imbalance with and without batch-binding. While the impact of batch-bindings is less noticeable when training on a simple dataset such as MNIST,  the convergence is significantly improved, particularly under imbalance, when training on a more complex dataset such as CIFAR10. 
    }
    \vspace{-10pt}
    \label{Fig:DenseNet_BindingsComparison}
\end{figure}

\noindent\textbf{Improving convergence for less powerful models.}~
When conducting our experiments, we noticed that DenseNet-40 converges slower compared to ResNet-18. A reason for this may be related to the very different complexities of the two models: ResNet-18 has substantially more trainable parameters compared to DenseNet-40. In an attempt to improve the convergence speed, we reduce the batch size for DenseNet experiments to $128$ to increase the number of SGD iterations. We also train DenseNet for $500$ epochs instead of $350$ while reducing the learning rate from $0.1$ to $0.01$ at epoch $300$. Yet, we observe that with much smaller batch sizes, the embedding geometry does not always converge to the OF geometry, especially when training with high imbalance ratios. Specifically when training with randomly reshuffled batches, there is a higher chance that the examples do not interact with each other even if trained for long, suggesting that the optimal solution for all batches is not necessarily OF. We hypothesize that this likelihood increases when using smaller batch sizes during training. In order to combat this effect, we ran the DenseNet experiments with the addition of binding examples to every batch. 
As shown in Fig.~\ref{Fig:DenseNet_BindingsComparison}, we find that adding such binding examples significantly improves the convergence to OF, particularly when training on more complex datasets (CIFAR10 compared to MNIST) and under more severe imbalances (STEP imbalance with $R= 50,100$). These results emphasize the impact of batch-binding and provide further evidence regarding our claims in Sec.~\ref{sec:mini_batch_theory}. While the convergence values of $\Delta_{\G_\M}$ are higher when compared to ResNet-18, we deem them reasonable  considering the difference in the number of parameters between the two architectures. Similar to ResNet-18, we provide geometric convergence results for DenseNet-40 on different datasets in Fig.~\ref{Fig:Deepnet_SCL_DenseNet_Geom} in the appendix.

\begin{figure}[t]
    \hspace*{-5pt}
    \begin{subfigure}[b]{\textwidth}
        \centering
        \begin{tikzpicture}
            \node at (0.0,0) 
            {\includegraphics[width=0.75\textwidth]{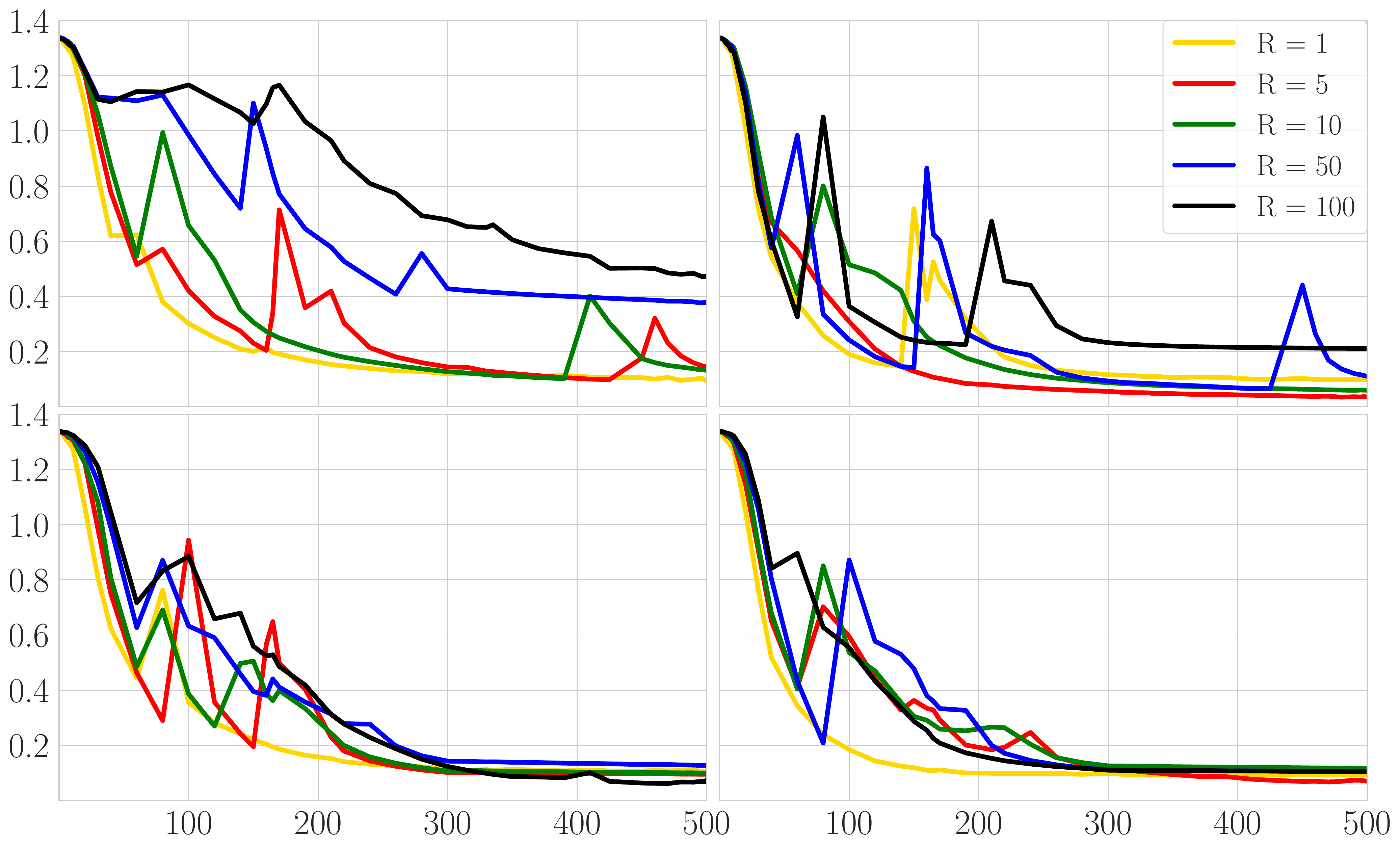}};
            \node at (-2.4,3.7) [scale=0.9]{\textbf{Batch Shuffling}};
            \node at (2.6,4.1) [scale=0.9]{\textbf{Batch Shuffling}};
            \node at (2.6,3.7) [scale=0.9]{\textbf{+ Batch-Binding}};
            \node at (-6.2,1.6)  [scale=0.9, rotate=90]{\textbf{Step}};
            \node at (-6.2,-1.2)  [scale=0.9, rotate=90]{\textbf{Long-tail}};
            \node at (0.2,-3.8) [scale=0.9]{\textbf{Epoch}};
        \end{tikzpicture}
    \end{subfigure}
    \caption
    {Convergence of embeddings geometry to OF as measured by $\Delta_{\G_\M} := \|\frac{\G_\M}{\| \G_\M\|_F} - \frac{\Id_{k}}{\|\Id_{k}\|_F}\|_F$ for ResNet-18 trained on imbalanced CIFAR100 with SCL and different batching schemes. 
    Adding binding examples helps with the convergence to the OF geometry, especially under STEP imbalance with larger imbalance ratios.}
    \label{Fig:Deepnet_SCL_CIFAR100_GMu}
    \vspace{-15pt}
\end{figure}

 In another experiment, we consider the convergence of ResNet-18 embeddings to the OF geometry when trained with CIFAR100. Models are trained for $500$ epochs with a constant learning rate of $0.1$ and a batch size of 1024. In Fig.~\ref{Fig:Deepnet_SCL_Geom_GMu}, we see that ResNet-18 easily converges to CIFAR10. However, without batch binding, in Fig.~\ref{Fig:Deepnet_SCL_CIFAR100_GMu} ResNet-18 struggles to convergence. With a large number of classes ($k = 100$ for CIFAR100), it becomes increasingly more likely for the randomly reshuffled batches to not allow sufficient interactions between examples. Particularly for large imbalance ratios (e.g., $R=100,50$), since the number of samples in each minority class could become as low as $5$-$10$, some batches might never encounter examples from minority classes. This is particularly noticeable for STEP-Imbalance and large imbalance ratios, where half of the classes are minorities. On the other hand, including the binding examples in each batch (right side) can improve the convergence of the feature geometry to OF.  

\subsubsection{Convergence in non-ReLU settings}
\begin{figure*}
            \centering
            \begin{tikzpicture}
                \node at (0.0,0) 
                {\includegraphics[width=0.4\textwidth]{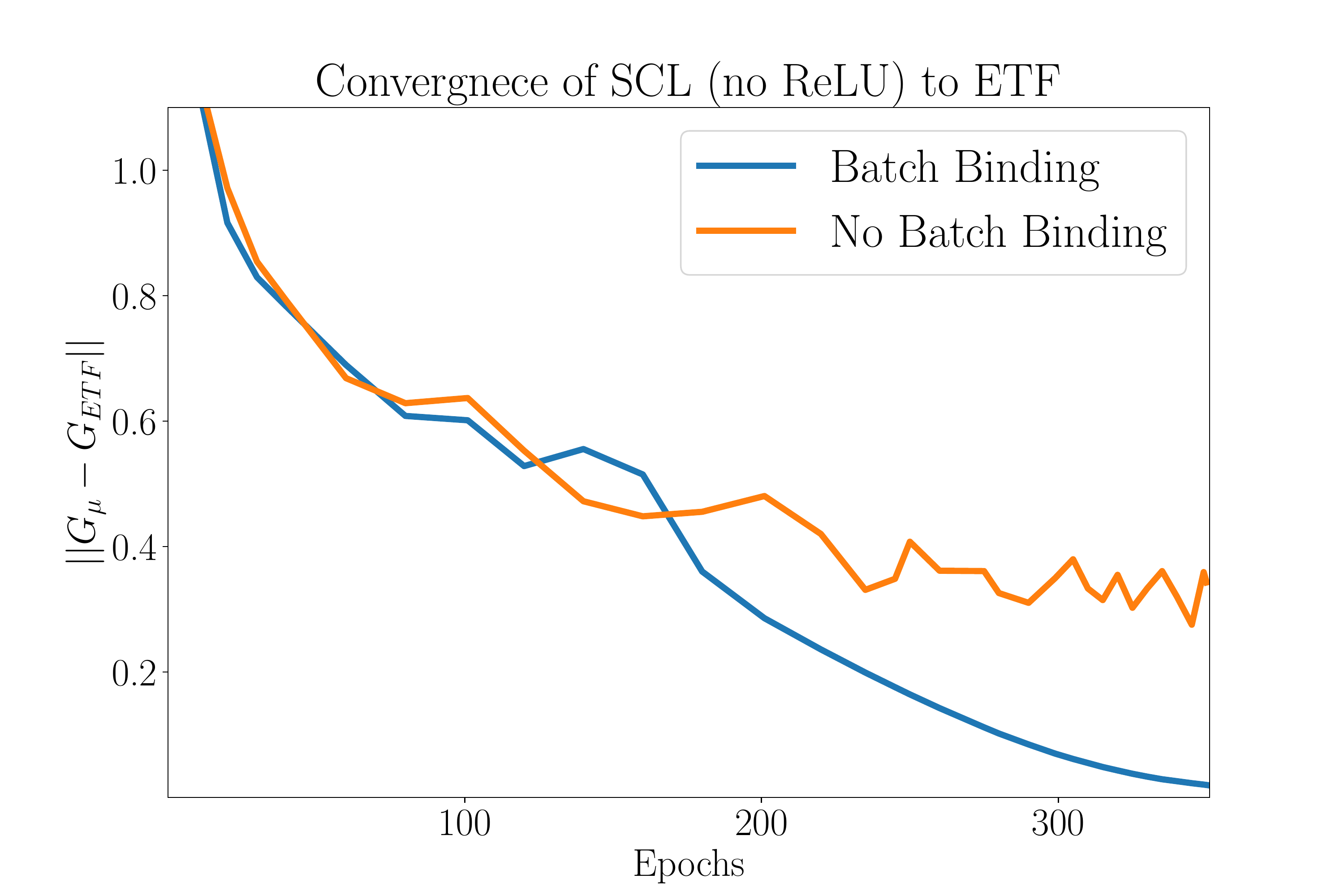}};
            \end{tikzpicture}
        \caption{Geometry convergence of $\mathbf{G}_\mu = \mathbf{M}^\top\mathbf{M}$ to the ETF geometry with and without binding examples. A batch size of 128 is used, there is no ReLU on the output features.}
        \label{fig:binding_ETF} 
\end{figure*}

In addition to observing that batch-binding improves the convergence of SCL with ReLU to OF, we consider the setting without ReLU. In particular in Fig.~\ref{fig:binding_ETF} we compare the convergence of the Gram matrix of mean embeddings to the ETF geometry with and without batch-binding. We train a ResNet-18 model with CIFAR-10. We observe that with a reduced batch size of 128 the convergence can be significantly improved with binding-examples.
\subsection{On the convergence of batch-shuffling to OF}\label{app:batch-shuffling}
{It can be readily observed  that a fixed mini-batch partition without shuffling (\emph{No batch-shuffling} in the experiments above) does not satisfy the conditions of  Cor.~\ref{cor:batch}. Consequently, OF is \emph{not} the unique minimizer geometry in such scenarios. This is clearly manifested by the inadequate convergence levels observed in our experiments, as depicted for example in Fig. \ref{Fig:Deepnet_SCL_Geom_NC_BindingBatch}. In contrast, in our experiments, when the examples are randomly shuffled prior to partitioning into mini-batches at each epoch (which we call \emph{Batch Shuffling}), the convergence behavior to OF shows a significant improvement  compared to the case of \emph{No batch-shuffling}. The observed improvement can be attributed to the fact that shuffling enables interactions among examples across epochs. To elaborate on this notion, we present a formalization of batch shuffling below.}


Let $\Sc$ denote the set of all permutations of $[n] = \{1,2,...,n\}$, where $n$ is the total number of training examples. Additionally, let $b$ 
denote {a fixed} batch-size. For simplicity, assume $n$ is an integer multiple of $b$. At the beginning of each training epoch $t$, we sample a permutation $s_t=(i_1,i_2,....,i_n)$ uniformly from $\mathcal{S}$ \emph{with} replacement. We then define $\Pc(s_t) := \{\{i_1,...,i_b\},\{i_{b+1},...,i_{2b}\},...,\{i_{n-b+1},...,i_{n}\}\}$ as the collection of total $\frac{n}{b}$ mini-batches obtained by partitioning $s_t$. These are the mini-batches used at epoch $t$. Thus, SCL in this epoch can be written as follows:
\begin{align*}
     \Lc_{s_t}(\Hb):=\sum_{B \in \Pc(s_t)}\sum_{i\in B} \Lc_B(\Hb;i), \quad \text{where } \Lc_B(\Hb;i):= \frac{1}{n_{B,y_i}-1}\sum_{\substack{j\in B\\j \neq i, y_j = y_i}}\log\big(\sum_{\substack{\ell \in B \\ \ell \neq i}}\exp{\left(\h_i^\top\h_\ell - \h_i^\top\h_j\right)}\big),
\end{align*}
and recall $n_{B,c} = |\{i:i\in B,y_i=c\}|$.
Consider also the loss over \emph{all} mini-batches obtained by partitioning each permutation of $\Sc$: 
\begin{align}\label{eq:scl_all_permutations}
     \Lc_{\Sc}(\Hb):=\sum_{s \in \Sc}\sum_{B \in \Pc(s)}\sum_{i\in B}\Lc_B(\Hb;i),
\end{align}
Now, since $s_t$ is uniformly sampled from $S$, we have the following relation for the expected per-epoch loss to the total loss given in Eqn. \eqref{eq:scl_all_permutations}:  $\mathbb{E}_{s_t}\Lc_{s_t}(\Hb) = \nicefrac{1}{|\Sc|}\,\Lc_{\Sc}(\Hb)$. Therefore, training with batch-shuffling can be regarded as a stochastic version of minimizing the total loss in Eq. \eqref{eq:scl_all_permutations}. Regarding the latter, it can be checked (see Rem.~\ref{rem:big_shuffling} below) that it satisfies the conditions of Cor.~\ref{cor:batch}, thereby making OF its unique minimizer geometry. Taken together, these findings suggest that although the per-epoch loss $\Lc_{s_t}$ obtained through batch-shuffling does not satisfy the conditions of Cor. \ref{cor:batch} for any specific epoch $t$, it does satisfy them in expectation. In particular, this can be regarded as preliminary justification of the experimental observation of improved convergence \emph{with batch-shuffling} compared to \emph{No batch-shuffling} schemes. The latter schemes fail to satisfy Cor. \ref{cor:batch} under any circumstances, unless in the extreme case of a large batch size where $b=n$ and we recover $\Lc_{\text{full}}$. However, it is important to note that the aforementioned argument is a rough approximation, as it is not feasible to average over all $|\Sc|=n!$ permutations when optimization is typically performed over a limited number of epochs. This can explain why, despite exhibiting better convergence compared to the No batch-shuffling schemes, batch-shuffling schemes still demonstrate non-smooth and inconsistent convergence patterns in our experiments. This behavior becomes particularly evident when comparing the convergence levels of batch-shuffling to our \emph{batch-binding} scheme, which is specifically designed to satisfy the conditions of Cor. \ref{cor:batch} at \emph{each} epoch. 

\begin{remark}\label{rem:big_shuffling}
    To show that Eqn.~\eqref{eq:scl_all_permutations} satisfies Cor.~\ref{cor:batch}, we will show that the corresponding \emph{batch interaction graph} $G$ is a complete graph. This suffices, because the induced subgraphs $G_c$ from Def.~\ref{def:big} for a complete graph are connected graphs, and there exist multiple edges between $G_{c_1}$ and $G_{c_2}$ for $c_1,c_2\in [k]$. To show that $G$ is a complete graph, we argue as follows. Consider the set $\Bc$ of all mini-batches obtained by partitioning all $n!$ elements of $\Sc$. Fix $b \geq 2$, so that the mini-batches have at least two examples. We will consider a subset $\Bc_1\subseteq\Bc$ of mini-batches and show that the corresponding batch interaction graph for $\Bc_1$ is a complete graph, which would then imply the batch interaction graph of $\Bc$ is also a complete graph. Specifically, let $\Bc_1\subseteq\Bc$ denote the mini-batches comprising of the first $b$ indices in every permutation $s\in\Sc$. In other words, from a given $s = (i_1,i_2,...,i_n) \in \Sc$, we let $\Bc_1$ include the mini-batch $\{i_1,i_2,...,i_b\}$. Since $\Sc$ includes all permutations of $[n]$, $\Bc_1$ contains all $b$-length permutations of the elements of $[n]$, possibly with repetitions. Thus, the \emph{batch interaction graph} created by the mini-batches in $\Bc_1$ is a \emph{complete graph}. In the definition of the \emph{batch interaction graph}, since a repeated presence of a pair of examples does not alter the graph, the batch interaction graph for $\Bc$ is the same complete graph. Thus, $\Bc$ satisfies Cor.~\ref{cor:batch}.
\end{remark}

\subsection{ReLU does not compromise Accuracy}\label{sec:generalization}
Having discussed the symmetry-preserving value of ReLU activation in SCL training, we now turn towards the impact on generalization accuracy. While the exact relationship between the training feature geometry and generalization is an open question, it is conceivable that maintaining symmetry in features is beneficial to generalization \citep{behnia2023implicit, zhu2022balanced}. In fact, our experiments with class-imbalance reveal that ReLU improves the test accuracy significantly in a subset of the cases, while not deteriorating in others. There have been suggestions of loss function modifications in such a way that the feature geometry is symmetric in presence of class-imbalance, consequently also achieving improvements in generalization \cite{behnia2023implicit, zhu2022balanced}. Since the role of a projector block is of interest when training with SCL, we compare the experimental test accuracies both when the projector is retained (post-projection) as well as discarded (pre-projection) during inference \citep{khosla2020supervised,chen2020simple,zhu2022balanced}.




\begin{table}[!ht]
\fontsize{6pt}{15pt}\selectfont
    \centering
    \begin{tabular}{|l|l|l|l|l|l|l|l|l|l|}
    \cline{3-10}
         \multicolumn{2}{c}{} & \multicolumn{4}{|c}{Pre-Projection}   & \multicolumn{4}{|c|}{Post-Projection}   \\  \cline{3-10}
         \multicolumn{2}{c}{} & \multicolumn{2}{|c}{MLP} & \multicolumn{2}{|c}{MLP + ReLU} & \multicolumn{2}{|c}{MLP} & \multicolumn{2}{|c|}{MLP + ReLU} \\ \cline{2-10}
        \multicolumn{1}{c|}{} & R & Step & LongTail & Step & LongTail & Step & LongTail & Step & LongTail \\ \cline{1-10}

         \parbox[t]{3mm}{\multirow{3}{*}{\rotatebox[origin=c]{90}{CIFAR-10 }}} & 1 & \multicolumn{2}{|c}{91.88 $\pm$ 0.29}  & \multicolumn{2}{|c}{92.04 $\pm$ 0.10} & \multicolumn{2}{|c}{91.94 $\pm$ 0.04} & \multicolumn{2}{|c|}{91.79 $\pm$ 0.13} \\ \cline{2-10}
        & 10 & 83.70 $\pm$ 1.09 & 83.82 $\pm$ 0.70 & 83.41 $\pm$ 0.64 & 84.40 $\pm$ 0.79 & 82.35 $\pm$ 0.67 & 82.97 $\pm$ 0.83 & 80.81 $\pm$ 1.01 & 82.99 $\pm$ 1.03 \\ \cline{2-10}
        & 100 & 60.19 $\pm$ 1.75 & 68.75 $\pm$ 0.31 & 67.12 $\pm$ 1.04 & 68.57 $\pm$ 1.69  & 55.31 $\pm$ 1.98 & 63.67 $\pm$ 0.77 & 59.83 $\pm$ 1.75 & 62.65 $\pm$ 2.37 \\ \cline{1-10}

         \parbox[t]{3mm}{\multirow{3}{*}{\rotatebox[origin=c]{90}{CIFAR-100 }}} & 1 & \multicolumn{2}{|c}{72.17 $\pm$ 0.23}  & \multicolumn{2}{|c}{72.32 $\pm$ 0.60} & \multicolumn{2}{|c}{72.30 $\pm$ 0.57} & \multicolumn{2}{|c|}{72.25 $\pm$ 0.35} \\ \cline{2-10}
        & 10 & 56.58 $\pm$ 0.50 & 57.10 $\pm$ 1.04 & 58.16 $\pm$ 0.76 & 57.88 $\pm$ 1.05 & 54.57 $\pm$ 0.49 & 56.27 $\pm$ 0.45 & 55.35 $\pm$ 0.33 & 57.08 $\pm$ 0.82 \\ \cline{2-10}
        & 100 & 43.49$\pm$ 0.30 & 37.19 $\pm$ 2.50 & 43.80 $\pm$ 0.25 & 39.71 $\pm$ 0.09  & 40.29 $\pm$ 0.33 & 35.29 $\pm$ 1.64 & 39.93 $\pm$ 0.33 & 37.21 $\pm$ 0.51 \\ \cline{1-10}
        
        \parbox[t]{3mm}{\multirow{3}{*}{\rotatebox[origin=c]{90}{Tiny ImageNet }}} & 1 & \multicolumn{2}{|c}{62.53 $\pm$ 1.23}  & \multicolumn{2}{|c}{62.26 $\pm$ 0.37} & \multicolumn{2}{|c}{62.71 $\pm$ 1.10} & \multicolumn{2}{|c|}{62.48 $\pm$ 0.40} \\ \cline{2-10}
        & 10 & 50.17 $\pm$ 1.44 & 50.8 $\pm$ 1.94 & 49.78 $\pm$ 0.67 & 49.81 $\pm$ 0.82 & 48.91 $\pm$ 1.6 & 50.23 $\pm$ 1.83 & 48.23 $\pm$ 0.53 & 48.76 $\pm$ 1.18 \\ \cline{2-10}
        & 100 & 39.13 $\pm$ 0.63 & 36.06 $\pm$ 0.89 & 40.02 $\pm$ 0.62 & 35.84 $\pm$ 1.33  & 37.41 $\pm$ 0.29 & 34.57 $\pm$ 0.95 & 37.19 $\pm$ 0.68 & 33.78 $\pm$ 1.34 \\ \cline{1-10}
    \end{tabular}
    \caption{Comparison of test accuracies for (i) CIFAR-10 (ii) CIFAR-100 (iii) Tiny ImageNet when ResNet-18 is trained with a 2-layer MLP projection head with and without ReLU on the output features. Comparisons in both Step and LT imbalance settings. Additionally a comparison of the performance using features before the projection head (pre-projection), and after the head (post-projection). }
    \label{tab:relu_noRelu_all_test}
\end{table}

Thus, as noted in Table \ref{tab:relu_noRelu_all_test} we make a remarkable observation that simply adding ReLU to the network can yield significant improvements in test accuracy in certain cases, while at least matching that of the case without ReLU in others. In light of this, our finding of the insensitivity of the embedding geometry in presence of ReLU serves a potential explanation. Our analysis therefore paves an important direction for future empirical investigations into the role of the activation functions in the projector network on generalization of SCL.

\begin{table}[h]
  \caption{NCC test accuracy for Batch-Binding}
  \hspace*{-20pt}
  \label{Tab:NCC_Bindings}
  \centering
  \begin{tabular}{l|ll|ll}
  \toprule
    & \multicolumn{2}{c}{Step} & \multicolumn{2}{c}{Long-tail}\\ 
    \cmidrule(r){2-3} \cmidrule(r){4-5}
    Ratio \\ ($R$)      & Reshuffling  & Reshuffling & Reshuffling  & Reshuffling \\ & & + Batch-Binding &  & + Batch-Binding  \\ 
    \midrule
    $10$ & $83.31 \pm 0.65 \%$ &   $83.27 \pm 0.27 \%$  & $85.04 \pm 0.28 \%$  &  $85.03 \pm 0.57 \%$  \\
    $100$     & $64.07 \pm 1.79 \%$ & $64.18 \pm 1.25 \%$   &  $67.75 \pm 1.21 \%$ & $68.08 \pm 0.72 \%$\\
  \end{tabular}
\end{table}
\begin{figure*}[h]
    \hspace*{-20pt}
    \begin{subfigure}[b]{\textwidth}
        \centering
        \begin{tikzpicture}
            \node at (0.0,0.9) 
            {\includegraphics[width=0.8\textwidth]{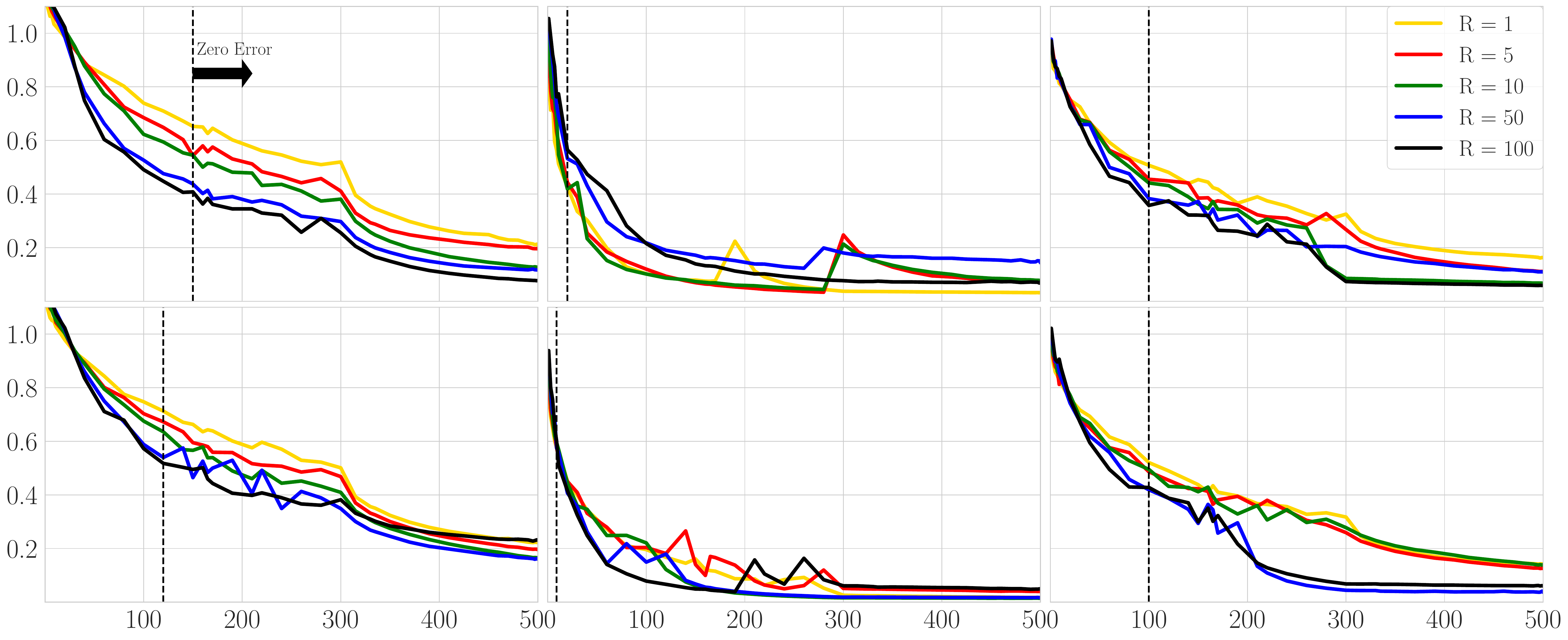}};
            \node at (-4,3.8) [scale=0.9]{\textbf{CIFAR10}};
            \node at (0.2,3.8) [scale=0.9]{\textbf{MNIST}};
            \node at (4.2,3.8) [scale=0.9]{\textbf{FMNIST}};
            \node at (-6.6,2.2)  [scale=0.9, rotate=90]{\textbf{Step}};
            \node at (-6.6,-0.2)  [scale=0.9, rotate=90]{\textbf{Long-tail}};
            \node at (0.1,-2) [scale=0.9]{\textbf{Epoch}};
        \end{tikzpicture}
        \caption{$\G_\M$ Convergence to OF}
    \end{subfigure}
    \hspace*{-20pt}
    \begin{subfigure}[b]{\textwidth}
        \centering
        \begin{tikzpicture}
            \node at (0.0,0.9) 
            {\includegraphics[width=0.8\textwidth]{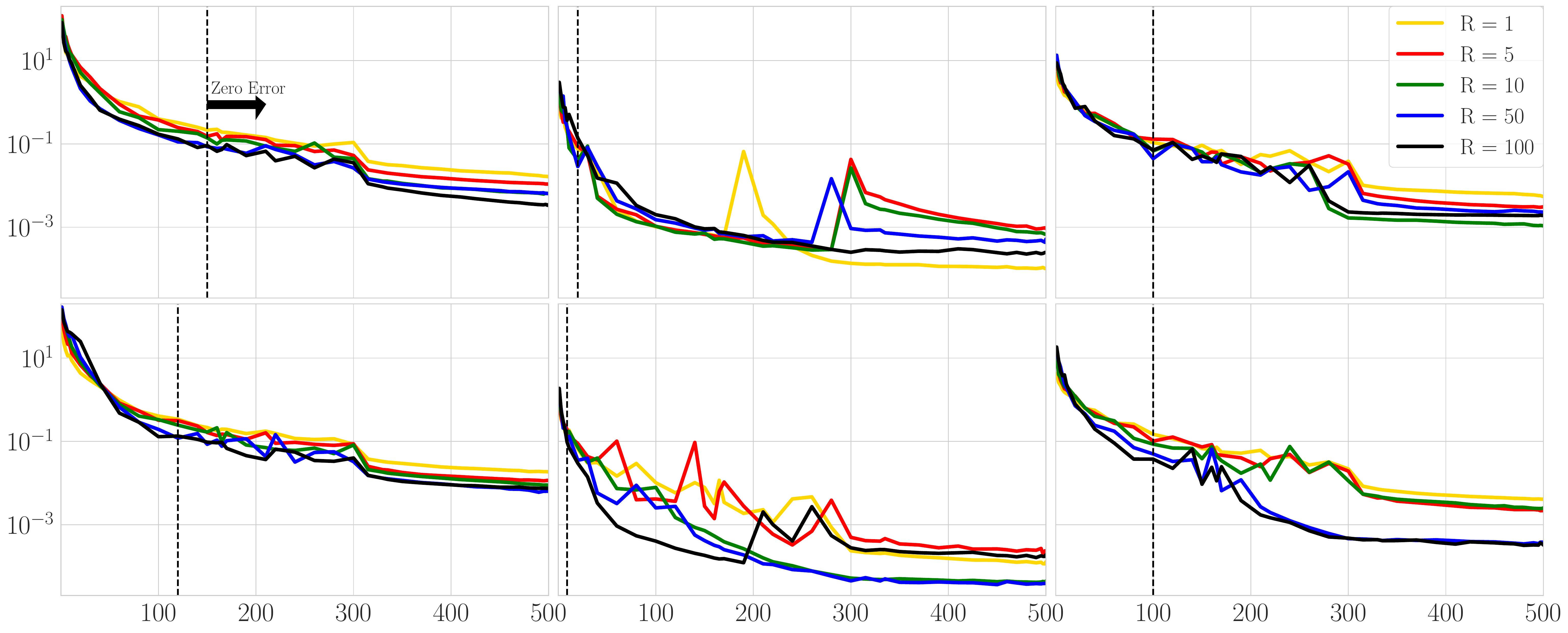}};
            \node at (-4,3.8) [scale=0.9]{\textbf{CIFAR10}};
            \node at (0.2,3.8) [scale=0.9]{\textbf{MNIST}};
            \node at (4.2,3.8) [scale=0.9]{\textbf{FMNIST}};
            \node at (-6.6,2.2)  [scale=0.9, rotate=90]{\textbf{Step}};
            \node at (-6.6,-0.2)  [scale=0.9, rotate=90]{\textbf{Long-tail}};
            \node at (0.1,-2) [scale=0.9]{\textbf{Epoch}};
        \end{tikzpicture}
        \caption{Neural Collapse}
    \end{subfigure}
    \begin{subfigure}[b]{\textwidth}
        \centering
        \begin{tikzpicture}
            \node at (0.0,0.9) 
            {\includegraphics[width=0.8\textwidth]{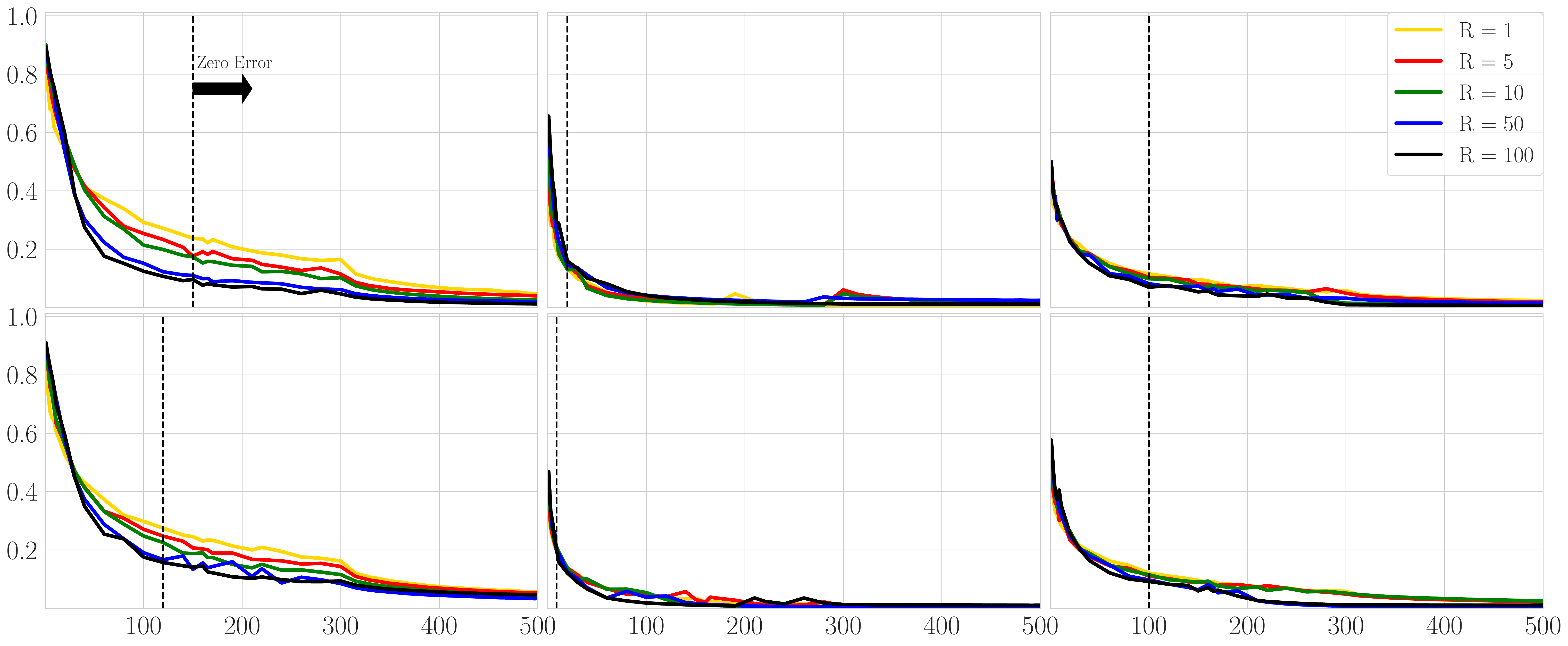}};
            \node at (-4,3.8) [scale=0.9]{\textbf{CIFAR10}};
            \node at (0.2,3.8) [scale=0.9]{\textbf{MNIST}};
            \node at (4.2,3.8) [scale=0.9]{\textbf{FMNIST}};
            \node at (-6.6,2.2)  [scale=0.9, rotate=90]{\textbf{Step}};
            \node at (-6.6,-0.2)  [scale=0.9, rotate=90]{\textbf{Long-tail}};
            \node at (0.1,-2) [scale=0.9]{\textbf{Epoch}};
        \end{tikzpicture}
        \caption{Angles between class-means}
    \end{subfigure}
    \caption
    {Geometric convergence of embeddings (as based on Def.~\ref{def:nc}, Def.~\ref{def:of}, and Def.~\ref{def:ncof}) of DenseNet-40 trained with batch-binding. We measure: \textbf{(a)} $\Delta_{\G_\M}$, \textbf{(b)} $\beta_{\text{NC}}$, and \textbf{(c)} $\text{Ave}_{c \neq c'} \alpha_{\text{sim}}(c,c')$ as defined in text.}
    \label{Fig:Deepnet_SCL_DenseNet_Geom}
\end{figure*}
\subsubsection{Impact of batch-binding on generalization}
To study the effect of binding examples on generalization, we  consider the \textit{Nearest Class Center} (NCC) classifier. For this, each test example is assigned the label of the class center $\mub_c$ learned during training that is closest to it (in Euclidean distance). We consider a simple setup of training a ResNet-18 model on CIFAR10 under imbalances $R = 10,100$ both with and without the binding examples. Models are trained with a batch size of $128$ for $350$ epochs. We ran the experiments with a smaller batch size to increase the number of back-propagation steps as adding data augmentations slows down convergence to OF geometry . We start with a learning rate of $0.1$, which is reduced by a factor of 10 on epochs $250$ and $300$. Weight decay is set to $5\times10^{-4}$.
In addition, we apply data augmentation as it is common practice when considering generalization and test accuracy. In particular, rather than simply adding horizontally flipped images (as described in Sec.~\ref{sec:Deepnet_SCLGeom_exp}) we allow for generic augmentations that include simple horizontal or vertical flips and random cropping with a probability of $0.5$. 
The NCC test accuracy was measured across $5$ versions of the experiment with the $k$ binding examples sampled randomly every time and $5$ versions without any batch-binding. The results for NCC balanced test accuracies are provided in Table~\ref{Tab:NCC_Bindings}. While making definitive conclusions regarding the impact of the embeddings geometries and binding examples on generalization requires further investigation, this preliminary investigation suggests that batch-binding does \emph{not} negatively impact NCC test accuracy.

Finally, Fig.~\ref{Fig:error_bars_plotGM} shows convergence of embeddings geometry to OF for the same experiments. As expected, the convergence is slightly slower in this case due to the inclusion of data augmentation (random crops and flips).





\end{document}